\documentclass{article}

\usepackage[final,nonatbib]{neurips_2022}

\usepackage{subcaption}
\usepackage{multirow}
\usepackage{multicol}
\usepackage{diagbox}
\usepackage{graphicx}
\usepackage{amsmath}
\usepackage{wrapfig}
\usepackage{algorithm}
\usepackage{algorithmic}

\usepackage[utf8]{inputenc} % allow utf-8 input
\usepackage[T1]{fontenc}    % use 8-bit T1 fonts
\usepackage{hyperref}       % hyperlinks
\usepackage{url}            % simple URL typesetting
\usepackage{booktabs}       % professional-quality tables
\usepackage{amsfonts}       % blackboard math symbols
\usepackage{nicefrac}       % compact symbols for 1/2, etc.
\usepackage{microtype}      % microtypography
\usepackage{xcolor}         % colors

\title{DreamShard: Generalizable Embedding Table Placement for Recommender Systems}

\author{%
Daochen Zha$^1$ \quad Louis Feng$^2$ \quad Qiaoyu Tan$^3$ \quad Zirui Liu$^1$ \quad Kwei-Herng Lai$^1$ \\
\textbf{Bhargav Bhushanam}$^2$ \quad \textbf{Yuandong Tian}$^2$ \quad \textbf{Arun Kejariwal}$^2$ \quad \textbf{Xia Hu}$^1$\\
$^1$Rice University \quad $^2$Meta Platforms, Inc. \quad $^3$Texas A\&M University\\
\texttt{\{daochen.zha,Zirui.Liu,khlai,Xia.Hu\}@rice.edu}\\
\texttt{\{lofe,bbhushanam,yuandong,akejariwal\}@fb.com}\\
\texttt{qytan@tamu.edu}
}

\begin{document}

\maketitle

% Embedding learning is a crucial technique in deep recommendation models by mapping sparse categorical features to dense vectors. However, the embedding tables can be extremely large, becoming storage and efficiency bottlenecks. As such, distributed training solutions have been developed to place the embedding tables to multiples devices such as GPUs and CPUs. In this work, 
\begin{abstract}
We study embedding table placement for distributed recommender systems, which aims to partition and place the tables on multiple hardware devices (e.g., GPUs) to balance the computation and communication costs. Although prior work has explored learning-based approaches for the device placement of computational graphs, embedding table placement remains to be a challenging problem because of 1) the operation fusion of embedding tables, and 2) the generalizability requirement on unseen placement tasks with different numbers of tables and/or devices. To this end, we present DreamShard, a reinforcement learning (RL) approach for embedding table placement. DreamShard achieves the reasoning of operation fusion and generalizability with 1) a cost network to directly predict the costs of the fused operation, and 2) a policy network that is efficiently trained on an estimated Markov decision process (MDP) without real GPU execution, where the states and the rewards are estimated with the cost network. Equipped with sum and max representation reductions, the two networks can directly generalize to any unseen tasks with different numbers of tables and/or devices without fine-tuning. Extensive experiments show that DreamShard substantially outperforms the existing human expert and RNN-based strategies with up to 19\% speedup over the strongest baseline on large-scale synthetic tables and our \emph{production} tables.
The code is available at \url{https://github.com/daochenzha/dreamshard}.
\end{abstract}
% , with neglectable performance drop on unseen tasks.

\section{Introduction}

%However, due to the large feature sizes, the embedding tables can be extremely large, causing storage and efficiency issues~\cite{zhao2020distributed}.

%The Meta recommendation model demands hundreds of embedding tables with multi-terabyte memory~\cite{acun2021understanding}.

Embedding learning is a commonly used technique to deal with categorical features in deep recommendation models by mapping sparse features into dense vectors~\cite{zhang2019deep,cheng2016wide,naumov2019deep,he2017neural,song2019autoint}. However, the embedding tables can be extremely large due to the large feature sizes~\cite{zhao2020distributed}. For example, in the YouTube recommendation model, a single categorical feature contains tens of millions of video IDs~\cite{covington2016deep}; the Meta recommendation model demands multi-terabyte memory~\cite{acun2021understanding}. Distributed training has been adopted to place the tables on multiple hardware devices such as GPUs~\cite{naumov2019deep,zhao2020distributed,amazon-dsstne,gupta2020architectural,naumov2020deep}. However, even with distributed training, the embedding tables are often still the efficiency bottlenecks. For instance, embedding lookup is shown to dominate the training throughput in the Meta recommendation model~\cite{acun2021understanding}. In our internal production model, which has hundreds of tables, embedding lookup accounts for 48\% and 65\% of the total computation and communication costs, respectively.

%(details not presented due to data privacy). 

%\textbf{Paragraph 1} The computational requirements for training recommendation models are large. Thus, distributed training solutions are adopted. A key challenge is how to split the model and put them into different devices. Embedding tables account for more 99\% of the models weights, and the majority of the computation time. Thus, embedding table placement is crucial.
%In this stage, all the devices will be ``synced up'' since they need to wait until all the dense vectors are received before they can start the follow-up dense computations (they are already well balanced by data partitioning, and thus omitted in the figure). 

\begin{wrapfigure}{R}{6.5cm}
  \centering
  \begin{subfigure}[b]{0.45\textwidth}
    \centering
    \includegraphics[width=0.99\textwidth]{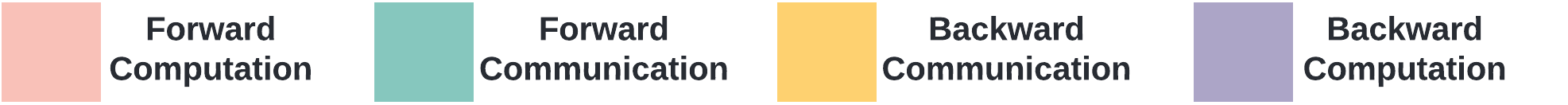}
  \end{subfigure}%
  
  \begin{subfigure}[b]{0.45\textwidth}
    \centering
    \includegraphics[width=0.99\textwidth]{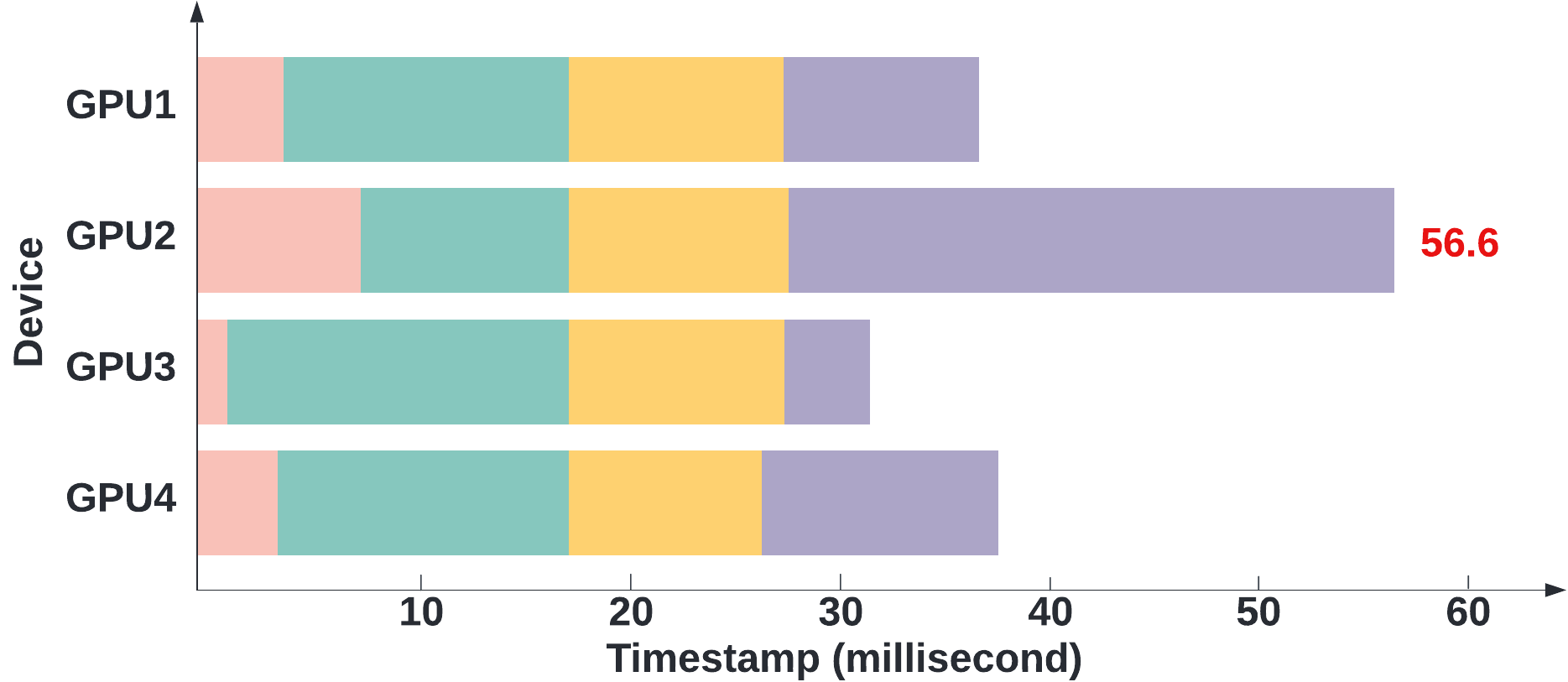}
    \subcaption{Random placement}
    \label{fig:casea}
  \end{subfigure}%
  
  \begin{subfigure}[b]{0.45\textwidth}
    \centering
    \includegraphics[width=0.99\textwidth]{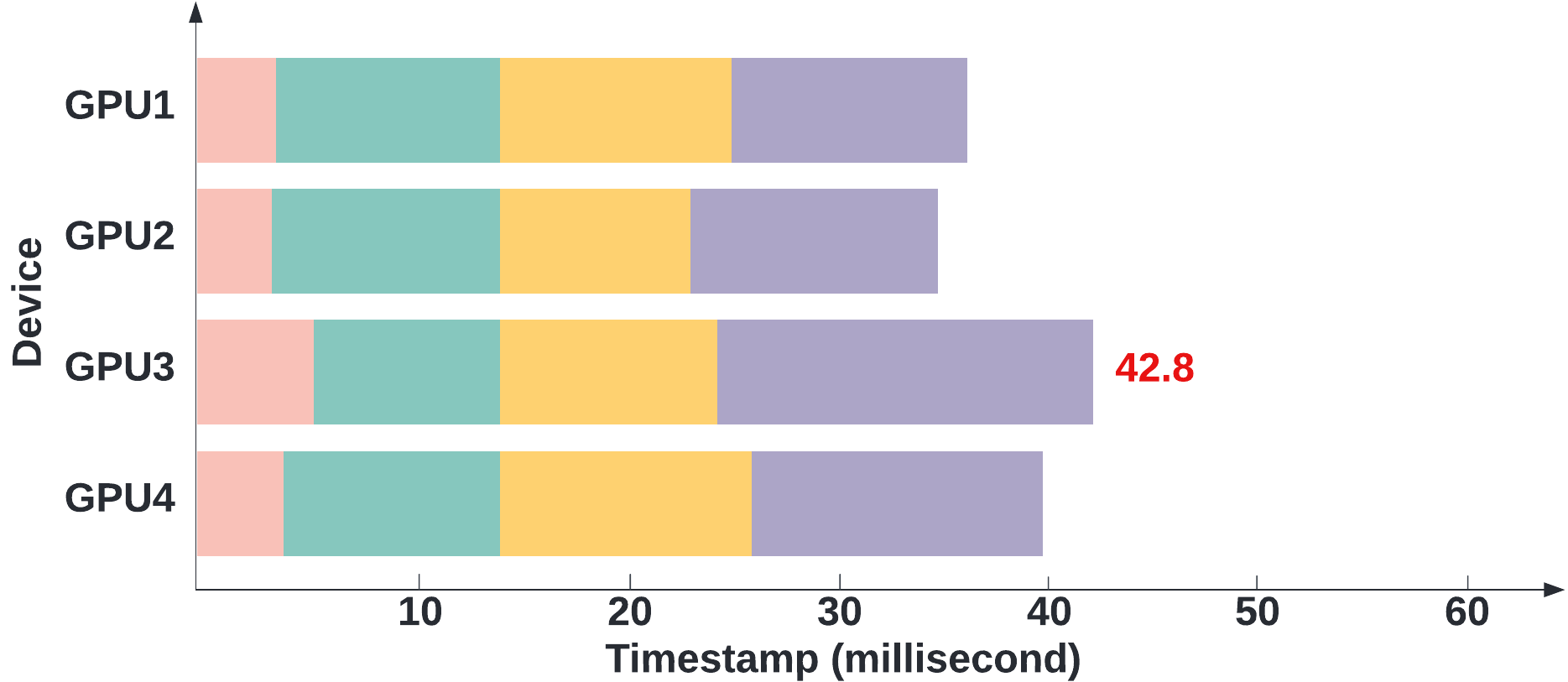}
    \subcaption{The existing best human expert strategy}
    \label{fig:caseb}
  \end{subfigure}%
  
  \begin{subfigure}[b]{0.45\textwidth}
    \centering
    \includegraphics[width=0.99\textwidth]{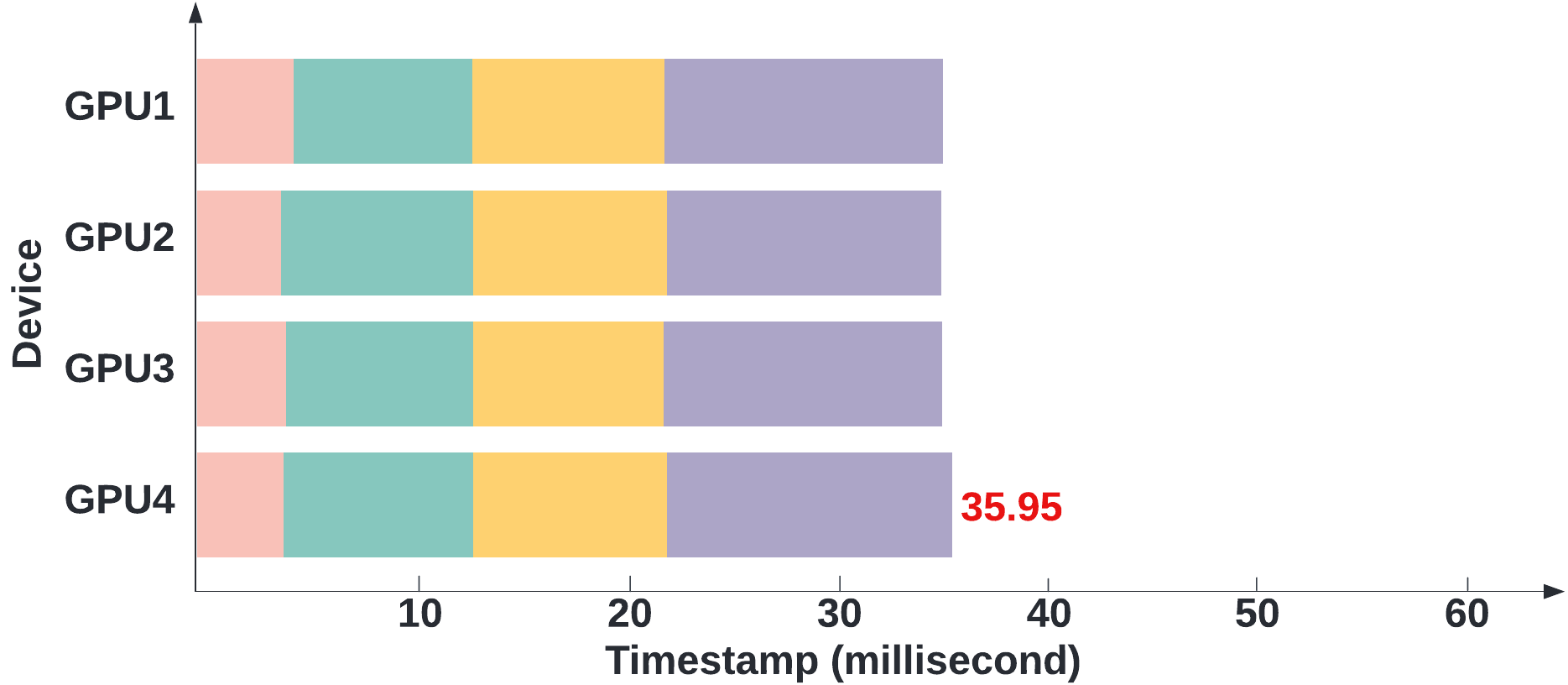}
    \subcaption{DreamShard}
    \label{fig:casec}
  \end{subfigure}%
 
  \caption{Visualization of random placement, the existing best human expert strategy, and DreamShard on a task of placing 50 tables on 4 GPUs. The dense computations and communications are omitted in the traces because they do not have an imbalance issue. We provide more visualizations in Appendix~\ref{appendix:K}.}
  \label{fig:case}
\end{wrapfigure}

{How the embedding tables are placed can significantly impact the costs. Figure~\ref{fig:case} shows the traces of different placement strategies on a task of placing 50 tables on 4 devices. Typically, embedding lookup consists of four stages. In the forward pass, the sparse indices are mapped into dense vectors (forward computation), which are then sent to the target devices (forward communication). \unskip\parfillskip 0pt \par}

\setlength\intextsep{0pt}

In the backward pass, the gradients of the embedding vectors are sent back from the target devices (backward communication) and applied to the embedding vectors (backward computation). The tables will easily lead to imbalances if not carefully partitioned. The random placement in Figure~\ref{fig:casea} is bottlenecked by GPU2 with a 56.6 milliseconds latency, while the more balanced placements in Figure~\ref{fig:caseb} and \ref{fig:casec} significantly reduce the costs to 42.8 and 35.95 milliseconds, respectively. This work asks: \emph{given a set of embedding tables, how can we identify the best placement of the tables to balance the costs?}

Device placement is essentially a partition problem, which is one of the classical NP-hard combinatorial optimization problems~\cite{wiki:Partition_problem}. A recent line of research uses reinforcement learning (RL) for device placement of computational graphs~\cite{mirhoseini2017device,mirhoseini2018hierarchical,gao2018spotlight,addanki2019placeto,zhou2019gdp,paliwal2019reinforced,gao2018post,goldie2020placement}. For example, \cite{mirhoseini2017device} proposed to train an RNN controller with content-based attention to predict the placement. Other studies advanced~\cite{mirhoseini2017device} in different ways, such as using hierarchical models~\cite{mirhoseini2018hierarchical}, more sophisticated RL algorithms~\cite{gao2018spotlight}, and graph neural networks~\cite{addanki2019placeto}.

However, embedding table placement remains to be an open and challenging problem due to the operation fusion~\cite{niu2021dnnfusion} of tables and the generalizability requirement. \textbf{1)} Modern embedding implementations (e.g., FBGEMM~\cite{fbgemm}), use a single operation to subsume multiple tables for acceleration. The speedup of the fused operation over the sum of the single-table operation costs is not constant and depends on the characteristics of the fused tables (e.g., table dimensions). Our analysis finds that the speedups vary significantly across different table combinations, ranging from 1X to 3X~(Figure~\ref{fig:multitableanalysis} in Appendix~\ref{appendix:a32}). Thus, we not only need to reason about cost balance but also how the tables should be fused to maximize the speedup. \textbf{2)} In real-world scenarios, the adopted embedding tables and the available devices can change frequently (e.g., machine learning engineers/researchers may conduct concurrent experiments with various table combinations and numbers of devices). Thus, a practical algorithm should generalize to tasks with unseen tables, different numbers of tables, and different numbers of devices. It is non-trivial to achieve this with the existing device placement approaches.

To this end, we introduce DreamShard, an RL approach for embedding table placement. DreamShard achieves the reasoning of operation fusion and generalizability with two novel ideas. \textbf{1)} It learns a cost network to directly predict the costs of the fused operations. Specifically, the network takes as input the table features (e.g., table dimension) of each single-table and outputs the computation and communication costs. \textbf{2)} It trains a policy network by interacting with an estimated Markov decision process (MDP) without real GPU execution, where the states and the rewards are estimated by the predictions of the cost network. Equipped with sum reductions for the table representations and max reductions for the device representations, the two networks can directly generalize to unseen placement tasks with different numbers of tables and/or devices without fine-tuning.

Extensive experiments show that DreamShard outperforms the existing human expert and RNN-based~\cite{mirhoseini2017device} strategies on open-sourced synthetic tables~\cite{dlrm_dataset} and our \emph{production} tables, achieving up to 19\% speedup over the strongest baseline. Moreover, it can generalize to unseen tasks that have different numbers of tables and/or devices with neglectable performance drop (< 0.5 milliseconds). Additionally, its inference is very efficient. It can place hundreds of tables in less than one second.

\section{Generalizable Embedding Table Placement Problem}
% It can be defined in a similar way as the placement for computational graphs~\cite{mirhoseini2017device}. 
The embedding table placement problem seeks a device placement\footnote{In this work, we focus on GPU devices, where all the GPU devices are identical, which is the most common configuration in our production. We defer the mixed scenarios of both GPUs and CPUs to future work.} of all the tables such that the overall cost (in terms of execution time) is minimized (we provide a background for the distributed training of recommendation models in Appendix~\ref{appendix:A1}). Consider $M$ embedding tables $\{\mathbf{e}_1, \mathbf{e}_2, ..., \mathbf{e}_M \}$ and $D$ devices, where $\mathbf{e}_i \in \mathbb{R}^{N}$ denotes the table features that characterize the embedding lookup patterns. In our work, we use 21 table features, including hash size, dimension, table size, pooling factor, and distribution (their definitions are provided in Appendix~\ref{appendix:A2}). A placement $\mathbf{a} = [a_1, a_2, ..., a_M]$, where $a_i \in \{1, 2, ..., D\}$, assigns each table to a device. Let $c(\mathbf{a})$ denote the cost measured on GPUs. The goal of embedding table placement is to find the $\mathbf{a}$ such that $c(\mathbf{a})$ is minimized. Due to the NP-hardness of the partition problem~\cite{wiki:Partition_problem}, identifying the exact solution demands extensive computational overhead. Thus, the state-of-the-art algorithms often approximate the optimal partition via sampling with RL~\cite{bello2016neural,mirhoseini2017device}. However, sampling remains expensive because obtaining $c(\mathbf{a})$ requires running operations on GPUs. Given that the embedding tables and the available devices can frequently change, we wish to approximate the best $\mathbf{a}$ without GPU execution.

Motivated by this, we study the \emph{generalizable embedding table placement} (GETP) problem. Let $\mathcal{E}$ be the space of all the embedding tables. A placement task can be denoted as $T_i=(\mathcal{E}_i, D_i)$, where $\mathcal{E}_i \subseteq \mathcal{E}$ is a set of tables, and $D_i$ is the number of devices. Given $N_\text{train}$ training tasks $\mathcal{T}_{\text{train}} = \{T_1, T_2, ..., T_{N_\text{train}}\}$, and $N_\text{test}$ testing tasks $\mathcal{T}_{\text{test}} = \{T_1, T_2, ..., T_{N_\text{test}}\}$, the goal is to train a placement policy based on $\mathcal{T}_{\text{train}}$ (GPU execution is allowed during training) such that the learned policy can minimize the costs for the tasks in $\mathcal{T}_{\text{test}}$ without GPU execution. 

%The main challenges of the GEPT problem are the difficulties of estimating $c(\mathbf{a})$ (we provide a comprehensive analysis in \textbf{Appendix~\ref{appendix:A3}}) and generating placements for any unseen tasks on the fly.

%it is hard to estimate $c(\mathbf{a})$ due to the non-linear relationship between the table cost and table features, operation fusion, and complex communication patterns. We provide a comprehensive analysis in \textbf{Appendix~\ref{appendix:A3}}.

\setlength\intextsep{0pt}
\begin{wrapfigure}{R}{6.5cm}
  \centering
  \vspace{-40pt}
  \includegraphics[width=0.46\textwidth]{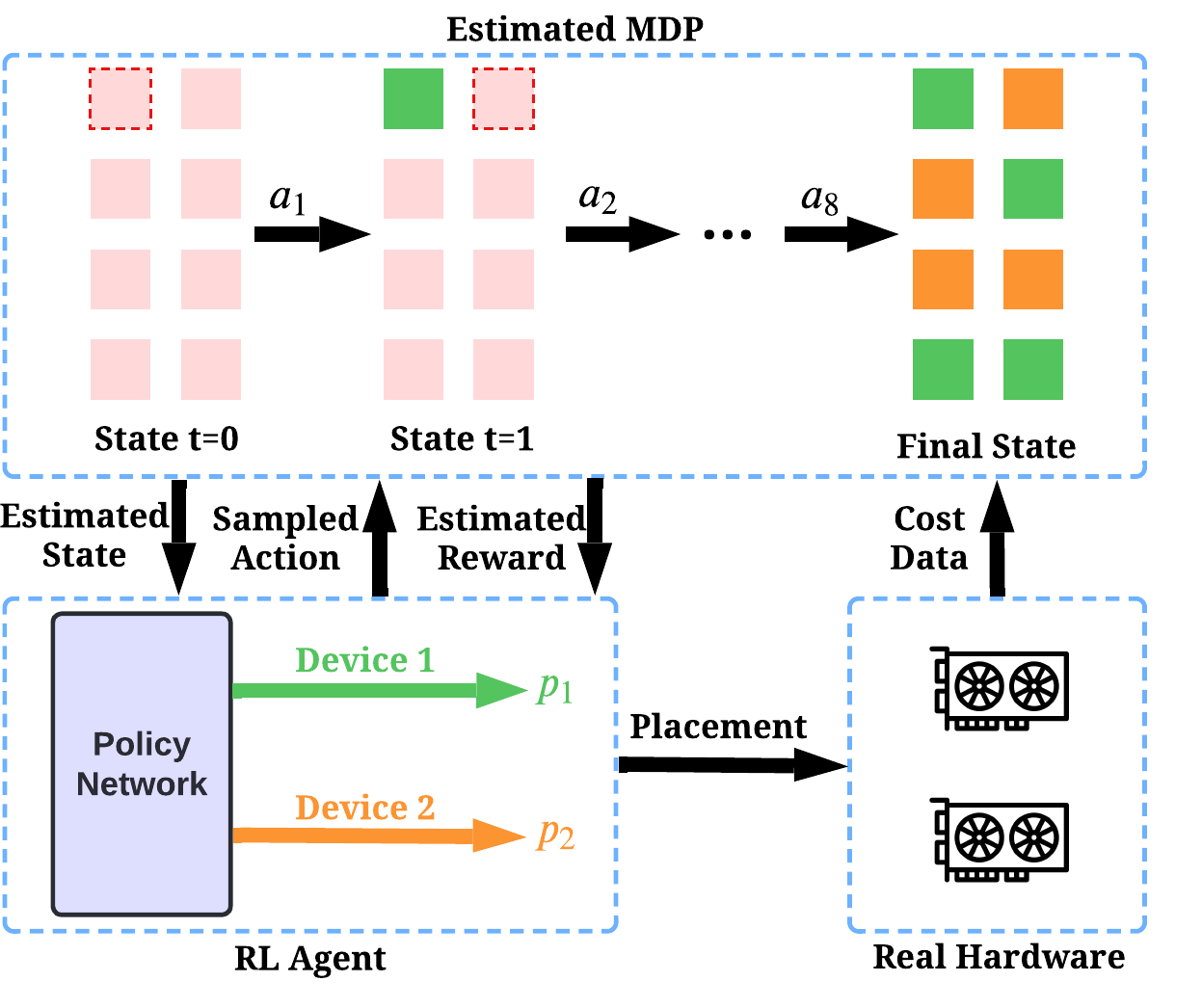}
  \vspace{-15pt}
  \caption{DreamShard framework. The agent interacts with the estimated MDP, which is trained with the cost data collected from GPUs.}
  \vspace{-20pt}
  \label{fig:overview}
\end{wrapfigure}

\section{DreamShard Framework}

We present DreamShard, an RL framework based on estimated MDP, to tackle the GETP problem. An overview of the framework is shown in Figure~\ref{fig:overview}. The key idea is to formulate the table placement process as an MDP (Section~\ref{sec:31}) and train a cost network to estimate its states and rewards (Section~\ref{sec:32}). A policy network with a tailored generalizable network architecture is trained by efficiently interacting with the estimated MDP (Section~\ref{sec:33}). The two networks are updated iteratively to improve the state/reward estimation and the placement policy.

%We jointly train the RL agent and the cost network, and generalize them to unseen placement tasks (Section~\ref{sec:34}).

\subsection{MDP Formulation}
\label{sec:31}

Given embedding tables $\{\mathbf{e}_1, \mathbf{e}_2, ..., \mathbf{e}_M \}$ and $D$ devices, we aim to generate a placement $\mathbf{a} = [a_1, a_2, ..., a_M]$. The key idea is to place the tables one by one at each step, where the state characterizes the tables that have been placed so far, the action is the device ID, and the reward represents the execution time on GPUs. Specifically, at a step $t$, the state $s_t=\{s_{t,d}\}_{d=1}^D$ is all the table features of the tables placed on all the devices, where $s_{t,d} = \{\mathbf{e}_i | i \in \mathcal{P}_d\}$ denotes all the table features corresponding to device $d$ ($\mathcal{P}_d$ is the set of table IDs that have been placed on device $d$). We further augment the raw features with cost features which are obtained by collecting the operation computation and communication times from GPUs (Appendix~\ref{appendix:A3} provides a comprehensive analysis of the cost features). Formally, the augmented state is defined as $\widetilde{s}_t=\{s_t, \{\mathbf{q}_{t,d}\}_{d=1}^{D}\}$, where $\mathbf{q}_{t,d} \in \mathbb{R}^3$ has three elements representing forward computation time, backward computation time, and backward communication time for the current operation in device $d$ (we provide detailed explanations of why forward communication time is 
excluded in Appendix~\ref{appendix:A4}). We find that the augmented cost features can significantly boost the performance, evidenced by the ablations in Table~\ref{tbl:ablation}. The action $a_t \in \mathcal{A}_t$ is an integer specifying the device ID, where $\mathcal{A}_t$ is the set of legal actions at step $t$. A device ID is considered legal if placing the current table on the corresponding device does not cause a memory explosion. The reward $r_t$ is $0$ for all the intermediate steps, and the reward at the final step $M$ is the negative of the cost, i.e., $r_{M} = -c(\mathbf{a})$, which encourages the agent to achieve lower cost.

\begin{figure}[t]
  \centering
    \includegraphics[width=\textwidth]{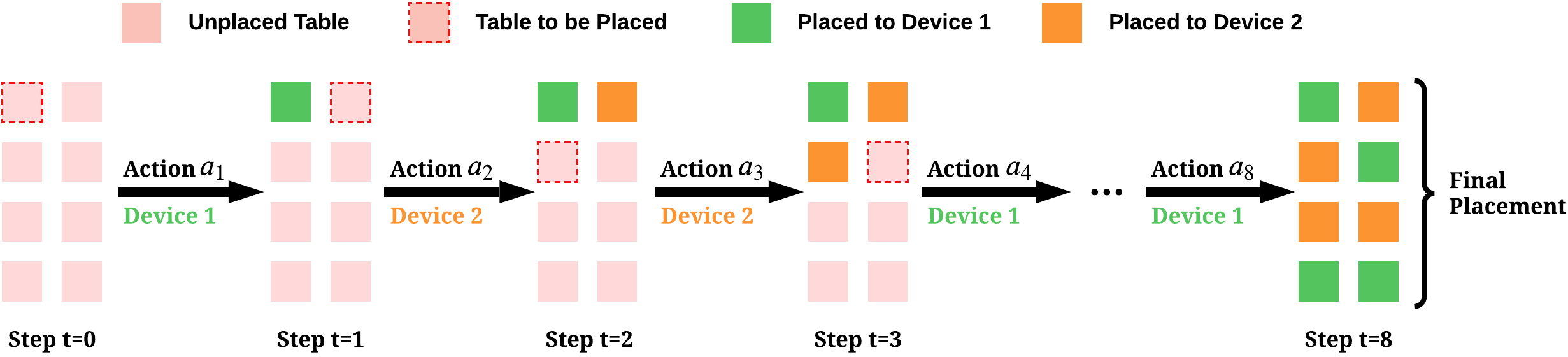}
  \caption{MDP formulation of embedding table placement.}
  \label{fig:mdp}
\end{figure}

\begin{figure}[t]
  \centering
    \includegraphics[width=0.9\textwidth]{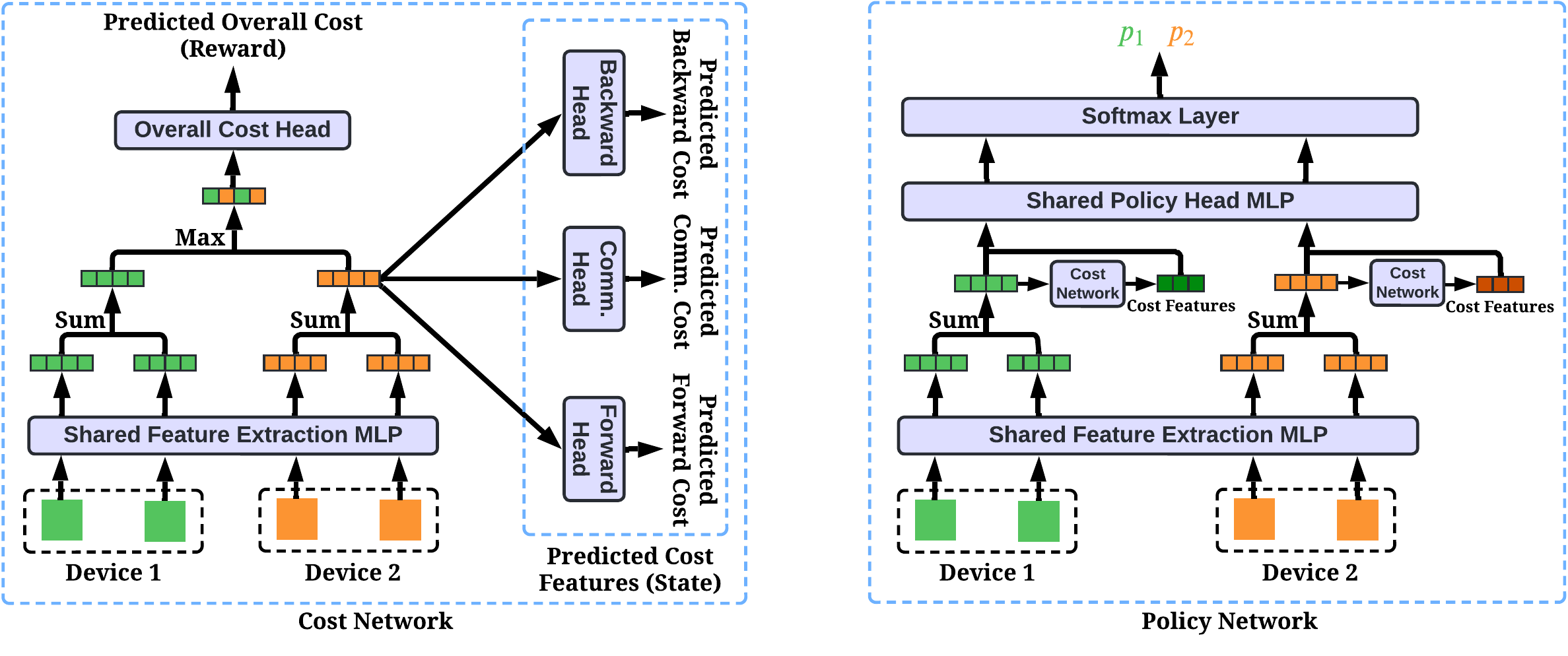}
  \caption{DreamShard's cost network (\textbf{left}) and policy network (\textbf{right}).}
  \label{fig:networks}
\end{figure}

The procedure is illustrated for an example task of placing $8$ tables $\{\mathbf{e}_1, \mathbf{e}_2, ..., \mathbf{e}_8 \}$ on $2$ devices in Figure~\ref{fig:mdp}. At step $0$, no table has been placed so $s_0 = \{\{\}, \{\}\}$ and the augmented state $\widetilde{s}_0=\{s_0, \{{\mathbf{q}_{0,0}, \mathbf{q}_{0,1}\}}\}$, where both $\mathbf{q}_{0,0}$ and $\mathbf{q}_{0,1}$ are zero vectors (i.e., $[0, 0, 0]$) since all the computation and communication times are 0 as well. Then the action $a_0=1$ makes the MDP transit to the next state $s_1 = \{\{\mathbf{e}_1\}, \{\}\}$ with its corresponding augmented state $\widetilde{s}_1=\{s_1, \{{\mathbf{q}_{1,0}, \mathbf{q}_{1,1}\}}\}$, where $\mathbf{q}_{1,0}$ becomes a non-zero vector containing the computation and communication costs by running $\{\mathbf{e}_1\}$ and $\{\}$ on GPUs. We repeat the above process, and finally at step $8$, we have $s_8 = \{\{\mathbf{e}_1, \mathbf{e}_4, \mathbf{e}_7, \mathbf{e}_8\}, \{\mathbf{e}_2, \mathbf{e}_3, \mathbf{e}_5, \mathbf{e}_6\}\}$. The corresponding $\mathbf{q}_{8,1}$, and $\mathbf{q}_{8,2}$ are the measured times of running $\{\mathbf{e}_1, \mathbf{e}_4, \mathbf{e}_7, \mathbf{e}_8\}$ and $\{\mathbf{e}_2, \mathbf{e}_3, \mathbf{e}_5, \mathbf{e}_6\}$ on two devices. The action sequence $\mathbf{a}=[a_1, a_2, ..., a_8]$ is the generated placement, which is then evaluated on GPUs to obtain the reward.

\textbf{Discussion 1.} The MDP enjoys two desirable properties. \textbf{1)} The legal action $\mathcal{A}_t$ can guarantee that the generated placement satisfies the memory constraints. \textbf{2)} The one-by-one placement enables the agent to be generalized across different numbers of tables. For example, an agent trained on an MDP with very few tables can be applied to another MDP with more tables by simply executing more steps.

\textbf{Discussion 2.} A straightforward idea to solve the MDP is to greedily place the current table on the device with the lowest cost at each step, where the cost function can be one of or a combination of the state features (e.g., the sum of the table dimensions, or the sum of all the cost features). However, greedy heuristics are often sub-optimal. Thus, we seek a learning-based algorithm to explore various placement possibilities and make comprehensive decisions based on all the state features.

% \zr{This formulation reminds me the classical Submodular optimization problem. I think they have the same setting. So from this, I think this greedy formulation makes sense if the embedding placement is a submodular. Otherwise I think it is sub-optimal.}

\subsection{Learning an Estimated MDP}
\label{sec:32}

%~\yuandong{This is not that accurate. World model means the transition/reward models.}
% which build parametric models to represent the world from past experience
% ~\yuandong{I would use a different phrase... ``estimated environment'' is better}

Interacting with the above MDP is computationally expensive since obtaining the cost features and the reward requires GPU execution. Motivated by world models~\cite{ha2018world,hafner2019dream}, we build an estimated MDP by approximating the cost features and the reward with a cost network. Let $f_{\text{cost}}$ denote the cost network. $f_{\text{cost}}$ takes as input the raw table features $s_t$, and predicts cost features $\{\mathbf{q}_{t,d}\}_{d=1}^{D}$ and the overall cost $c(\mathbf{a})$. $f_{\text{cost}}$ is trained with mean squared error (MSE) loss using the cost data collected from the GPUs. Once trained, it can predict the cost features or the reward with a single forward pass without GPU execution. However, it is non-trivial to design the architecture of $f_{\text{cost}}$ because it needs to accommodate different numbers of devices (i.e., $s_t$ can have variable sizes), and different numbers of tables in each device (i.e., $s_{t,d}$ can have variable lengths).

%~\yuandong{This part is important. I would put a figure to demonstrate the core design idea, and omit details.}

The left-hand side of Figure~\ref{fig:networks} shows DreamShard's generalizable design of $f_{\text{cost}}$, which is based on two key ideas. \textbf{First,} it uses a shared MLP to map raw table features into table representations. For any unseen tables, this MLP can be directly applied to extract table representations. \textbf{Second,} it enables a fixed-dimension representation for each device with sum reductions (i.e., the element-wise sum of the table representations in the device), and similarly for the overall representation across devices with max reductions (Appendix~\ref{appendix:B3} compares different reduction choices and finds that this sum-max combination leads to the most accurate prediction). The reduced representations are then followed by multiple MLP heads for cost predictions. For unseen tasks with different numbers of tables and/or devices, the reductions will always lead to fixed-dimension device/overall representations, so that the prediction heads can be directly applied. Appendix~\ref{appendix:B1} provides more details.

\begin{algorithm}[t]
\caption{Training of DreamShard}
\label{alg:1}
\setlength{\intextsep}{0pt} 
\begin{algorithmic}[1]
\STATE \textbf{Input:} Training tasks $\mathcal{T}_{\text{train}}$, the number of data collection steps $N_{\text{collect}}$, the number of cost network update steps $N_\text{cost}$, the batch size for updating the cost network $N_\text{batch}$, the number RL update steps $N_{\text{RL}}$, the number of considered episodes in each RL update $N_{\text{episode}}$
\STATE Initialize a cost network, a policy network, and a buffer
\FOR{iteration = 1, 2, ... until convergence}
    \FOR[Collect cost data from GPUs]{step = 1, 2, ... $N_{\text{collect}}$}
        \STATE Randomly sample a training task from $\mathcal{T}_{\text{train}}$
        \STATE Generate a placement by interacting with the estimated MDP using the policy network
        \STATE Evaluate the placement on the hardware and store the collected cost data to the buffer
    \ENDFOR
    \FOR[Update the cost network (no GPU execution)]{step = 1, 2, ... $N_{\text{cost}}$}
        \STATE Randomly sample $N_\text{batch}$ cost data from the buffer
        \STATE Update the cost network based on MSE loss (Eq.~\ref{eq:1} in Appendix~\ref{appendix:B41})
    \ENDFOR
    \FOR[Update the policy network (no GPU execution)]{step = 1, 2, ... $N_{\text{RL}}$}
        \STATE Randomly sample a training task from $\mathcal{T}_{\text{train}}$
        \STATE Collect $N_{\text{episode}}$ episodes by interacting with the estimated MDP using the policy network
        \STATE Update the policy network based on the policy gradient loss (Eq.~\ref{eq:2} in Appendix~\ref{appendix:B41})
    \ENDFOR
\ENDFOR
\end{algorithmic}
\end{algorithm}

\subsection{Training the Policy Network on the Estimated MDP}
\label{sec:33}

%\yuandong{The architecture can be summarized by a figure. We can move all the detailed descriptions to Appendix.}

\textbf{Generalizable policy network architecture.} Let $\pi$ be the policy network. $\pi$ maps the augmented state $\widetilde{s}_t=\{s_t, \{\mathbf{q}_{t,d}\}_{d=1}^{D}\}$ to action $a_t$, i.e., $a_t = \pi(\widetilde{s}_t)$. $\pi$ also adopts a generalizable design, shown in the right-hand side of Figure~\ref{fig:networks}. Like $f_{\text{cost}}$, $\pi$ uses a shared MLP and sum reductions to produce a fixed-dimension representation, which is then concatenated with the cost features to obtain the device representation. To accommodate the potentially variable action space (i.e., the number of available devices may vary), a shared MLP will process each device representation separately to obtain a confidence score, followed by a Softmax layer to produce action probabilities. This design allows $\pi$ to generalize across different numbers of devices. Appendix~\ref{appendix:B2} provides more details.

\textbf{Training and inference.} Algorithm~\ref{alg:1} summarizes the training procedure of DreamShard, which iteratively executes the following: \textbf{1)} collect cost data from GPUs based on the placements generated by the current policy, \textbf{2)} update the cost network with the previously collected cost data, and \textbf{3)} update the policy network by interacting with the current estimated MDP. Throughout the training process, the estimated MDP gradually becomes more accurate, and the resultant policy network tends to generate better placements. Appendix~\ref{appendix:B42} provides more details of the training procedure. For the inference, the trained cost network and policy network can be directly applied to unseen tasks to generate placements without GPU execution, which is summarized by Algorithm~\ref{alg:2} in Appendix~\ref{appendix:B43}.

%\yuandong{An algorithm block would be a lot cleaner. Also, make sure the audience knows there is no real GPU execution needed.}
%DreamShard is trained by iteratively collecting cost data from GPUs, and updating the cost network and the policy network as follows. \textbf{First,} we use the current policy $\pi$ to sample $N_{\text{collect}}$ placements from a randomly selected training task based on the current cost network $f_{\text{cost}}$, and collect the forward/communication/backward latencies and the overall latency. The collected data is stored in a buffer. \textbf{Second,} we sample $N_\text{cost}$ mini-batches of cost data from the buffer with a batch size of $N_\text{batch}$ to update $f_{\text{cost}}$. \textbf{Third,} we run $N_{\text{RL}}$ update steps for $\pi$. In each update step, we collect $N_{\text{episode}}$ episodes from a randomly selected training task via interacting with the estimated MDP with $\pi$, and update it with standard policy gradients~\cite{williams1992simple}. Here, $N_{\text{collect}}$, $N_\text{cost}$, $N_\text{batch}$, $N_{\text{RL}}$, $N_{\text{episode}}$ are hyperparameters. The inference of DreamShard is straightforward. We can simply generate a placement for any unseen task by interacting with the estimated MDP with the trained $f_{\text{cost}}$ and $\pi$. We provide more details such as the losses for training in \textbf{Appendix~\ref{appendix:B3}}.

\section{Experiments}
%\zr{I think we should report the training time to comprehensively compare the algorithm. If I missed, just ignore this comment.}
Our experiments aim to answer the following research questions. \textbf{RQ1:} How does DreamShard compare with the existing human expert and RL-based placement strategies? \textbf{RQ2:} Can DreamShard generalize to placement tasks with different numbers of tables and/or devices? \textbf{RQ3:} How efficient is the training of DreamShard? \textbf{RQ4:} How do the hyperparameters influence the performance of DreamShard? \textbf{RQ5:} How does each component of DreamShard contribute to the performance?  \textbf{RQ6:} How accurate is the estimated MDP and to what extend can it accelerate the training and inference?

\subsection{Experimental Setup}

\textbf{Datasets.} Academic recommendation datasets are often too small to enable a meaningful evaluation because the cost will always be very small no matter how the tables are placed. Thus, we use two industrial-scale datasets. \textbf{DLRM}\footnote{\label{dlrm_dataset}\url{https://github.com/facebookresearch/dlrm_datasets}} is a large-scale synthetic dataset with 856 tables, recently released by Meta. It shares memory access reuse patterns similar to those arising in Meta production workloads. \textbf{Prod} is an internal large-scale dataset for production recommendation models. It has a similar scale as DLRM. The main difference is that DLRM only has a fixed dimension for all the tables, while Prod is more challenging with diverse table dimensions, ranging from 4 to 768. For reproducibility, we mainly focus on the DLRM dataset since it is open-sourced. We only report the main results on the Prod dataset for verification purposes. We provide more details in Appendix~\ref{appendix:C}.

\textbf{Baselines.} We compare DreamShard against human expert strategies from previous work~\cite{sethi2022recshard,acun2021understanding,lui2021understanding}, including \textbf{size-based}, \textbf{dim-based}, \textbf{lookup-based}, \textbf{size-lookup-based} greedy balancing strategies. We also include an \textbf{RNN-based} RL algorithm~\cite{mirhoseini2017device}, which uses RNN architecture to map operators to devices. Since the feature extraction layers of the RNN-based method were designed for other operations instead of embedding tables, for a fair comparison, we adapt~\cite{mirhoseini2017device} by making it have the same feature extraction layers as in DreamShard. We provide more details in Appendix~\ref{appendix:D}.

%~\yuandong{Why DreamShard does not use RNN? Seem to be an easy try?}

\textbf{Configurations.} To evaluate the generalizability of DreamShard, we randomly divide the tables into a training pool $\mathcal{E}^{\text{train}}$ and a testing pool $\mathcal{E}^{\text{test}}$. The two pools have the same number of tables but they are not overlapped. A sharding task $T_i$ is constructed by randomly sampling a subset of $|\mathcal{E}_i|$ tables from a pool, where the number of tables $|\mathcal{E}_i| \in \{10, 20, 30, 40, 50, 60, 70, 80, 90, 100\}$ for the DLRM dataset, and $|\mathcal{E}_i| \in \{20, 40, 80\}$ for the Prod dataset. For all the experiments, we randomly sample 50 training and 50 testing tasks from $\mathcal{E}^{\text{train}}$ and $\mathcal{E}^{\text{test}}$, respectively. DreamShard is trained on the training tasks and will be evaluated on unseen tables in the testing tasks. We denote placement tasks with different numbers of tables and devices using the format of \texttt{dataset}-\texttt{num\_tables} (\texttt{num\_devices}). For example, DLRM-30 (4) suggests that there are 30 tables sampled from the DLRM dataset in each training/testing task with $4$ available devices. We provide more details in Appendix~\ref{appendix:E}.

% The number after ``-'' indicates the number of tables in each task, and the number in the parenthesis indicates the number of devices. 

\textbf{Implementation Details.} We use the same hyperparameters for all the experiments with $N_{\text{collect}} = 10$, $N_\text{cost} = 300$, $N_\text{batch} = 64$, $N_{\text{RL}} = 10$, $N_{\text{episode}} = 10$, $10$ training iterations, and an entropy weight of $0.001$ in the policy gradient. 2080 Ti GPUs and V100 GPUs are used for the DLRM (except that we use V100 for experiments with 8 GPUs) and Prod datasets, respectively. All the experiments are run 5 times, and we report the mean and the standard deviation. Appendix~\ref{appendix:B} provides more details.

%same Max-Memory = 5GB 
\subsection{Results and Analysis}
\label{sec:42}

\textbf{Evaluation of DreamShard against baselines (RQ1).} We perform qualitative and quantitative comparisons of DreamShard against the baselines. \textbf{Qualitatively,} Figure~\ref{fig:case} visualizes the traces of DreamShard and the baselines on one of the tasks from DLRM-50 (4). DreamShard achieves significant better overall cost than the best baseline (35.95 vs. 42.8) with \textbf{1)} a better balance of forward and backward commutation workloads, and \textbf{2)} less communication time, possibly due to a better balance of table dimensions. \textbf{Quantitatively,} Table~\ref{tbl:mainresults} presents comprehensive evaluations on tasks with different numbers of tables and devices on the DLRM and the Prod datasets. \textbf{Observations:} \textbf{1)} DreamShard outperforms the baselines on \emph{all} the tasks. \textbf{2)} DreamShard shows strong generalizability on unseen tables, achieving the same level of performance on \emph{all} the testing and training tasks. \textbf{3)} DreamShard appears to be more advantageous on harder tasks. Specifically, DreamShard achieves more improvement over the baselines on tasks with more tables/devices on the Prod dataset. In particular, DreamShard achieves 19\% improvment over the strongest baseline on Prod-80 (8). \textbf{4)} RNN-based method is only better than the random strategy on tasks with few tables/devices, but is worse on harder tasks. A possible reason is that RNN-based algorithm does not have a cost network, and using RL alone could lead to unstable performance. \textbf{5)} Lookup-based strategy is the best baseline on the DLRM dataset, while dim-based strategy is better on the Prod dataset. A potential reason is that the tables in the Prob dataset have very diverse table dimensions, while the tables in the DLRM dataset have the same dimension. As such, the dimensions on Prod tasks can more easily become imbalanced, leading to poor communication efficiency. Dim-based strategy can better balance the dimensions, which leads to a better overall performance. DreamShard outperforms the baselines on both the DLRM and the Prod datasets, showing its flexibility in dealing with different scenarios.

\begin{table}[]
\centering
%~\yuandong{Do we have number of steps, which is independent of the machine used for reproducibility.}
% \yuandong{I would prefer a figure rather than a table, or at least a table with decimated rows.}
\caption{Overall cost comparison in milliseconds and relative speedups over random placement, averaged over 50 different randomly sampled tasks. The placement tasks are denoted as \texttt{dataset}-\texttt{num\_tables} (\texttt{num\_devices}). For example, DLRM-30 (4) suggests that 30 tables are sampled from the DLRM dataset for each task with $4$ GPUs. More results are provided in Appendix~\ref{appendix:F}.}

%The numbers in the parentheses indicate the numbers of GPUs, e.g., DLRM-10 (4) suggest using 4 GPUs on DLRM-10, and Prod-80 (8) indicates suggest using 8 GPUs on Prod-80. More results of using 2 GPUs are provided in Appendix~\ref{appendix:F}.}
% Results of using 2 GPUs on DLRM tasks are provided in \textbf{Appendix~\ref{appendix:C}}
\label{tbl:mainresults}
\tiny
\setlength{\tabcolsep}{1.3pt}
\begin{tabular}{l|l|l|l|l|l|l|l|l}
\toprule
\multicolumn{2}{c|}{\multirow{2}{*}{Task}} & No strategy & \multicolumn{4}{c|}{Human Experts} & \multicolumn{2}{c}{RL} \\
\cline{3-9}
\multicolumn{2}{c|}{~} & \multicolumn{1}{c|}{Random} & \multicolumn{1}{c|}{Size-based} & \multicolumn{1}{c|}{Dim-based} & \multicolumn{1}{c|}{Lookup-based} & \multicolumn{1}{c|}{Size-lookup-based} & \multicolumn{1}{c|}{RNN-based} & \multicolumn{1}{c}{DreamShard} \\
\midrule
\multirow{2}{*}{DLRM-20 (4)} & Train & 24.0$\pm$0.6 & 22.7$\pm$0.0 (+5.7\%) & 21.3$\pm$0.0 (+12.7\%) & 19.1$\pm$0.0 (+25.7\%) & 19.1$\pm$0.0 (+25.7\%) & 22.4$\pm$0.5 (+7.1\%) & \textbf{18.6$\pm$0.2 (+29.0\%)} \\
~ & Test & 23.0$\pm$0.5 & 21.7$\pm$0.0 (+6.0\%) & 19.9$\pm$0.0 (+15.6\%) & 18.3$\pm$0.0 (+25.7\%) & 18.4$\pm$0.0 (+25.0\%) &20.9$\pm$0.3 (+10.0\%) & \textbf{17.6$\pm$0.2 (+30.7\%)} \\
\midrule
\multirow{2}{*}{DLRM-40 (4)} & Train & 41.3$\pm$0.2 & 39.6$\pm$0.0 (+4.3\%) & 37.4$\pm$0.1 (+10.4\%) & 33.6$\pm$0.0 (+22.9\%) & 33.6$\pm$0.1 (+22.9\%) & 39.2$\pm$0.7 (+5.4\%) & \textbf{32.8$\pm$0.3 (+25.9\%)} \\
~ & Test & 41.1$\pm$0.5 & 40.3$\pm$0.0 (+2.0\%) & 37.3$\pm$0.0 (+10.2\%) & 33.0$\pm$0.1 (+24.5\%) & 33.2$\pm$0.0 (+23.8\%) & 39.2$\pm$1.1 (+4.8\%) & \textbf{32.4$\pm$0.3 (+26.9\%)} \\
\midrule
\multirow{2}{*}{DLRM-60 (4)} & Train & 57.7$\pm$0.8 & 56.6$\pm$0.1 (+1.9\%) & 52.9$\pm$0.0 (+9.1\%) & 49.2$\pm$0.1 (+17.3\%) & 49.3$\pm$0.0 (+17.0\%) & 55.5$\pm$0.9 (+4.0\%) & \textbf{47.6$\pm$0.4 (+21.2\%)} \\
~ & Test & 58.1$\pm$0.6 & 59.6$\pm$0.1 (-2.5\%) & 53.7$\pm$0.0 (+8.2\%) & 48.7$\pm$0.2 (+19.3\%) & 49.1$\pm$0.1 (+18.3\%) & 56.0$\pm$0.7 (+3.8\%) & \textbf{47.9$\pm$0.7 (+21.3\%)} \\
\midrule
\multirow{2}{*}{DLRM-80 (4)} & Train & 75.7$\pm$1.0 & 76.0$\pm$0.0 (-0.4\%) & 70.0$\pm$0.3 (+8.1\%) & 64.8$\pm$0.0 (+16.8\%) & 65.3$\pm$0.1 (+15.9\%) & 73.2$\pm$2.7 (+3.4\%) & \textbf{62.2$\pm$0.2 (+21.7\%)} \\
~ & Test & 74.5$\pm$0.8 & 77.7$\pm$0.2 (-4.1\%) & 69.9$\pm$0.4 (+6.6\%) & 64.1$\pm$0.2 (+16.2\%) & 65.1$\pm$0.0 (+14.4\%) & 72.9$\pm$2.4 (+2.2\%) & \textbf{62.7$\pm$0.3 (+18.8\%)} \\
\midrule
\multirow{2}{*}{DLRM-100 (4)} & Train & 91.8$\pm$1.7 & 94.1$\pm$0.3 (-2.4\%) & 86.7$\pm$0.3 (+5.9\%) & 81.2$\pm$0.4 (+13.1\%) & 82.2$\pm$0.2 (+11.7\%) & 94.5$\pm$10.7 (-2.9\%) & \textbf{78.4$\pm$0.6 (+17.1\%)} \\
~ & Test & 94.5$\pm$6.5 & 95.4$\pm$0.0 (-0.9\%) & 84.7$\pm$0.4 (+11.6\%) & 79.5$\pm$0.3 (+18.9\%) & 80.8$\pm$0.3 (+17.0\%) & 94.8$\pm$13.0 (-0.3\%) & \textbf{77.8$\pm$0.8 (+21.5\%)} \\
\midrule
\midrule
\multirow{2}{*}{DLRM-40 (8)} & Train & 15.6$\pm$0.4 & 14.1$\pm$0.0 (+10.6\%) & 13.4$\pm$0.1 (+16.4\%) & \textbf{9.8$\pm$0.0 (+59.2\%)} & 9.9$\pm$0.0 (+57.6\%) & 16.2$\pm$0.8 (-3.7\%) & \textbf{9.8$\pm$0.6 (+59.2\%)} \\
~ & Test & 15.2$\pm$0.2 & 14.5$\pm$0.0 (+4.8\%) & 13.2$\pm$0.0 (+15.2\%) & 9.5$\pm$0.0 (+60.0\%) & 9.5$\pm$0.0 (+60.0\%) & 16.0$\pm$1.1 (-5.0\%) & \textbf{9.4$\pm$0.5 (+61.7\%)} \\
\midrule
\multirow{2}{*}{DLRM-80 (8)} & Train & 25.0$\pm$0.2 & 24.0$\pm$0.0 (+4.2\%) & 21.7$\pm$0.0 (+15.2\%) & 17.1$\pm$0.0 (+46.2\%) & 17.5$\pm$0.0 (+42.9\%) & 51.4$\pm$3.9 (-51.4\%) & \textbf{16.1$\pm$0.3 (+55.3\%)} \\
~ & Test & 25.2$\pm$1.3 & 25.6$\pm$0.5 (-1.6\%) & 20.8$\pm$0.0 (+21.2\%) & 16.7$\pm$0.2 (+50.9\%) & 16.9$\pm$0.1 (+49.1\%) & 53.4$\pm$4.6 (-52.8\%) & \textbf{16.1$\pm$0.4 (+56.5\%)} \\
\midrule
\multirow{2}{*}{DLRM-120 (8)} & Train & 34.0$\pm$0.3 & 32.3$\pm$0.0 (+5.3\%) & 29.8$\pm$0.0 (+14.1\%) & 24.5$\pm$0.0 (+38.8\%) & 25.3$\pm$0.0 (+34.4\%) & 58.6$\pm$2.7 (-42.0\%) & \textbf{23.3$\pm$0.2 (+45.9\%)} \\
~ & Test & 33.5$\pm$0.5 & 35.0$\pm$0.0 (-4.3\%) & 29.2$\pm$0.0 (+14.7\%) & 23.7$\pm$0.0 (+41.4\%) & 24.5$\pm$0.0 (+36.7\%) & 58.7$\pm$3.1 (-42.9\%) & \textbf{22.8$\pm$0.2 (+46.9\%)} \\
\midrule
\multirow{2}{*}{DLRM-160 (8)} & Train & 42.8$\pm$0.3 & 41.6$\pm$0.0 (+2.9\%) & 39.0$\pm$0.0 (+9.7\%) & 32.0$\pm$0.0 (+33.7\%) & 32.7$\pm$0.0 (+30.9\%) & 58.3$\pm$3.5 (-26.6\%) & \textbf{30.3$\pm$0.2 (+41.3\%)} \\
~ & Test & 41.1$\pm$0.0 & 42.4$\pm$0.0 (-3.1\%) & 36.4$\pm$0.0 (+12.9\%) & 30.8$\pm$0.0 (+33.4\%) & 31.6$\pm$0.0 (+30.1\%) & 59.3$\pm$5.4 (-30.7\%) & \textbf{29.6$\pm$0.2 (+38.9\%)} \\
\midrule
\multirow{2}{*}{DLRM-200 (8)} & Train & 51.5$\pm$1.2 & 48.2$\pm$0.0 (+6.8\%) & 48.0$\pm$0.0 (+7.3\%) & 38.9$\pm$0.0 (+32.4\%) & 39.9$\pm$0.0 (+29.1\%) & 68.7$\pm$2.4 (-25.0\%) & \textbf{37.2$\pm$0.2 (+38.4\%)} \\
~ & Test & 50.7$\pm$0.2 & 50.8$\pm$0.0 (-0.2\%) & 44.8$\pm$0.0 (+13.2\%) & 38.0$\pm$0.0 (+33.4\%) & 38.6$\pm$0.0 (+31.3\%) & 70.4$\pm$2.8 (-28.0\%) & \textbf{36.4$\pm$0.3 (+39.3\%)} \\
\midrule
\midrule
\multirow{2}{*}{Prod-20 (2)} & Train & 41.3$\pm$0.7 & 43.4$\pm$0.0 (-4.8\%) & 37.0$\pm$0.0 (+11.6\%) & 44.2$\pm$0.0 (-6.6\%) & 45.8$\pm$0.0 (-9.8\%) & 38.0$\pm$0.3 (+8.7\%) & \textbf{36.3$\pm$0.3 (+13.8\%)} \\
~ & Test & 42.8$\pm$0.4 & 46.1$\pm$0.0 (-7.2\%) & 39.5$\pm$0.0 (+8.4\%) & 45.9$\pm$0.0 (-6.8\%) & 45.7$\pm$0.0 (-6.3\%) & 39.3$\pm$0.6 (+8.9\%) & \textbf{37.5$\pm$0.2 (+14.1\%)} \\
\midrule
\multirow{2}{*}{Prod-40 (4)} & Train & 35.1$\pm$0.3 & 39.4$\pm$0.0 (-10.9\%) & 31.3$\pm$0.0 (+12.1\%) & 36.4$\pm$0.0 (-3.6\%) & 38.8$\pm$0.0 (-9.5\%) & 33.9$\pm$2.5 (+3.5\%) & \textbf{28.3$\pm$0.3 (+24.0\%)} \\
~ & Test & 38.3$\pm$0.3 & 43.6$\pm$0.0 (-12.2\%) & 33.5$\pm$0.0 (+14.3\%) & 37.4$\pm$0.0 (+2.4\%) & 40.1$\pm$0.0 (-4.5\%) & 36.7$\pm$2.8 (+4.4\%) & \textbf{30.4$\pm$0.7 (+26.0\%)} \\
\midrule
\multirow{2}{*}{Prod-80 (8)} & Train & 43.2$\pm$0.2 & 44.3$\pm$0.0 (-2.5\%) & 39.0$\pm$0.0 (+10.8\%) & 43.7$\pm$0.0 (-1.1\%) & 49.3$\pm$0.0 (-12.4\%) & 56.6$\pm$6.8 (-23.7\%) & \textbf{33.6$\pm$0.9 (+28.6\%)} \\
~ & Test & 47.7$\pm$0.4 & 53.9$\pm$0.0 (-11.5\%) & 41.9$\pm$0.0 (+13.8\%) & 46.1$\pm$0.0 (+3.5\%) & 49.6$\pm$0.0 (-3.8\%) & 62.5$\pm$4.2 (-23.7\%) & \textbf{35.2$\pm$0.8 (+35.5\%)} \\

\bottomrule
\end{tabular}
\end{table}

\textbf{Analysis of generalizability (RQ2).} In Table~\ref{tbl:generalizability}, we directly apply a DreamShard model trained from one task to another task \emph{without any fine-tuning} (the rightmost column), where the source and the target tasks have different numbers of tables and/or devices. DreamShard shows neglectable performance drop, suggesting that it is generalizable across different numbers of tables and/or devices.

\begin{table}[]
\centering
\caption{Generalization performance of DreamShard on target tasks w.r.t. to different numbers of tables (\textbf{top}) and devices (\textbf{bottom}), averaged over 50 randomly sampled tasks. Appendix~\ref{appendix:G} provides the results of more challenging tasks, such as mixed table-wise and device-wise transferring scenarios.}
%.~\yuandong{What is transferred DreamShard? Do you mean fine-tune?}
\label{tbl:generalizability}
\scriptsize
\setlength{\tabcolsep}{1.0pt}
\begin{tabular}{l|c|c|c|c}
\toprule

\multicolumn{1}{c|}{Source Task $\to$ Target Task} & Random & Best baseline strategy & DreamShard (trained on target task) & DreamShard (trained on source task)  \\
\midrule
DLRM-20 (4) $\to$ DLRM-100 (4) & 94.5$\pm$6.5 & 79.5$\pm$0.3 (+18.9\%) & \textbf{77.8$\pm$0.8 (+21.5\%)} & 77.9$\pm$0.4 (+21.3\%) \\
DLRM-40 (4) $\to$ DLRM-80 (4) & 74.5$\pm$0.8 & 64.1$\pm$0.2 (+16.2\%) & \textbf{62.7$\pm$0.3 (+18.8\%)} & \textbf{62.7$\pm$0.5 (+18.8\%)} \\
DLRM-80 (4) $\to$ DLRM-40 (4) &  41.1$\pm$0.5 & 33.0$\pm$0.1 (+24.5\%) & \textbf{32.4$\pm$0.3 (+26.9\%)} & \textbf{32.4$\pm$0.2 (+26.9\%)} \\
DLRM-100 (4) $\to$ DLRM-20 (4) & 23.0$\pm$0.5 & 18.3$\pm$0.0 (+25.7\%) & \textbf{17.6$\pm$0.2 (+30.7\%)} & 17.7$\pm$0.3 (+29.9\%) \\
\midrule
DLRM-20 (4) $\to$ DLRM-20 (2) & 29.9$\pm$0.4 & 26.0$\pm$0.0 (+15.0\%) & \textbf{25.8$\pm$0.2 (+15.9\%)} & \textbf{25.8$\pm$0.1 (+15.9\%)} \\
DLRM-40 (4) $\to$ DLRM-40 (2) & 58.6$\pm$0.7 & 52.4$\pm$0.0 (+11.8\%) & \textbf{51.9$\pm$0.1 (+12.9\%)} & \textbf{52.0$\pm$0.3 (+12.7\%)} \\
DLRM-20 (2) $\to$ DLRM-20 (4) & 23.0$\pm$0.5 & 18.3$\pm$0.0 (+25.7\%) & \textbf{17.6$\pm$0.2 (+30.7\%)} & 17.8$\pm$0.3 (+29.2\%) \\
DLRM-40 (2) $\to$ DLRM-40 (4) & 41.1$\pm$0.5 & 33.0$\pm$0.1 (+24.5\%) & \textbf{32.4$\pm$0.3 (+26.9\%)} & 32.6$\pm$0.3 (+26.1\%) \\

\bottomrule
\end{tabular}
\end{table}

\begin{figure}[!tbp]
  \centering
  \begin{minipage}[b]{0.49\textwidth}
      \begin{subfigure}[b]{0.5\textwidth}
        \centering
        \includegraphics[width=0.99\textwidth]{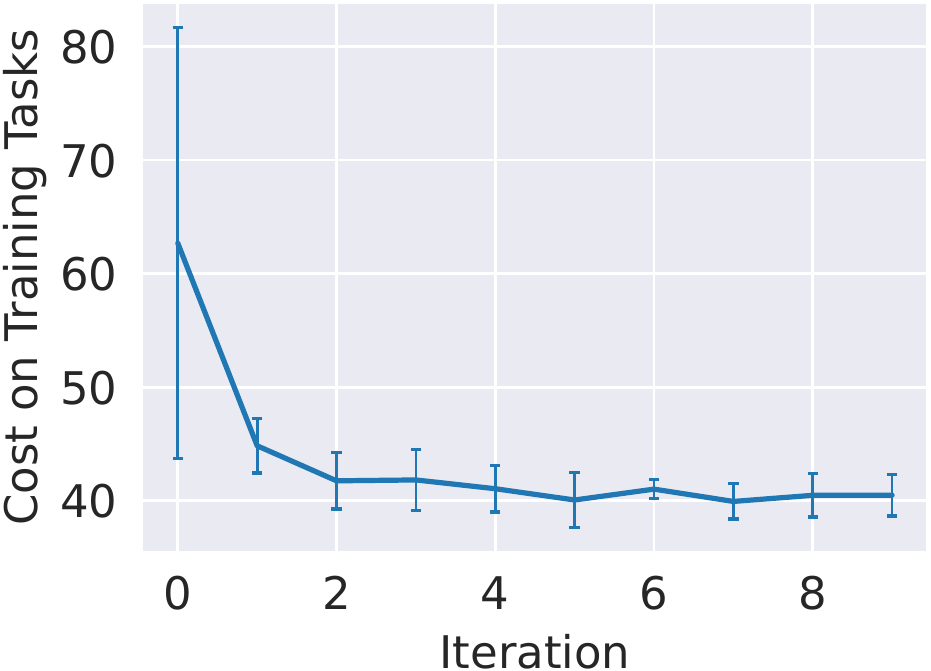}
      \end{subfigure}%
      \begin{subfigure}[b]{0.5\textwidth}
        \centering
        \includegraphics[width=0.99\textwidth]{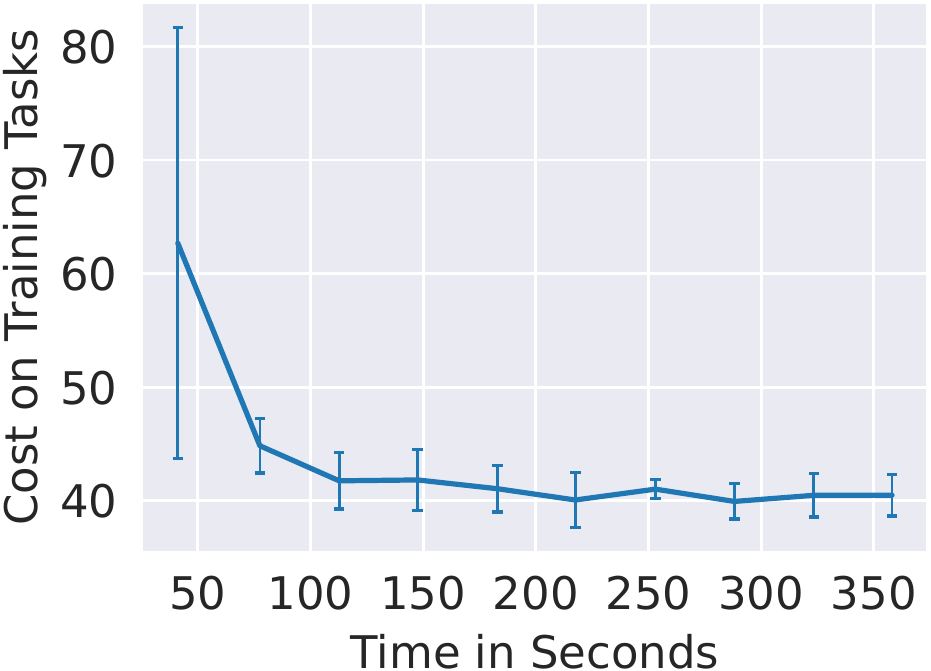}
      \end{subfigure}%
      \caption{Performance ($\downarrow$) of DreamShard on the DLRM-50 (4) datasets w.r.t. the numbers of iterations (\textbf{left}) and running time (\textbf{right}). Curves for other tasks are in Appendix~\ref{appendix:H}.}
      \label{fig:rq3}
  \end{minipage}
  \hfill
  \begin{minipage}[b]{0.49\textwidth}
      \begin{subfigure}[b]{0.5\textwidth}
        \centering
        \includegraphics[width=0.99\textwidth]{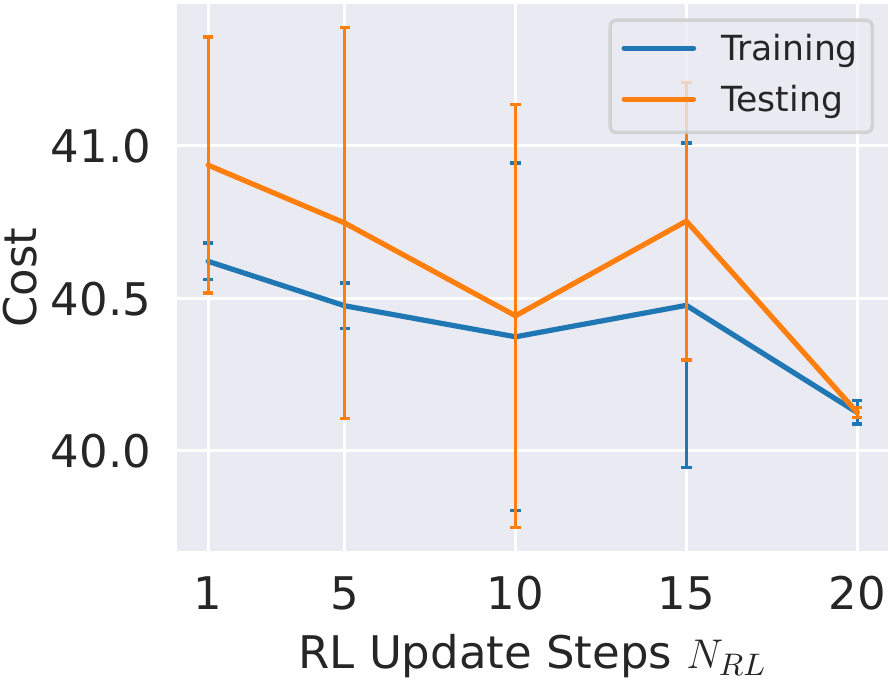}
      \end{subfigure}%
      \begin{subfigure}[b]{0.5\textwidth}
        \centering
        \includegraphics[width=0.99\textwidth]{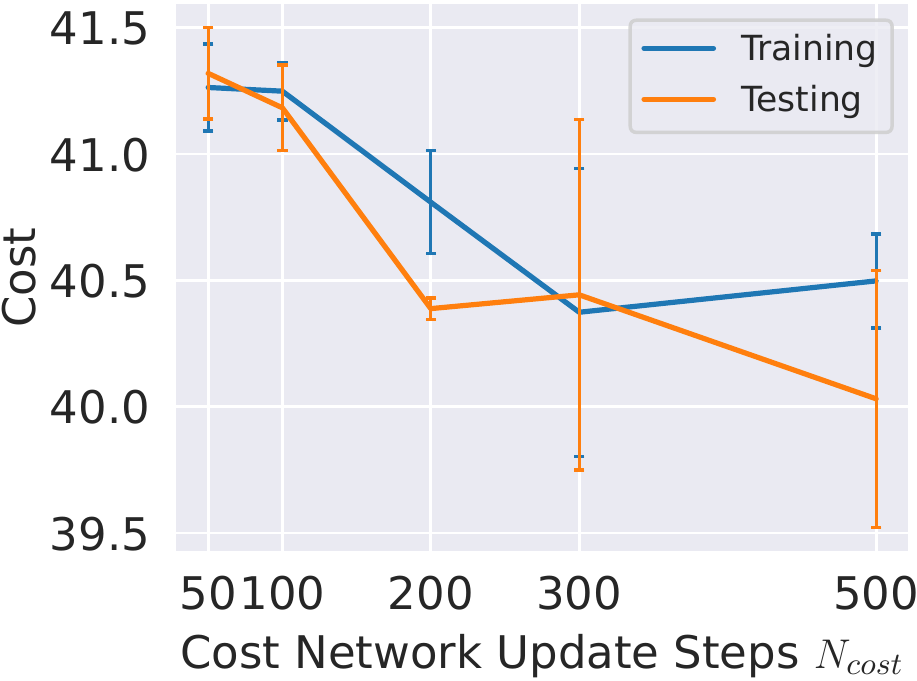}
      \end{subfigure}%
      \caption{Impacts of the number of RL update steps $N_{RL}$ (\textbf{left}) and cost network update steps $N_{\text{cost}}$ (\textbf{right}) on the DLRM-50 (4) dataset. Curves for other tasks are in Appendix~\ref{appendix:I}. }
      \label{fig:rq4}
  \end{minipage}
\end{figure}

\textbf{Analysis of training efficiency (RQ3).} Figure~\ref{fig:rq3} plots the performance of DreamShard w.r.t. the number of iterations and running time in seconds on four 1080Ti GPUs. The training of DreamShard is highly efficient. On DLRM-50 (4), it can achieve strong performance in less than 5 iterations or 200 seconds. Note that we only need to re-train or fine-tune DreamShard when the table pools have significant changes. Once trained, it only needs a forward pass for inference.

\textbf{Hyperparameter study (RQ4).} We study the impacts of two key hyperparameters: \textbf{1)} $N_{\text{RL}}$, which controls the RL update frequency, and \textbf{2)} $N_{\text{cost}}$, which determines the cost network update frequency. We vary one of them with the other fixed, shown in Figure~\ref{fig:rq4}. Increasing $N_{\text{RL}}$ or $N_{\text{cost}}$ will both lead to improvement, suggesting that both the cost network and the policy network need to be sufficiently trained. However, when $N_{\text{RL}}$ and $N_{\text{cost}}$ are large enough, increasing them will not bring more improvement. Considering that larger values will lead to more computational costs, we set $N_{\text{RL}}=10$ or $N_{\text{cost}}=300$ as a trade-off between the performance and the training efficiency.

\textbf{Ablation study (RQ5).} We study the importance of each table feature and check whether the RNN architecture helps, with the following ablations. \textbf{1)} We remove each of the features in the state. \textbf{2)} We add an RNN upon the device representation in the policy network. We makes several observations from the results in Table~\ref{tbl:ablation}. \textbf{1)} Cost features play a significant role, which demonstrates the effectiveness of our proposed augmented state. \textbf{2)} The most contributing table features are the pooling factor and the dimension, which aligns with our intuitions since these two feature are the determining factors of computation and communication workloads. \textbf{3)} Using more features leads to consistently good performance. \textbf{4)} While the policy makes decisions sequentially, RNN does not provide clear benefits. This is why we have kept the architecture simple with only MLP in DreamShard.

\begin{table}[t]
\centering
\caption{{Ablation study of DreamShard. Results on other tasks are in Appendix~\ref{appendix:J}}.}
\setlength{\tabcolsep}{8.0pt}
\label{tbl:ablation}
\scriptsize
\setlength{\tabcolsep}{1.0pt}
\begin{tabular}{l|c|c|c|c|c|c|c|c|c}
\toprule

\multicolumn{2}{c|}{Task} & w/o dim & w/o hash size & w/o pooling factor & w/o table size & w/o distribution & w/o cost & w/ RNN & DreamShard  \\
\midrule
\multirow{2}{*}{DLRM-50 (4)} & Train & 40.8$\pm$0.4 & 40.7$\pm$0.1 & 46.3$\pm$0.3 & 40.8$\pm$0.4 & 40.6$\pm$0.2 & 47.5$\pm$1.2 & 40.5$\pm$0.2 & \textbf{40.4$\pm$0.5}  \\
~ & Test & 40.9$\pm$0.6 & 40.6$\pm$0.3 & 47.2$\pm$0.1 & 40.6$\pm$0.7 & 40.5$\pm$0.2 & 46.3$\pm$0.1 & 40.5$\pm$0.1 & \textbf{40.4$\pm$0.6}  \\

\bottomrule
\end{tabular}
\end{table}

\begin{figure}[!tbp]
  \centering
  \begin{minipage}[b]{0.49\textwidth}
      \begin{subfigure}[b]{0.50\textwidth}
        \centering
        \includegraphics[width=0.99\textwidth]{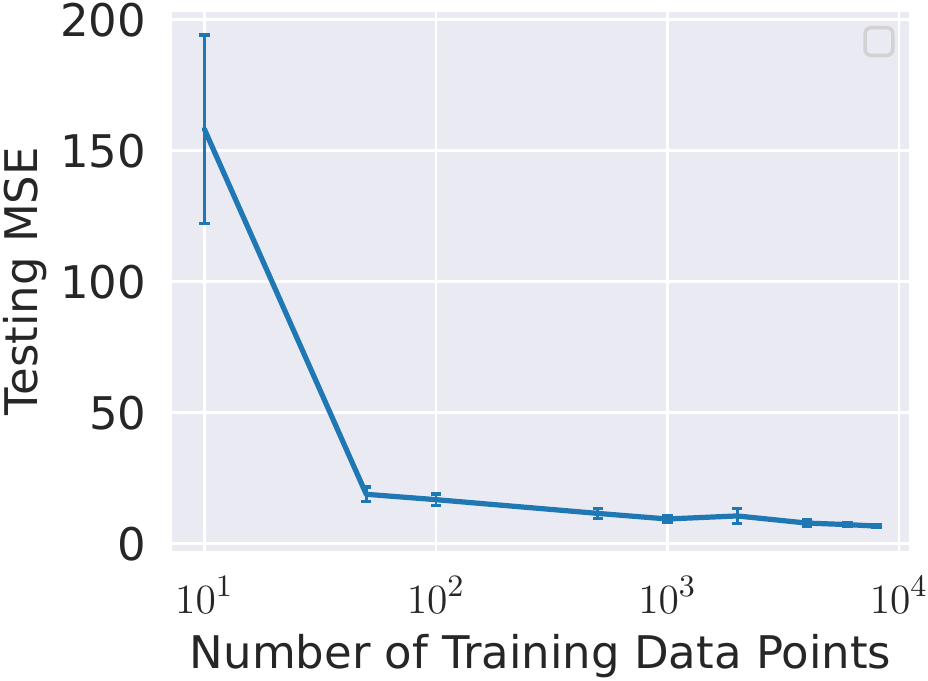}
      \end{subfigure}%
      \begin{subfigure}[b]{0.50\textwidth}
        \centering
        \includegraphics[width=0.99\textwidth]{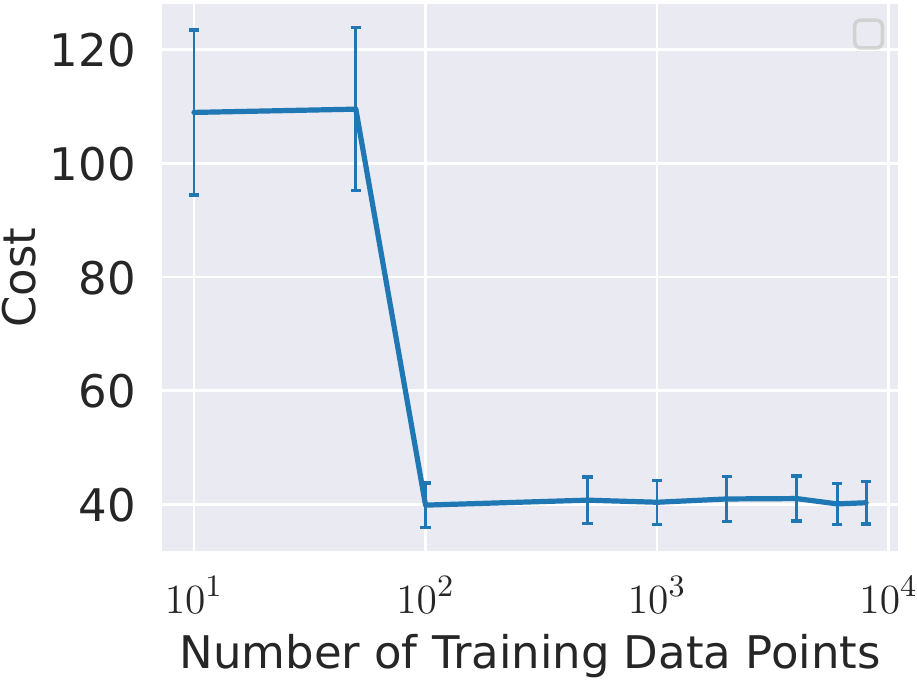}
      \end{subfigure}%
      \caption{MSE of the cost network w.r.t. the number of training data points collected from GPUs (\textbf{left}), and the corresponding performance of a fully trained policy network based on the trained cost network (\textbf{right}) on the DLRM-50 (4) dataset.}
      \label{fig:rq6-1}
  \end{minipage}
  \hfill
  \begin{minipage}[b]{0.49\textwidth}
      \begin{subfigure}[b]{0.50\textwidth}
        \centering
        \includegraphics[width=0.99\textwidth]{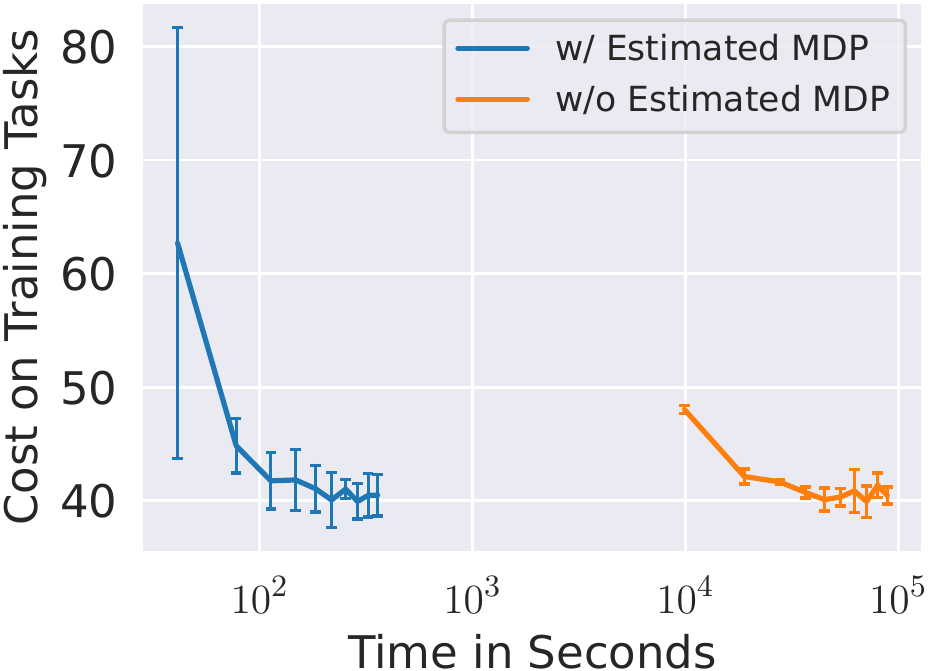}
      \end{subfigure}%
      \begin{subfigure}[b]{0.50\textwidth}
        \centering
        \includegraphics[width=0.99\textwidth]{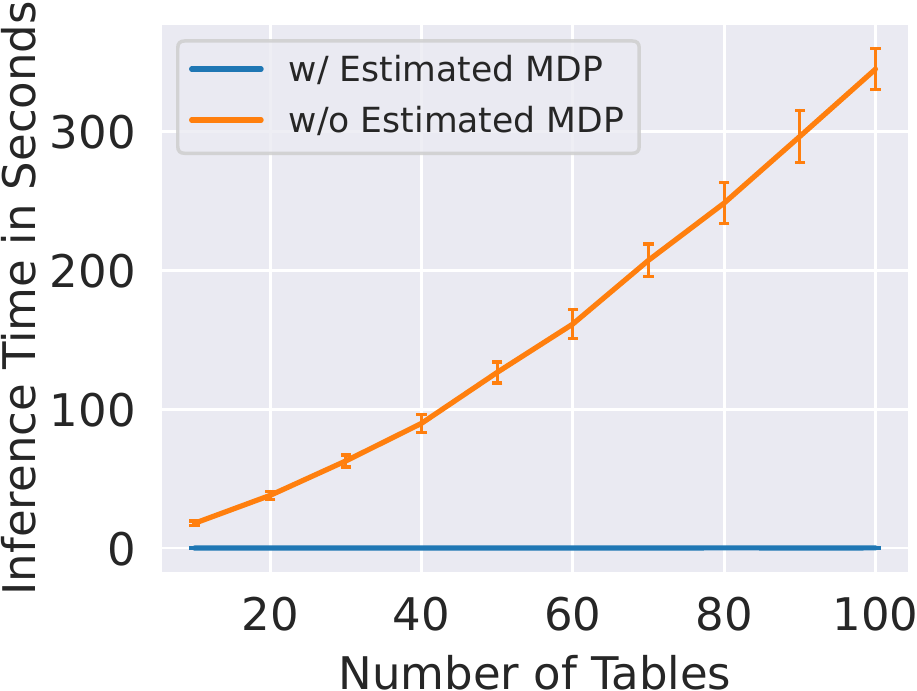}
      \end{subfigure}%
      \caption{Training curves w/ or w/o the estimated MDP on DLRM-50 (4) dataset (\textbf{left}), and their inference times w.r.t. the number of tables (\textbf{right}). Using cost network to estimate the MDP leads orders of magnitudes faster training and inference. }
      \label{fig:rq6-2}
  \end{minipage}
\end{figure}

\textbf{Study of the estimated MDP (RQ6).} \textbf{First,} we study how many data points are required to train an accurate cost network, and how accurate the cost network needs to be to enable a strong policy. Specifically, we randomly sample 10,000 cost data points from the DLRM-50 (4) dataset. Then we use 20\% for testing, and vary the size of the training data to train the cost network. Further, we fully train a policy network with 100 iterations based on each of the trained cost networks. We make two observations from the results in Figure~\ref{fig:rq6-1}. \textbf{1)} As expected, more data points lead to a more accurate cost network. \textbf{2)} Interestingly, after around 100 data points, the policy network does not keep improving even though the cost network becomes more accurate. Thus, we only need a sufficiently (but not perfectly) accurate cost estimation to achieve the best performance. This also partially explains why DreamShard can generalize: even though the cost network could be not very accurate on unseen tables, DreamShard can still find strong placements with the policy network. \textbf{Second,} we study the necessity of the estimated MDP. We consider a variant that obtains the cost features and rewards directly from GPUs, shown in Figure~\ref{fig:rq6-2}. Using the estimated MDP can make the training and the inference orders of magnitudes faster, while achieving the same level of performance. In particular, the inference time is less than one second even with a hundred tables.

%show the training and inference time with or without the estimated MDP. The results suggest that the estimated MDP can boost the efficiency, making DreamShard orders of magnitudes faster in both training and inference.

\section{Related Work}

%\zr{To me, RL-based method is also heuristic. The difference is that we have better/more accurate heuristics/cost model. I think in the CS, the opposite of heuristic is provable.}

%Thus, distributed training solutions have been developed to place the embedding tables on multiple devices~\cite{}. This leads to the embedding table placement problem since the placements have a significant impact on the computation and communication costs.

\textbf{Embedding tables.} Embedding tables are commonly used to deal with sparse features in recommendation models~\cite{zhang2019deep,cheng2016wide,naumov2019deep,he2017neural,song2019autoint,wang2018billion,zhang2016collaborative,zhao2019aibox}. However, the extremely large embedding tables are often the storage and efficiency bottlenecks~\cite{zhao2020distributed,covington2016deep,acun2021understanding,naumov2019deep,zhao2020distributed,amazon-dsstne,gupta2020architectural,naumov2020deep,lian2021persia}. To our knowledge, the only two studies that target the embedding table placement problem are RecShard~\cite{sethi2022recshard} and our previous work AutoShard~\cite{zha2022autoshard}. RecShard approaches table placement at a per-row granularity through exploitation of the underlying feature distributions. RecShard leverages these distributions, along with characteristics of the training system, to shard embedding tables across a tiered memory hierarchy using a mixed integer linear program, with more frequent rows placed in GPU HBM and the remaining placed in CPU DRAM. In contrast, our work develops a neural cost network for cost estimation and an RL-based optimization algorithm, and focuses on sharding across a single memory layer as opposed to a tiered memory hierarchy. AutoShard also leverages RL for embedding table sharding. However, it only balances the computational costs, which will lead to sub-optimal solutions since communication also account for significant costs. In addition, our DreamShard is more efficient in training than AutoShard due to the design of the estimated MDP. Another line of work focuses on reducing the embedding table sizes~\cite{zhang2020model,shi2020compositional,zhao2020autoemb,joglekar2020neural,liu2020learnable,kang2020learning,kang2021learning}, which is orthogonal to our work since DreamShard is also applicable to compressed tables.

\textbf{Device placement optimization.} The existing device placement techniques can be mainly grouped into two categories: RL-based algorithms, and cost modeling methods. \textbf{1)} RL-based algorithms treat the device placement as a black box and optimize the cost objective in a trial-and-error fashion~\cite{mirhoseini2017device,mirhoseini2018hierarchical,gao2018spotlight,addanki2019placeto,zhou2019gdp,paliwal2019reinforced,gao2018post,goldie2020placement}. Unfortunately, these methods are computationally expensive and often require training from scratch on unseen tasks. While Placeto~\cite{addanki2019placeto} shows generalizabily on computational graphs with graph embeddings, it can not deal with various table combinations since there is no graph structure, and it can not handle different numbers of devices. \textbf{2)} Cost modeling methods build a cost model to reflect the real performance and adopt offline algorithms (e.g., scheduling, and dynamic programming) to optimize the placement~\cite{lawler1993sequencing,jia2019beyond,jia2018exploring,narayanan2019pipedream,tarnawski2020efficient}. However, the cost model could be inaccurate. In particular, they can not deal with the operation fusion of embedding tables. Whereas, DreamShard combines the advances of RL with a neural cost model for accurate cost prediction.

%enabling an effective, efficient, and generalizable solution for embedding table placement.

\textbf{Deep RL.} Deep RL has recently made significant progress in games~\cite{mnih2013playing,silver2017mastering,zha2021douzero,schulman2017proximal,zha2021simplifying,lillicrap2016continuous,andrychowicz2017hindsight,zha2020rank,zha2021rlcard}. Our work is related to using RL to optimize machine learning model designs, such as neural architecture search~\cite{zoph2016neural,wang2022auto,li2021automated,pham2018efficient,li2021autood}, data augmentation~\cite{cubuk2018autoaugment}, data sampling~\cite{zha2020meta,zha2022towards}, pipeline search~\cite{drori2021alphad3m,zha2021autovideo,lai2021tods}. However, these methods often only focus on one task and can not generalize to unseen tasks. Our work is related to meta-learning~\cite{finn2017model,zhao2021automatic,zhao2022towards}. Instead of performing meta-learning for machine learning tasks, we focus on machine learning system design. Our work is also related to solving combinational optimization problems with RL~\cite{bello2016neural,barrett2020exploratory}. Unlike the above studies, we show that RL can tackle a practical problem of embedding table placement and the learned policy is generalizable.

%Unlike~\cite{barrett2020exploratory}, our work is application-driven, and we tackle a real-world combinational optimization challenge in industrial recommender systems.

%  Moreover, the above methods cannot deal with the complex communication patterns brought by the combined model- and data-parallelism in recommender systems.

\section{Conclusions and Future Work}

We present DreamShard for embedding table placement in recommender systems. We formulate the problem as an MDP which places the tables one by one at each step. Then we leverage RL to solve the MDP. To accelerate the training and the inference, we build an estimated MDP by training a cost network to approximate the state features (i.e., computation and communication times) and the reward (i.e., the overall cost), leading to orders of magnitudes faster training and inference speeds. Extensive experiments on the open-sourced DLRM dataset and our production dataset demonstrate the superiority of DreamShard over the existing algorithms. Moreover, DreamShard shows strong generalizability, making it a desirable choice in real-world applications. In the future, we will extend DreamShard to tiered memory hierarchy and large-scale training clusters with complex topologies.

\section*{Acknowledgements}
The work is, in part, supported by NSF (\#IIS-2224843). The views and conclusions in this paper are those of the authors and should not be interpreted as representing any funding agencies. We would also like to thank the helpful feedback from the anonymous reviewers.

\bibliographystyle{unsrt}
\bibliography{reference}

%%%%%%%%%%%%%%%%%%%%%%%%%%%%%%%%%%%%%%%%%%%%%%%%%%%%%%%%%%%%

%%%%%%%%%%%%%%%%%%%%%%%%%%%%%%%%%%%%%%%%%%%%%%%%%%%%%%%%%%%%

\newpage

\appendix

\section{Background of Embedding Table Placement}
\label{appendix:A}

In this section, we provide a background of \emph{embedding table placement} problem (also called \emph{embedding table sharding}~\cite{sethi2022recshard,acun2021understanding} since it essentially partitions the tables across different devices). In Section~\ref{appendix:A1}, we introduce a background of distributed training of recommendation models. Section~\ref{appendix:A2} lists some important table features, which characterize the table accessing patterns and are highly related to computation/communication costs. Section~\ref{appendix:A3} further provides an in-depth analysis of the correlation between the computation/communication costs and the table features. Finally, we discuss the difference between forward and backward communication times in Section~\ref{appendix:A4}.

\subsection{Distributed Training of Recommendation Models and Embedding Tables}
\label{appendix:A1}

Industrial recommendation models often require massive memory consumption and high training throughput. Thus, distributed training solutions have been developed to train recommendation models. While various recommendation models have been developed in the past decades, they often rely on embedding tables to map sparse categorical features to dense vectors~\cite{acun2021understanding,amazon-dsstne,covington2016deep,zhou2019deep,liu2017related,gomez2015netflix}. We take DLRM~\cite{naumov2019deep} as an example to introduce distributed training design since DLRM is the core of the official package of PyTorch for recommendation models\footnote{\url{https://github.com/pytorch/torchrec}}  and is commonly used in both academia and industry.

\begin{figure}[h!]
  \centering
    \includegraphics[width=0.70\textwidth]{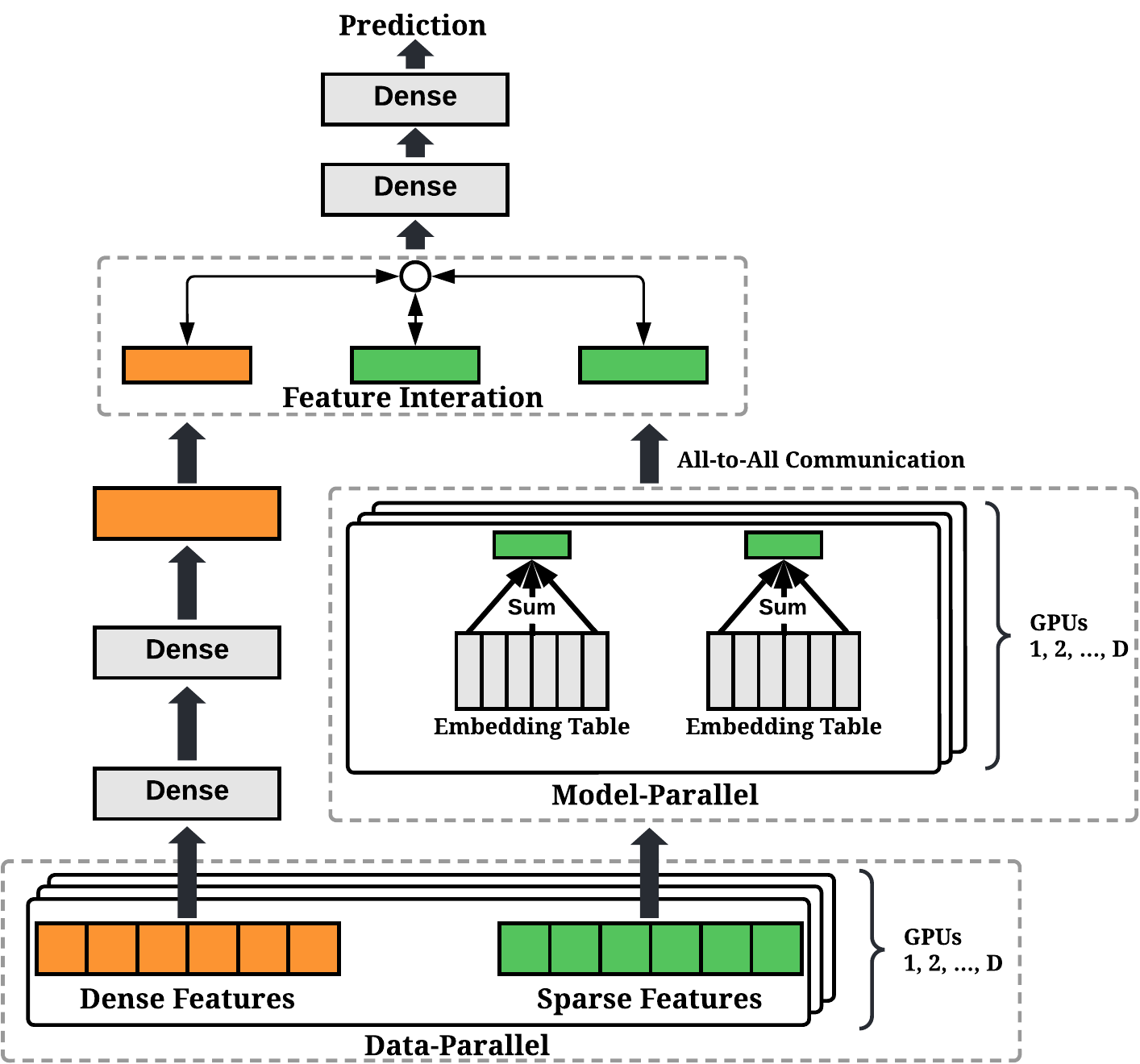}
  \caption{A typical distributed training framework for recommendation models~\cite{naumov2019deep}. The figure is adapted from~\cite{zha2022autoshard}.}
  \label{fig:dlrm}
\end{figure}

Figure~\ref{fig:dlrm} shows an overview of the DLRM model. DLRM processes two types of features, i.e., dense features, and sparse features. Dense features are numerical values and are directly processed with MLPs in DLRM. Sparse features are categorical features. For example, in the context of YouTube video recommendation, a possible sparse feature can be video IDs. For the sparse features, DLRM adopts embedding tables to map the categorical features to dense vectors. Specifically, each row of an embedding table corresponds to a feature value (i.e., video ID), and the number of columns corresponds to the vector dimension. Given a list of feature values, an embedding table lookup is performed to obtain the vectors. For each feature value, a corresponding vector is obtained from the table via hashing. Then all the obtained vectors are summed to obtain a fixed-dimension vector. The embedding lookup is performed for all the tables. The obtained vectors are processed by MLPs, and then will be interacted with the dense representations to obtain the final representation. The final MLP will be processed by another MLP, which maps the representation to the predictions (e.g., click-through rate).

However, in real-world applications, the embedding tables can become extremely large and can not be fed into a single GPU. Meanwhile, the dataset can also be extremely large so using one GPU may not meet the high training throughput requirement. To accommodate the massive memory and training throughput requirements, DLRM adopts a combination of data-parallelism and model-parallelism. For the data-parallelism, DLRM replicates MLPs on each device and partitions training data into different devices. In this way, each device only needs to process its own mini-batch of data, achieving higher training throughput. For the model-parallelism, the embedding tables are placed on different devices. With this design, in the forward pass, the embedding lookup for a certain feature value will be performed by querying the device that actually holds the corresponding table. For example, suppose a feature value in the training data of device 1 corresponds to table 1. If table 1 is unfortunately placed in device 2, then device 1 will query device 2 to obtain the vector via communication. The above communication is essentially very frequent if we feed a batch of data for training. Thus, in actual implementations, such communication is often batched by sending a batch of data at a time (termed all-to-all communication since there is often communication between each pair of the devices). In the backward pass, the accumulated gradients will be similarly sent back to the device that actually holds the table. In the above example, device 1 will calculate the gradient and send the gradient tensor back to device 2 to update the embedding table.

We summarize the overall training procedure of DLRM as follows. In the forward pass, each device samples its own mini-batch of data, which contains a batch of dense features and a batch of sparse features. The dense features will be simply processed by the duplicated MLP to obtain dense representations. The sparse features (i.e., feature values, or the indices of embedding tables) will be sent to the corresponding devices for embedding lookup. Then each device will perform the embedding lookup for the tables that are placed on the device (\textbf{forward computation}). The obtained vectors are sent back to the device that launches the query, which is essentially an all-to-all communication since each device will communicate with all the other devices (\textbf{forward communication}). The obtained vectors will be interacted with the dense vectors to obtain a final representation, followed by a prediction head to make the predictions. In the backward pass, the gradient will be passed backwardly from the prediction loss. Updating the dense part is straightforward since it is the same as the standard backpropagation. For the sparse counterpart, the gradient of each vector needs to be sent back to the device that actually holds the corresponding table, which leads to another all-to-all communication (\textbf{backward communication}). Then, the gradient will be applied to the embedding tables to update the embedding weights (\textbf{backward computation}). At the end of the backward pass, the weights of the duplicated MLP will be synchronized.

We only focus on optimizing the cost of the sparse part of the model, i.e., the cost of embedding tables, including forward computation, forward communication, backward communication, backward computation. This is because the costs of embedding computation and communication often dominate the overall training efficiency. For example, in our internal training pipeline of production recommendation models (which is already well optimized with numerous iterations), the cost of embedding tables account for 48\% and 65\% of the total computation and communication costs, respectively. Meanwhile, the embedding table cost is orthogonal to other costs, such as data loading, dense feature processing, etc. This means embedding table cost optimization can be considered as an independent task, which will contribute to the overall training throughput. We note that the embedding computation and communication can be performed simultaneously with the computation of the dense MLP. The bottleneck depends on which part takes more time. However, we observe in production models that embedding cost is often significantly larger than the dense MLP cost due to the extremely large embedding tables, which aligns with the observations from previous studies~\cite{acun2021understanding,zhao2020distributed}. Thus, the dense MLP cost is often ``hidden`` by the embedding table cost, and embedding table cost becomes the bottleneck during model training.

Optimizing the embedding table cost is very challenging because it has very complex computation and communication patterns. \textbf{First,} embedding computation or communication alone has very complex relationship with the embedding lookup patterns (we will provide detailed quantitative analysis of this in Section~\ref{appendix:A3}. \textbf{Second,} the forward/backward computation/communication costs can have interactive effects. For example, if the forward communication of a device is significantly larger than those of the other devices. Then the other devices need to wait until this device finishes the forward computation so that they can obtain the queried embedding vectors. Similarly, the backward computation for a device can only start after the device receives all the gradients in the backward communication.

As a result, different embedding table placements will significantly impact the embedding cost in several aspects. \textbf{First,} a good combination of embedding tables may lead to faster forward/backward computation since it may enable a more efficient kernel implementation. \textbf{Second,} balancing the forward computation time can reduce the waiting time before the forward communication starts. \textbf{Third,} balancing the backward computation time will also reduce the waiting time. \textbf{Forth,} balancing the number of amount of data being sent can accelerate the all-to-all communication.

However, optimizing embedding table placement is very challenging. This partition problem is known to be NP-hard\footnote{\url{https://en.wikipedia.org/wiki/Partition_problem}}, which means the number of possible placements grows exponentially with more tables. Additionally, due to the complexity of the embedding table cost discussed above, it is hard to optimize the placement in an analytical way. This motivates our work of DreamShard, which leverages RL to optimize the embedding table placement in a trial-and-error fashion.

\subsection{Embedding Table Features}
\label{appendix:A2}

We define several embedding table features to characterize embedding tables. These table features are highly correlated to the computation and communication workloads. Thus, in DreamShard, they serve as the input of the cost network and the policy network. In total, we use 21 features, which are defined as follows.

\begin{itemize}
    \item \textbf{Dimension (dim, 1 feature)}: It is the dimension of each embedding vector, i.e., the number of columns of the embedding table. It is a critical table feature since it determines the workloads of both computation and communication. For computation, the forward pass requires fetching the embedding vectors, and the backward pass will apply gradients to the embedding vectors, both of which have a computational complexity that increases linearly with the vector dimension. For communication, the vector dimension determines the data size, which will impact the communication time.
    \item \textbf{Hash size (1 feature)}: It is the number of embedding vectors in the embedding table, i.e., the number of rows of the table. It is called hash size because embedding lookup is essentially a hashing operation. While hashing is often believed to have $O(1)$ time complexity, which means the lookup time does not depend on the hash size, we find that hash size can still impact the lookup time because of caching mechanism. Specifically, modern GPUs often have L1/L2 caches, which are small but faster memories. If hash size is small, a larger portion of the embedding vectors can be put into the caches such that the lookup will be faster. In contrast, a large hash size will lead to a smaller portion of the embedding vectors being cached such that the lookup time will be larger.
    \item \textbf{Pooling factor (1 feature)}: It is the number of embedding indices in a lookup. For example, in YouTube video recommendation, a user may have watched multiple videos in the past. If a feature corresponds to ``all the videos that were watched in the past'', then we need to fetch all the embedding vectors that correspond to these video IDs from the table. In this context, pooling factor refers to the number of video IDs. Like dimension, pooling factor decides the workload of computation. In the forward pass, a larger pooling factor will result in more embedding vectors being fetched and summed, which will naturally lead to more computation. Similarly, in the backward pass, more computation will be required to update the embedding vectors with the gradients. Note that pooling factor usually will not impact communication since it does not decide the data size in communication. Since we often adopt mini-batch training, which means a batch of indices will be used to perform embedding lookup, we use the mean pooling factor as the table feature. Specifically, for a batch of indices, we calculate mean value of the pooling factors of all the training samples in the batch.
    \item \textbf{Table size (1 feature)}: Table size is the memory consumption of the embedding table in GBs. It can help the agent reason about satisfying the memory constraints of devices.
    \item \textbf{Distribution (17 feature)}: It refers to the accessing frequencies of all the indices of a table. Specifically, certain indices can be accessed far more frequently than other indices. Modern embedding table implementation will exploit such patterns with caching. The indices that are frequently accessed will tend to be put into the L1/L2 cache for acceleration. For a batch of $65,536$ indices, we use 17 bins, including $(0, 1]$, $(1, 2]$, $(2, 4]$, $(4, 8]$, $(8, 16]$, $(16, 32]$, $(32, 64]$, $(64, 128]$, $(128, 256]$, $(256, 512]$, $(512,$ $1024]$, $(1024, 2048]$, $(2048, 4096]$, $(4096, 8192]$, $(8192, 16384]$, $(16384, 32768]$, and $(32768, \infty)$. We count the number of appearances of each index and assign the count to the corresponding bin. Finally, we normalize the counts and make them sum to 1, which leads to a probability distribution with 17 table feature values.
\end{itemize}

\subsection{Quantitative Analysis of Computation and Communication Times}
\label{appendix:A3}

Embedding table placement is a very challenging problem because it is hard to estimate the costs without running the operations on GPUs. The main challenges include the non-linear relationship between the table cost and table features, operation fusion, and complex communication patterns. Here, we provide a quantitative analysis of these phenomena. All the results are collected using a modern embedding bag implementation from FBGEMM\footnote{\label{fbgem_link}\url{https://github.com/pytorch/FBGEMM/}}~\cite{fbgemm} from 2080Ti GPUs. Note that the results in Section~\ref{appendix:a31} and Section~\ref{appendix:a32} are originally collected in~\cite{zha2022autoshard}.

\subsubsection{Relationship Between Table Cost and Table Features}
\label{appendix:a31}

Recall that in Section~\ref{appendix:A2}, we have defined some table features, which can quantify the workloads of computation. However, due to the parallelism of GPUs, the actual table cost has a non-linear relationship with the table features. Here, we study the relationship between single-table cost and dimension, hash size, pooling factor, and distribution (table size is excluded because it can be essentially inferred from dimension and hash size). Dimension and hash size describe the table itself since they define the numbers of rows and columns of the table, respectively. The pooling factor and distribution characterize the indices assessing patterns, where the pooling factor measures the overall workload, and the distribution features measure the sparsity of the indices distributions. Now we analyze the above two types of features separately with synthetic embedding tables and indices.

\begin{figure}[h!]
  \centering
    \includegraphics[width=0.50\textwidth]{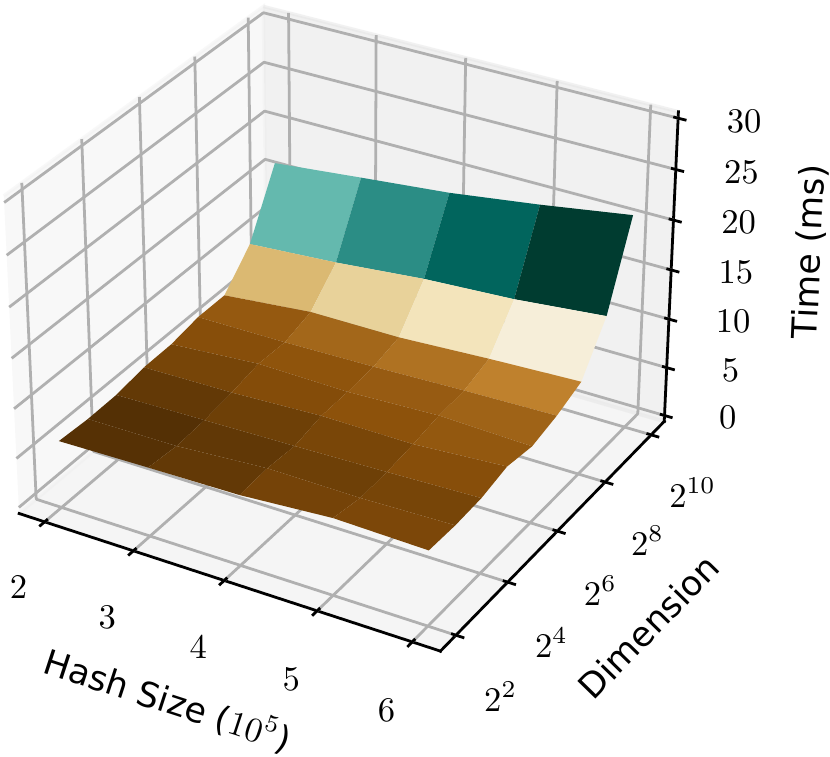}
  \caption{Influence of the hash size and the dimension on single-table cost. The results are originally collected from~\cite{zha2022autoshard}.}
  \label{fig:singleanalysis1}
\end{figure}

We study the impact of dimension and hash size with the pooling factor fixed as 32 and indices to be uniformly distributed. We vary the hash size from $2 \times 10^5$ to $6 \times 10^5$ and dimension from $2^2$ to $2^{10}$. We measure the kernel time (the sum of the forward and backward computation times) of the embedding operation for each of the combinations of hash size and dimension. The heat map of embedding cost is shown in Figure~\ref{fig:singleanalysis1}. We make three observations. \textbf{First,} a higher dimension will significantly increase the kernel time. This is expected since the embedding dimension corresponds to the size of the data to be fetched in the forward pass and the size of the data to be updated with the gradients in the backward pass. \textbf{Second,} while hash size only has a moderate impact on the table cost, a large hash size leads to a higher table cost. This also aligns with our intuition since a larger hash size will lead to a smaller portion of the indices being cached. \textbf{Third,} we find that the table cost has a non-linear relationship with both dimension and hash size.

\begin{figure}[h!]
  \centering
    \includegraphics[width=0.50\textwidth]{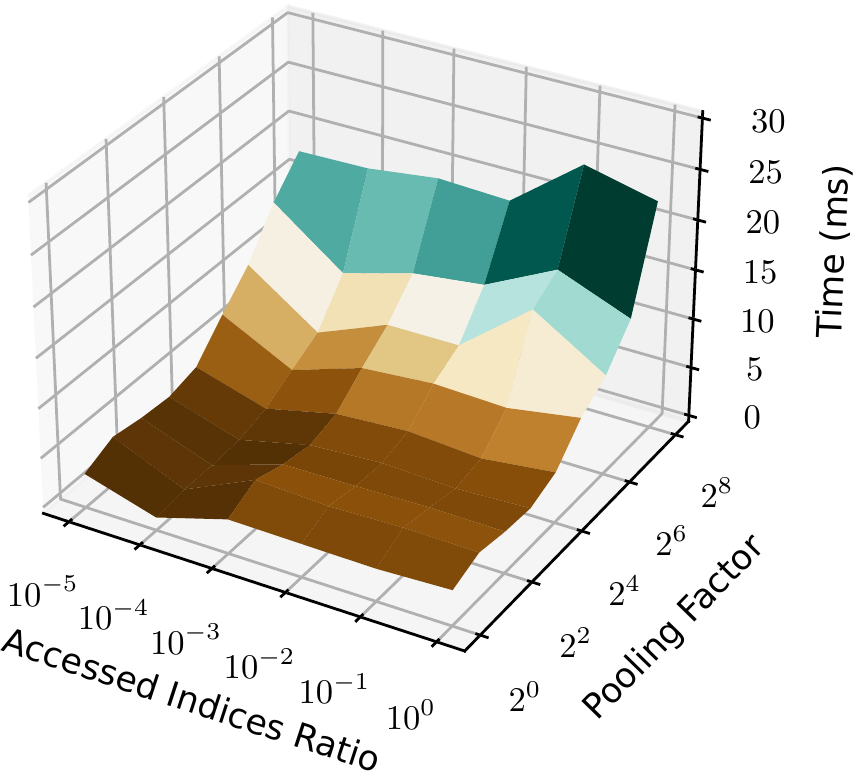}
  \caption{Influence of the indices distribution and the pooling factor  on single-table cost. The results are originally collected from~\cite{zha2022autoshard}.}
  \label{fig:singleanalysis2}
\end{figure}

Next, we study the impact of pooling factor and indices distributions with the hash size fixed as $10^6$ and dimension fixed as $32$. We vary the mean pooling factor from $2^0$ to $2^8$. For the indices distribution, some indices could be accessed far more frequently than others~\cite{acun2021understanding}. We simulate this phenomenon in our synthetic indices by only allowing a subset of all the embedding vectors to be accessed. Specifically, we define \emph{accessed indices ratio} as the ratio of the embedding vectors that can be accessed in the embedding table. For example, a ratio of 1.0 suggests the indices are uniformly distributed. A ratio of $10^{-3}$ means only $0.1\%$ of all the embedding vectors can be accessed. This means that those $0.1\%$ of embedding vectors are ``warm'' vectors and can be accelerated with caching. Note that there can be many ways to simulate the indices distributions. Here, we only focus on the most simple one, which masks a subset of the embedding vectors. The impacts of the pooling factor and accessed indices ratio are illustrated in Figure~\ref{fig:singleanalysis2}. We make three observations. \textbf{First,} a larger pooling factor will significantly increase the table cost. This is because a larger pooling factor suggests more computation cost of fetching and updating the embedding vectors. \textbf{Second,} sparser indices distribution tends to have lower table cost, which could be explained by the caching mechanism. \textbf{Third,} the table cost has a complex and non-linear relationship with pooling factor and indices distributions.

In the above analysis, we separately studied two table features with the other features fixed. However, it is possible that these features have an interaction effect, which could make the table cost even more challenging to estimate. Thus, when developing DreamShard, we are motivated to use a cost network to directly predict the table cost in a data-driven manner.

\subsubsection{Analysis of Operation Fusion}
\label{appendix:a32}

\begin{figure}[h!]
  \centering
    \includegraphics[width=0.50\textwidth]{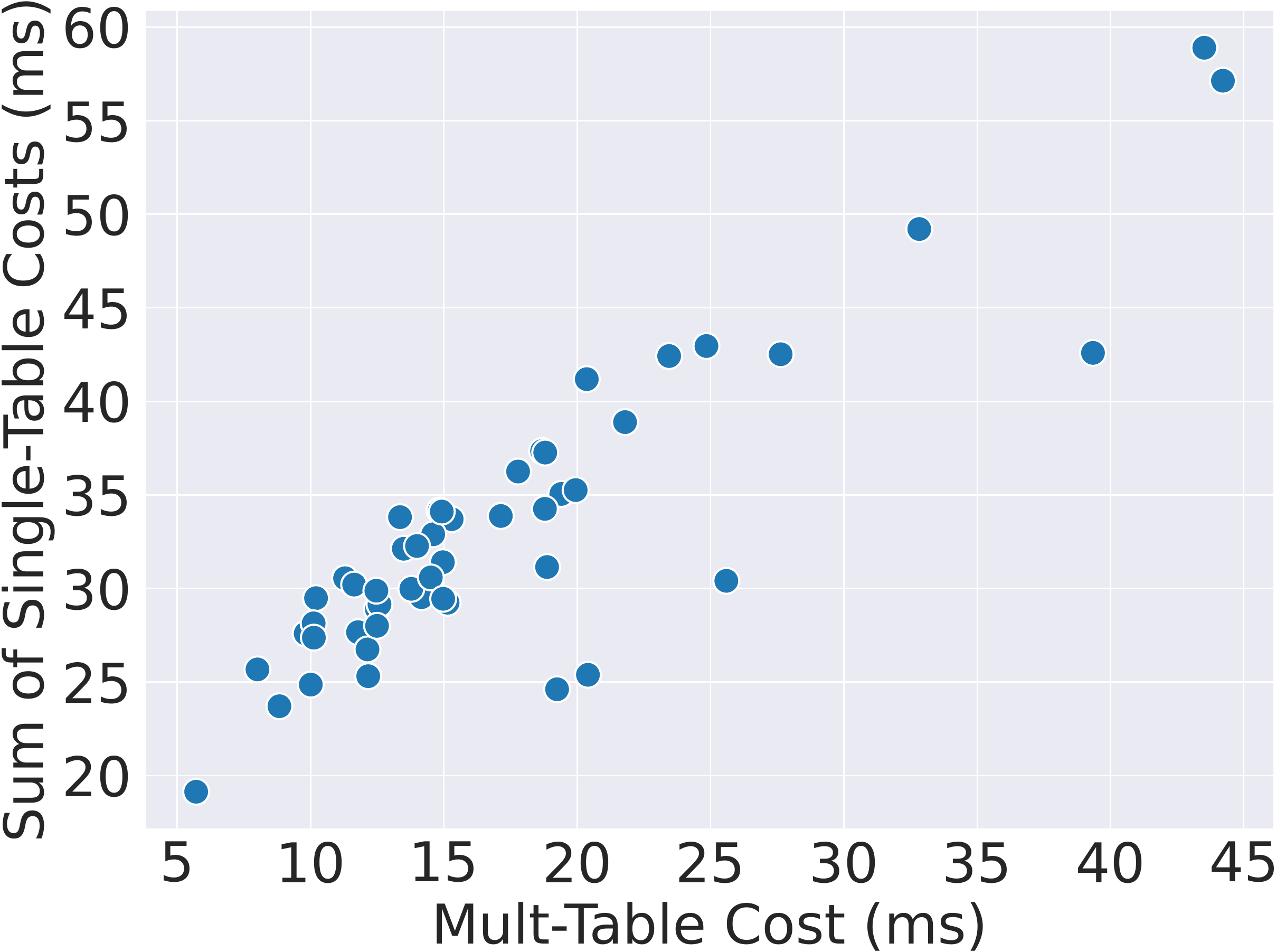}
  \caption{Mult-table cost = the sum of single-table costs? Mult-table cost is smaller than the sum of single-table costs, where the speedup ranges from 1X to 3X. They do not have a clear linear correlation. Assuming that the multi-table cost has a linear correlation with the sum of single-table costs and the single table cost estimation is perfect, we grid-search the correlation efficiency in the range of $[1.0, 2.0]$ with a step size of 0.001. The best MSE we found is 77.97, which is significantly worse than our cost network (our cost network can have less than 1.0 MSE). The results are originally collected from~\cite{zha2022autoshard}.}
  \label{fig:multitableanalysis}
\end{figure}

Operation fusion~\cite{niu2021dnnfusion} is a common acceleration strategy that uses a single operation to subsume the computation performed by multiple operations. It is particularly effective for embedding tables due to batching. It can often lead to significant speedup in operation computation time. Unfortunately, the operation fusion also makes the multi-table costs hard to predict. Here, we analyze the operation fusion by randomly sampling 10 tables from the DLRM dataset and comparing its multi-table cost and the sum of the single-table costs. We consider the sum of the single-table costs as the baseline because it represents the case without any acceleration. We repeat the sampling 50 times and plot the results in Figure~\ref{fig:multitableanalysis}.

We make two observations as follows. \textbf{First,} the multi-table cost is significantly lower than that of single-table cost. This is expected since operation fusion can accelerate the operation. The results show that operation fusion can lead to roughly 1.5X speedup when we have 10 tables. \textbf{Second,} while the multi-table cost is in general positively correlated with the sum of single-table costs, they are not linearly correlated. Specifically, the actual speedup is case-by-case, which may depend on many factors. The results suggest that simply using the sum of single-table costs to estimate the multi-table cost is inaccurate.

In our embedding placement process, we inevitably need to estimate the multi-table costs. Unfortunately, the above analysis suggests that we may not be able to get an accurate estimation without actually running the multi-table operations on GPUs. This motivates us to develop a neural cost network to directly approximate the multi-table costs.

\subsubsection{Analysis of Communication}
\label{appendix:a33}
Embedding tables in recommendation models have very complex communication patterns because of combined model-parallelization and data-parallelization. Specifically, we require all-to-all communication to send the embedding vectors or gradients from device to device. Since there are often limited bandwidths among GPUs, if the data are not distributed in a balanced way, then it may take significantly more time for communication. Here, we analyze the communication costs with different degrees of balance.

The communication cost depends on the amount of data to be sent to each device. In the context of embedding table placement, we mainly need to send the summed embedding vectors, whose sizes are determined by the batch size and the table dimension. Since batch size is pre-determined, the communication cost is essentially decided by the sum of table dimensions in each GPU device. Thus, in our empirical analysis, we adjust the sums of table dimensions for the GPUs to simulate different levels of imbalance. Specifically, we fix the batch size to be $65,536$ and construct 16 embedding tables, where each table has a dimension of $64$. Then we place these tables on the GPU devices to simulate different degrees of balance. 

\begin{table}[h!]
\centering
\caption{Communication time of 4 GPUs with different sums of table dimensions (the tables have $1,024$ dimensions in total) in each device with a batch size of $65,536$. We highlight the max communication time across devices since the slowest device becomes the bottleneck.}
\label{tbl:communicationanalysis}
\scriptsize
\setlength{\tabcolsep}{4.0pt}

\begin{tabular}{c|c|c|c|c|c|c|c|c|c}
\toprule
\multirow{2}{*}{Category} & \multicolumn{4}{c|}{Sum of Table dimensions} & \multicolumn{4}{c|}{Communication cost} & \multirow{2}{*}{Max cost} \\ 
\cline{2-9}
~ & \multicolumn{1}{c|}{GPU 1} &  \multicolumn{1}{c|}{GPU 2} &  \multicolumn{1}{c|}{GPU 3} &  \multicolumn{1}{c|}{GPU 4} & \multicolumn{1}{c|}{GPU 1} & \multicolumn{1}{c|}{GPU 2} &  \multicolumn{1}{c|}{GPU 3} &  \multicolumn{1}{c|}{GPU 4} & ~ \\
\midrule

Perfectly Balanced & 256 & 256 & 256 & 256 & 11.24$\pm$0.17 & 11.15$\pm$0.12 & 11.08$\pm$0.17 & 11.08$\pm$0.17 & 11.24$\pm$0.17 \\

\midrule

\multirow{4}{*}{Slightly Imbalanced} & 192 & 256 & 320 & 384 & 13.20$\pm$0.16 & 13.88$\pm$0.04 & 14.18$\pm$0.04 & 14.15$\pm$0.06 & 14.15$\pm$0.06 \\
~ & 192 & 192 & 320 & 320 & 11.74$\pm$0.08 & 13.01$\pm$0.13 & 12.89$\pm$0.10 & 12.93$\pm$0.11 & 13.01$\pm$0.13 \\
~ & 128 & 192 & 320 & 384 & 12.28$\pm$0.06 & 13.82$\pm$0.08 & 14.03$\pm$0.08 & 14.02$\pm$0.07 & 14.03$\pm$0.08 \\
~ & 128 & 128 & 384 & 384 & 12.02$\pm$0.10 & 14.67$\pm$0.09 & 14.73$\pm$0.08 & 14.47$\pm$0.11 & 14.73$\pm$0.08 \\

\midrule

\multirow{4}{*}{Very Imbalanced} & 64 & 128 & 384 & 448 & 12.91$\pm$0.80 & 16.00$\pm$0.81 & 16.11$\pm$0.81 & 15.82$\pm$0.08 & 16.11$\pm$0.81 \\
~ & 64 & 64 & 448 & 448 & 12.50$\pm$0.05 & 16.65$\pm$0.06 & 16.67$\pm$0.08 & 16.29$\pm$0.06 & 16.67$\pm$0.08 \\
~ & 64 & 64 & 320 & 576 & 12.56$\pm$0.13 & 15.61$\pm$0.15 & 16.93$\pm$0.17 & 16.89$\pm$0.11 & 16.93$\pm$0.17 \\
~ & 64 & 64 & 64 & 832 & 13.01$\pm$0.14 & 12.96$\pm$0.15 & 17.65$\pm$0.21 & 17.65$\pm$0.22 & 17.65$\pm$0.21 \\

\bottomrule
\end{tabular}
\end{table}

Table~\ref{tbl:communicationanalysis} presents the communication costs under different degrees of balance using 4 GPUs. We can see that when the sums of table dimensions become more imbalanced, the communication cost also increases. Thus, we need to balance the table dimensions to minimize communication costs. However, balancing dimension alone cannot achieve an overall good result, since a placement with a balanced dimension may still be not balanced in computation. To tackle this challenge, DreamShard jointly optimizes both computation and communication in a data-driven manner with RL.

\subsection{Why Backward Communication Time but not Forward Communication Time? }
\label{appendix:A4}

Forward communication and backward communication will send data of the same size in all-to-all communication (but in different directions). Specifically, in the forward pass, the obtained sparse representations will be sent, while in the backward pass, the gradient of the sparse representations will be sent back. Recall that the cost network in DreamShard only predicts the backward communication cost instead of the forward communication cost. One may ask why they are different given that the amount of data is the same.

We do not predict forward communication because we find that a considerable portion of the forward communication time is not spent on communication, but instead on idle time waiting for other devices. For example, if a device finishes forward computation very quickly, then it has to wait for other devices to finish computation before it can start communication. However, such waiting time is also counted in the forward communication with PyTorch even though it does not communicate. On the contrary, the devices will be ``synced'' when the forward communication is finished, so that the backward communication often does not have idle time. Thus, we only predict backward communication, which can better reflect the true communication cost.

\section{Implementation Details}
\label{appendix:B}
In this section, we introduce the implementation details of DreamShard. We will first introduce the neural architectures of the cost network and the policy network. Then we provide more details of the training and inference procedures. Further, we summarize the hyperparameter configurations. Finally, we describe the hardware and software used in our experiments. \textbf{To ensure reproducibility, we will open-source our code}.

\subsection{Neural Architecture of cost network}
\label{appendix:B1}

The cost network consists of three sub-networks, including 1) a shared feature extraction MLP (denoted as $\text{MLP}_{\text{table}}$), which maps the 21 table features to latent representations, 2) backward/communication/forward heads, which predict cost features based on the device representation, and 3) an overall cost head, which takes the final representation for all the devices as input and predicts the overall cost. 

We provide the detailed procedure of a forward pass as follows. For a raw state $s_t$, we first use an MLP to process all the raw table features with $\mathbf{h}^{\text{table}}_i = \text{MLP}_{\text{table}}(\mathbf{e}_i) \in \mathbb{R}^l$, where $l$ is the hidden dimension, and $\text{MLP}_{\text{table}}$ is shared across all the tables. This leads to a set of hidden representations for each device $\{\mathbf{h}^{\text{table}}_i | i \in \mathcal{P}_d\}$. Then we obtain the device representation with element-wise sum by $\mathbf{h}^{\text{device}}_d = \sum_{i \in \mathcal{P}_d} \mathbf{h}^{\text{table}}_i \in \mathbb{R}^l$, which has a fixed dimension regardless of the number of tables in the device. The motivation for using element-wise sum is that $\mathbf{h}^{\text{table}}_i$ is expected to describe the computational cost patterns of a table so it is natural to accumulate $\mathbf{h}^{\text{table}}_i$ to represent the potentially accumulated computational costs when we have multiple tables. Then $\mathbf{h}^{\text{device}}_d$ serves as the input of the backward/communication/forward heads to predict the cost features $\mathbf{q}_{t,d}$. To predict the overall cost, we similarly obtain a fixed-dimension representation for all the devices $\mathbf{h}$ by applying an element-wise max to $\mathbf{h}^{\text{device}}_d$, i.e., $\mathbf{h}$ is defined by $h_k := \max_{1 \le d \le D} h^{\text{device}}_{d,k}$, where $h_k$ and $h^{\text{device}}_{d,k}$ denote the $k^{th}$ element of $\mathbf{h}$, and $\mathbf{h}^{\text{device}}_d$, respectively. The motivation of element-wise max is that the slowest device is usually the bottleneck of the overall cost. Then $\mathbf{h}$ is followed by an overall cost head to predict the reward

We elaborate on the neural architectures of the three sub-networks as follows. 

\begin{itemize}
    \item \textbf{Shared table feature extraction MLP:} The dimension of the latent representation is set to be $32$. We instantiate the shared feature extraction MLP with a 2-layer neural network with a size of 21-128-32.
    \item \textbf{Backward/communication/forward heads:} We use three MLPs to implement these three heads. Each MLP is a 2-layer neural network with a size of 32-64-1.
    \item \textbf{Overall cost head:} Similarly, we instantiate it with a 2-layer neural network with a size of 32-64-1.
\end{itemize}

For all the above three sub-networks, we use the ReLU activation function and the default parameter initialization in PyTorch.

\subsection{Neural Architecture of Policy Network}
\label{appendix:B2}

The policy network consists of three sub-networks, including 1) a shared feature extraction MLP that is independent of that of the cost network (denoted as $\widetilde{\text{MLP}}_{\text{table}}$), which maps the 21 table features to latent representations, 2) cost feature MLP, which processes the cost features by mapping them to latent representations, and 3) a policy head, which maps device representations to probability distributions.

We provide the detailed procedure of a forward pass as follows. \textbf{First,} following the cost network, another MLP is used to process the raw table features $s_t$ with $\widetilde{\mathbf{h}}^{\text{table}}_i = \widetilde{\text{MLP}}_{\text{table}}(\mathbf{e}_i) \in \mathbb{R}^l$, where $\widetilde{\text{MLP}}_{\text{table}}$ is shared across all the tables (but independent of $\text{MLP}_{\text{table}}$ in $f_{\text{cost}}$). A fixed-dimension device representation for each device can be similarly obtained with element-wise sum by $\widetilde{\mathbf{h}}^{\text{device}}_d = \sum_{i \in \mathcal{P}_d} \widetilde{\mathbf{h}}^{\text{table}}_i \in \mathbb{R}^l$. \textbf{Second,} we augment $\widetilde{\mathbf{h}}^{\text{device}}_d$ with the cost features $\mathbf{q}_{t,d}$. Specifically, we use an MLP to process $\mathbf{q}_{t,d}$ by $\mathbf{h}^{\text{cost}}_d = \text{MLP}_{\text{cost}}(\mathbf{q}_{t,d}) \in \mathbb{R}^l$, where $\text{MLP}_{\text{cost}}$ is shared across the cost features of all the devices. The augmented device representation is the concatenation of $\widetilde{\mathbf{h}}^{\text{device}}_d$ and $\mathbf{h}^{\text{cost}}_d$, denoted as $[\widetilde{\mathbf{h}}^{\text{device}}_d; \mathbf{h}^{\text{cost}}_d]$. \textbf{Third,} we use a shared policy head to process the augmented device representation, followed by a Softmax layer to produce action probabilities. Let $\text{MLP}_{\text{policy}}$ be the policy head. The probabilities for all the legal actions are obtained by $\mathbf{p} = \text{Softmax}\{\text{MLP}_{\text{policy}}[\widetilde{\mathbf{h}}^{\text{device}}_d; \mathbf{h}^{\text{cost}}_d] | d \in \mathcal{A}_t\}$, where $\text{MLP}_{\text{policy}}$ is shared across all the devices. \textbf{Finally,} we sample an action $a_t$ based on the action probabilities $\mathbf{p}$. Our design allows $\pi$ to be trained on one task and generalize to other tasks with different numbers of tables and/or devices.
 
We elaborate on the neural architectures of three sub-networks as follows.

\begin{itemize}
    \item \textbf{Shared table feature extraction MLP:} It has the same architecture like that of the cost network (but the weights are not shared). The dimension of the latent representation is set to be $32$. We instantiate the shared feature extraction MLP with a 2-layer neural network with a size of 21-128-32.
    \item \textbf{Cost feature MLP:} This network maps the three cost features into a 32-dimension cost representation. We instantiate it with an MLP of size 3-64-32.
    \item \textbf{Policy head:} It takes as input the concatenated representation of the device representation and the cost representation. Thus, its input size is 64 (32 for the device representation and 32 for the cost representation). Then we use a 1-layer MLP of size 64-1 to map the representations to a ``confidence score''. After obtaining the score for each device, we use a Softmax layer to produce the action probabilities, i.e., the probability of selecting each of the devices. 
\end{itemize}

For all the above three sub-networks, we use the ReLU activation function and the default parameter initialization in PyTorch.

\subsection{Comparison of Different Reductions}
\label{appendix:B3}
Recall that we use the element-wise sum to aggregate table representations in a device and element-wise max to aggregate device representations. Here, we justify our design choices by comparing with other reduction methods. Specifically, we randomly sample 10,000 cost data points from the DLRM-50 (4) dataset. Then we use 20\% for testing and vary the size of the training data to compare the performances of different reductions using different numbers of data points. For all the experiments, we use a batch size of 64, an Adam optimizer with a learning rate of 0.0005, and we train 50,000 batches. We report the sum of testing MSE for all the predicted costs. All the experiments are repeated 5 times, and the mean and standard deviation are reported.

\begin{figure}[h!]
  \centering
    \includegraphics[width=0.50\textwidth]{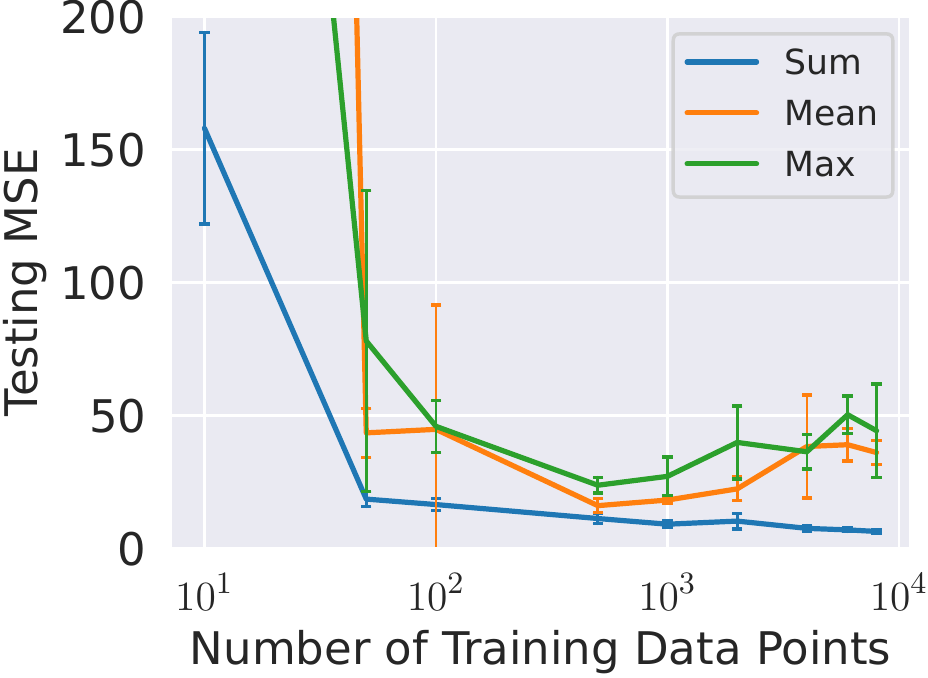}
  \caption{Comparison of different reductions for table representations (max reduction is applied to device representations).}
  \label{fig:tablereduction}
\end{figure}

\begin{figure}[h!]
  \centering
    \includegraphics[width=0.50\textwidth]{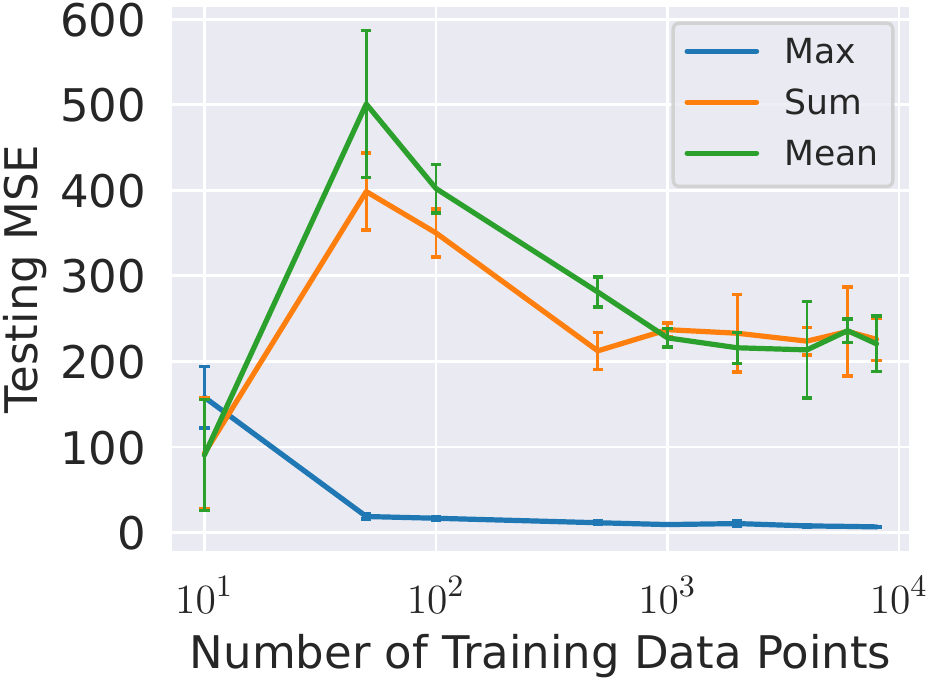}
  \caption{Comparison of different reductions for device representations (sum reduction is applied to table representations).}
  \label{fig:devicereduction}
\end{figure}

In the first experiment, we try max and mean reductions for the table representations and use max reduction for the device representations . The results are reported in Figure~\ref{fig:tablereduction}. We observe that sum reduction is the best choice for table representations. In the second experiment, we try sum and mean reductions for the device representations and use sum reduction for the table representations. The results are reported in Figure~\ref{fig:devicereduction}. We observe that max reduction is the best choice for table representations. Thus, in DreamShard we use sum reduction for the table representations and max reduction for the device representations.

\subsection{Details of DreamShard Training and Inference}
\label{appendix:B4}

In this subsection, we elaborate on the training and inference procedures. We will first present the loss functions for updating the cost network and the policy network. Then we summarize the training procedure. Finally, we describe how DreamShard performs inference on unseen embedding table placement tasks.

\subsubsection{Loss Functions}
\label{appendix:B41}

We update the cost network with mean squared error (MSE). Recall that we use a buffer to collect cost data by using the current policy $\pi$ to interact with the environment (we will elaborate on this procedure in Section~\ref{appendix:B42}). Suppose we have already collected some cost data. Then we use the cost data to update the cost network. Specifically, the cost network maps the raw state $s_t$ into the predicted cost features $\{\mathbf{\hat{q}}_{t,d}\}^D_{d=1}$ and the predicted overall cost $\hat{c}$. Let $\{\mathbf{q}_{t,d}\}^D_{d=1}$ and $c(\mathbf{a})$ be the ground truth of the cost features and the overall cost, respectively. We use the sum of their mean squared errors (MSE) to update the cost network:
\begin{equation}
    L_{\text{cost}} = \sum_{d=1}^D \text{MSE}(\mathbf{\hat{q}}_{t,d} - \mathbf{q}_{t,d}) + \text{MSE}(\hat{c}, c(\mathbf{a})),
    \label{eq:1}
\end{equation}
where $\text{MSE}(\cdot, \cdot)$ represents the MSE loss. Note that it is possible to use a weighted loss to prioritize the prediction of the cost features or the overall cost by introducing additional hyperparameters. We will explore this possibility in our future work.

We update the policy network $\pi$ with the standard policy gradient loss~\cite{williams1992simple} enhanced by a baseline to reduce the variance, and an entropy loss to enhance exploration:
\begin{equation}
    L_{\text{RL}} = \sum_{t=0}^{M-1} \log \pi(a_i|s_i) (\sum_{j=i}^{M} r_j - b_j) + w_{\text{entropy}} \sum_{a=1}^D \pi(a|s_i) \log \pi(a|s_i),
    \label{eq:2}
\end{equation}
where $\pi(a|s)$ is the predicted probability probability of performing action $a$ in state $s$, $w_{\text{entropy}}$ is the weight of the entropy. $b_j$ is the reward obtained at step $t$; $r_j$ is the negative of the overall cost when $j=M$, and $r_j$ is $0$ for all the other steps. Thus, $\sum^{M}_{j=i}r_j$ essentially reduces to $r_M$ (we use $\sum^{M}_{j=i}r_j$ so that it is consistent with the formulas used in the RL literature)\footnote{In the RL literature, a discount factor is often applied to make the early decisions have a smaller reward. In our context, we simply set the discount factor to be $1$ (i.e., no discount) because the reward in the MDP is sparse and the early decisions are very important.}.  $b_j$ is a baseline to reduce variance and stabilize training. In each update step, we run $N_{\text{episode}}$ episodes at a time and use their mean reward as $b_j$. Then we update the policy $\pi$ by calculating the loss with a batch of episodes. The policy network $\pi$ can be updated with the loss using the standard backpropagation.

\subsubsection{Training Procedure}
\label{appendix:B42}

The training procedure is iterative. In each iteration, we sequentially do the following: 1) use the current policy $\pi$ to sample some placements, collect the costs from GPUs, and store the collected cost data into the buffer, 2) update the cost network with the cost data collected in the buffer, and 3) update the RL agent by interacting with the estimated MDP simulated by the cost network. We provide details for each of the three stages below.

\textbf{Data collection.} In this stage, we use the current policy $\pi$ to generate table placements and evaluate the placements on GPUs. Specifically, we first randomly select a training task from $\mathcal{T}_{\text{train}}$. Then we generate a placement for this task by interacting with the estimated MDP with $\pi$. Before starting an episode, we first sort the tables in descending order based on the single-table cost, which is predicted using the cost network. The motivation is that it will be more likely to achieve a good balance if we put the costly table at the beginning of the MDP. Then we follow the MDP to place the tables one by one, where in each step, we obtain the augmented state using the current cost network, and then feed the augmented state to the policy $\pi$ to predict the action probabilities. Then we sample an action based on the action probabilities to make the placement decision. After generating a placement, we evaluate the placement on GPUs to collect the computation and communication costs using PARAM Benchmark~\footnote{\url{https://github.com/facebookresearch/param}}, which is the official micro-benchmarking tool for PyTorch. To precisely measure the cost, the benchmarking consists of three steps: 1) the initialization step will initialize the operators with the specified embedding table arguments and load the indices data to the GPU, 2) the warmup step will run all the computation and communication for 5 times to allow CUDA to complete the necessary preparations, and 3) the benchmarking step will run all the computations and communications again for 10 times. The median latency in the benchmarking step will be returned since the median value is less sensitive to outliers. The returned latency will be stored in a buffer for training the cost network later. We find that the above benchmarking strategy is very stable, and the obtained latency has very low variance. There is one hyperparameter in this stage, i.e., $N_{\text{collect}}$, which specifies the number of placements to be generated.

\textbf{Training the cost network.} In this stage, we sample multiple mini-batches of cost data from the buffer to update the cost network. Specifically, in each update step, we sample a batch of cost data with a size of $N_{\text{batch}}$. Then we feed the data to the cost network and update it based on the loss in Eq.~\ref{eq:1}. We update it for $N_{\text{cost}}$ times. $N_{\text{batch}}$ and $N_{\text{cost}}$ are hyperparameters controlling the update of the cost network.

\textbf{Training the policy network.} We use the current policy $\pi$ to interact with the estimated MDP, which is akin to the data collection stage. The only difference is that we do not evaluate the generated placement on GPUs. Instead, the final reward is simply obtained by a forward pass of the cost network. The design of the estimated MDP can significantly improve the training efficiency of RL since it isolates the RL training from the evaluation on GPUs. In each update step of RL, we first randomly select a training task. Then we generate $N_{\text{episode}}$ episodes through interacting with the estimated MDP. Next, we update the policy network $\pi$ based on Eq.~\ref{eq:2}. We repeat the above procedure $N_{\text{RL}}$ times.

\subsubsection{Inference Procedure}
\label{appendix:B43}
The inference of DreamShard is straightforward. The procedure is similar to the data collection except that we choose the action with the highest predicted probability instead of sampling an action based on the probabilities. This is because in training, we require the agent to explore different actions and discover the best strategy. Whereas, during inference, we no longer need exploration. As such, we can simply choose the most confident action. We summarize the procedure of performing inference on testing tasks in Algorithm~\ref{alg:2}. We note that the inference does not require GPUs.

\begin{algorithm}[h!]
\caption{Inference of DreamShard}
\label{alg:2}
\setlength{\intextsep}{0pt} 
\begin{algorithmic}[1]
\STATE \textbf{Input:} Some testing tasks $\mathcal{T}_{\text{test}}$, the trained cost network, trained policy network
\FOR{each task in $\mathcal{T}_{\text{test}}$}
    \FOR{step = 1, 2, ... until episode ends}
        \STATE Construct the augmented state using the cost network
        \STATE Predict the action probability using the policy network
        \STATE Take and record the action that has the highest probability
    \ENDFOR
    \STATE Record the action sequence (placement)
\ENDFOR
\end{algorithmic}
\end{algorithm}

\subsection{Hyperparameter Configuration}
\label{appendix:B45}
We summarize all the hyperparameters of DreamShard below.
\begin{itemize}
    \item \textbf{Data collection:} We set $N_{\text{collect}} = 10$.
    \item \textbf{cost network training:} We set $N_\text{cost} = 300$, and $N_\text{batch} = 64$.
    \item \textbf{Policy network training:} We set $N_{\text{RL}} = 10$, $N_{\text{episode}} = 10$, and the entropy weight $w_{\text{entropy}} = 0.001$.
    \item \textbf{Optimizer:} For both the cost prediction and policy networks, we adopt Adam optimizer with an initial learning rate of $0.0005$, with the other hyperparameters as default. A linear scheduler is used to linearly decay the learning rate to zero throughout the training process.
    \item \textbf{Embedding operation:} We use the embedding bag implementation in FBGEM$^{\ref{fbgem_link}}$~\cite{fbgemm}. For the parameters of embedding tables, we randomly initialize them with fp16 precision.
    
\end{itemize}

\subsection{Hardware and Software Description}
\label{appendix:B6}
For the DLRM dataset, all the experiments are conducted on a server with 48 Intel(R) Xeon(R) Silver 4116 CPU @ 2.10GHz processors, 188 GB memory, and four NVIDIA GeForce RTX 2080 Ti GPUs. For the Prod dataset, the server has similar hardware configurations but with NVIDIA V100 GPUs to accommodate the larger sizes of the tables. For software, we use Python 3.8.4, and PyTorch 1.9.1.

\section{Details of the Datasets}
\label{appendix:C}

We note that our goal is not to evaluate the accuracy of a recommendation model, but rather the training efficiency of embedding tables. The public recommendation datasets are not suitable for evaluation since they cannot match the scale of real-world industrial models. They are often too small with very few categorical features so the latency of embedding operations will always be very small no matter how the embedding tables are placed.

\begin{table}[h!]
\centering
\caption{Comparison of embedding table feature statistics between some popular public recommendation datasets and the industrial-scale DLRM dataset. Our production data has an even larger scale than the DLRM dataset (details are not shown due to data privacy). The public datasets are not suitable to evaluate embedding table placement algorithms. They only have very few small tables, and the average pooling factor is only 1, which suggests there is only one feature in each table per instance when performing embedding lookup.}
\label{tbl:publicvsindustrial}
\scriptsize

\begin{tabular}{l|l|c|c|c}
\toprule
\multicolumn{2}{c|}{Dataset} & \# of Tables & Avg. hash size  & Avg. pooling factor \\
\midrule
\multirow{3}{*}{Public} & Criteo\footnote{\url{https://www.kaggle.com/c/criteo-display-ad-challenge}} & 26 & 17,839 & 1 \\
~ & Avazu\footnote{\url{https://www.kaggle.com/c/avazu-ctr-prediction/data}} & 23 & 67,152 & 1 \\
~ & KDD\footnote{\url{https://www.kaggle.com/c/kddcup2012-track2/data}} & 10 & 601,908 & 1 \\

\midrule
\multirow{1}{*}{Industrial-Scale} & DLRM & 856 & 4,107,458 & 15 \\
%~ & Our production  &  &  &  \\
\bottomrule
\end{tabular}
\end{table}

Fortunately, Meta recently released the DLRM$^{\ref{dlrm_dataset}}$ dataset, which is a synthetic dataset that shares memory access reuse patterns similar to those arising in Meta production recommendation workloads. This dataset is an ideal benchmark to evaluate embedding table placement algorithms because it can well simulate the real workloads under different table placements in industrial models, and the results obtained on it will be reproducible since the dataset is open-sourced. Table~\ref{tbl:publicvsindustrial} compares the scales of some large-scale public recommendation datasets and the DLRM dataset. We can observe a clear gap between the public datasets and the DLRM dataset. The DLRM dataset has around one order of magnitude more tables, average hash size (i.e., the number of rows of the table), and average pooling factor (i.e., the number of rows extracted in a table for one instance when performing lookup). In what follows, we introduce and visualize the DLRM dataset. We will not provide more details of our production dataset due to data privacy.

\subsection{Data Format of the DLRM Dataset}
The DLRM dataset is stored as three PyTorch tensors, which are pickled in a single file. The three tensors include an indices tensor, an offsets tensor, and a length tensor. For brevity, we denote them as \texttt{indices}, \texttt{offsets}, and \texttt{lengths}, respectively. \texttt{indices} is a vector, where each element is an integer. The indices are ordered by the keys of \texttt{(table\_id, batch\_offset)}. For example, the first batch of indices (the size is determined by the offset) is for the first table, and the second batch of indices (the size is determined by another offset) is for the second table, etc. \texttt{offsets} is also a vector. It indicates the starting position and the ending position of \texttt{indices} for one lookup. It is also ordered by \texttt{(table\_id, batch\_offset)}. For instance, suppose the batch size is 45. Then \texttt{offsets[45]} and \texttt{offsets[46]} specify the starting and ending positions of the 45$^\text{th}$ indices lookup in the first table. The slice between the starting and ending positions, i.e., \texttt{indices[offsets[45]:offsets[46]]} corresponds to the 45$^\text{th}$ instance in the batch for the first table. \texttt{lengths} is a matrix and is of the shape of \texttt{[num\_tables, batch\_size]}, where each element is the pooling factor of the corresponding indices lookup. \texttt{lengths} is provided for correctness validation purposes.

\subsection{Data Visualization of the DLRM dataset}

We visualize the 856 tables in the DLRM dataset. Specifically, we focus on the distributions of hash size, mean pooling factor, and the relation between the hash size and pooling factor. We also visualize the distribution of indices accessing frequency since it may impact the caching mechanism. Note that all the results are originally collected in~\cite{zha2022autoshard}.

\begin{figure}[h!]
  \centering
    \includegraphics[width=0.5\textwidth]{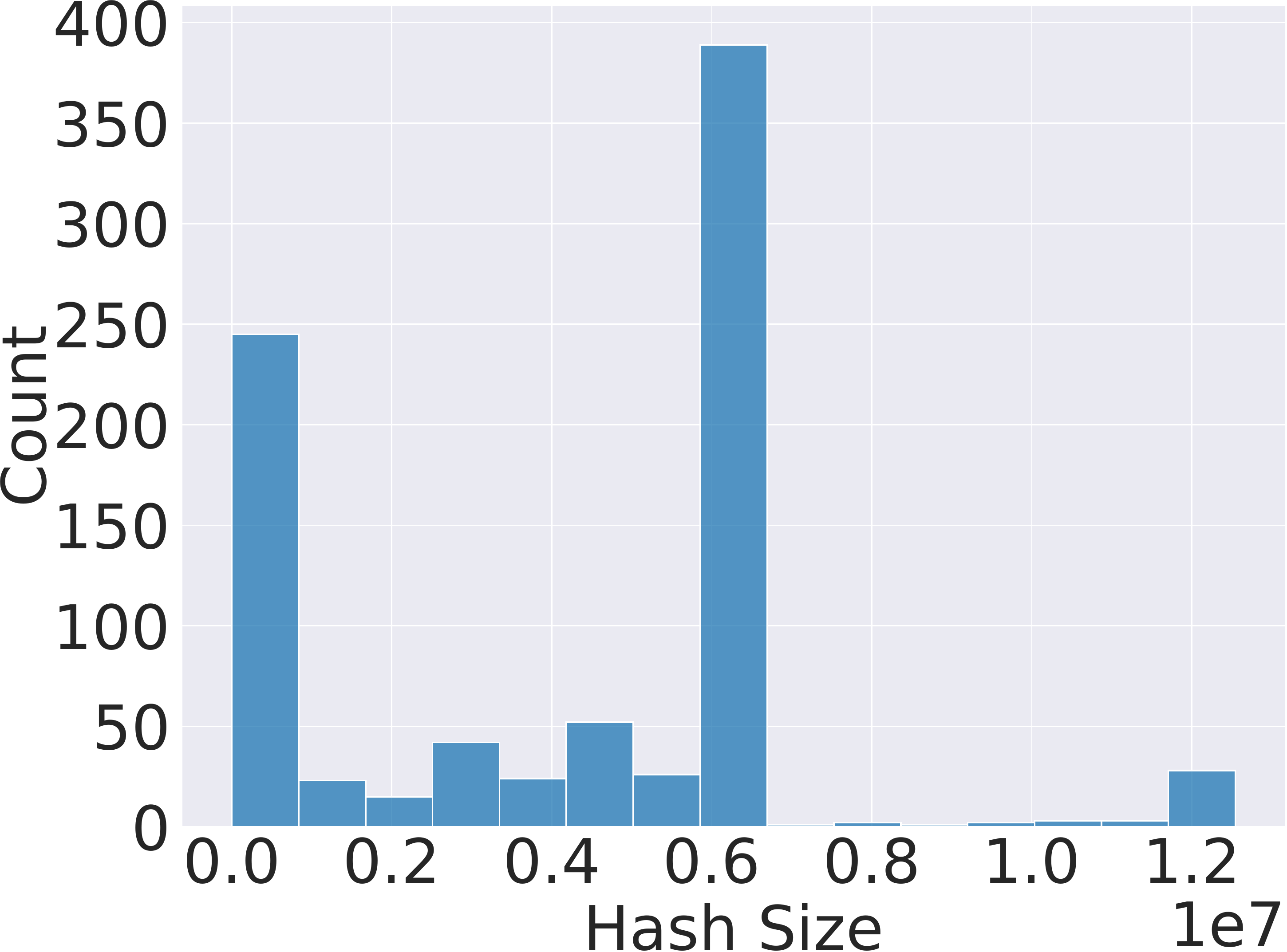}
  \caption{Hash size distribution.}
  \label{fig:distributionhashsize}
\end{figure}

Figure~\ref{fig:distributionhashsize} visualizes the distribution of hash size. We observe that the hash sizes for most tables are around $10^6$, while some can reach $10^7$. The tables with large hash sizes could lead to very large tables, making it challenging to balance the size of the tables. 

\begin{figure}[h!]
  \centering
    \includegraphics[width=0.5\textwidth]{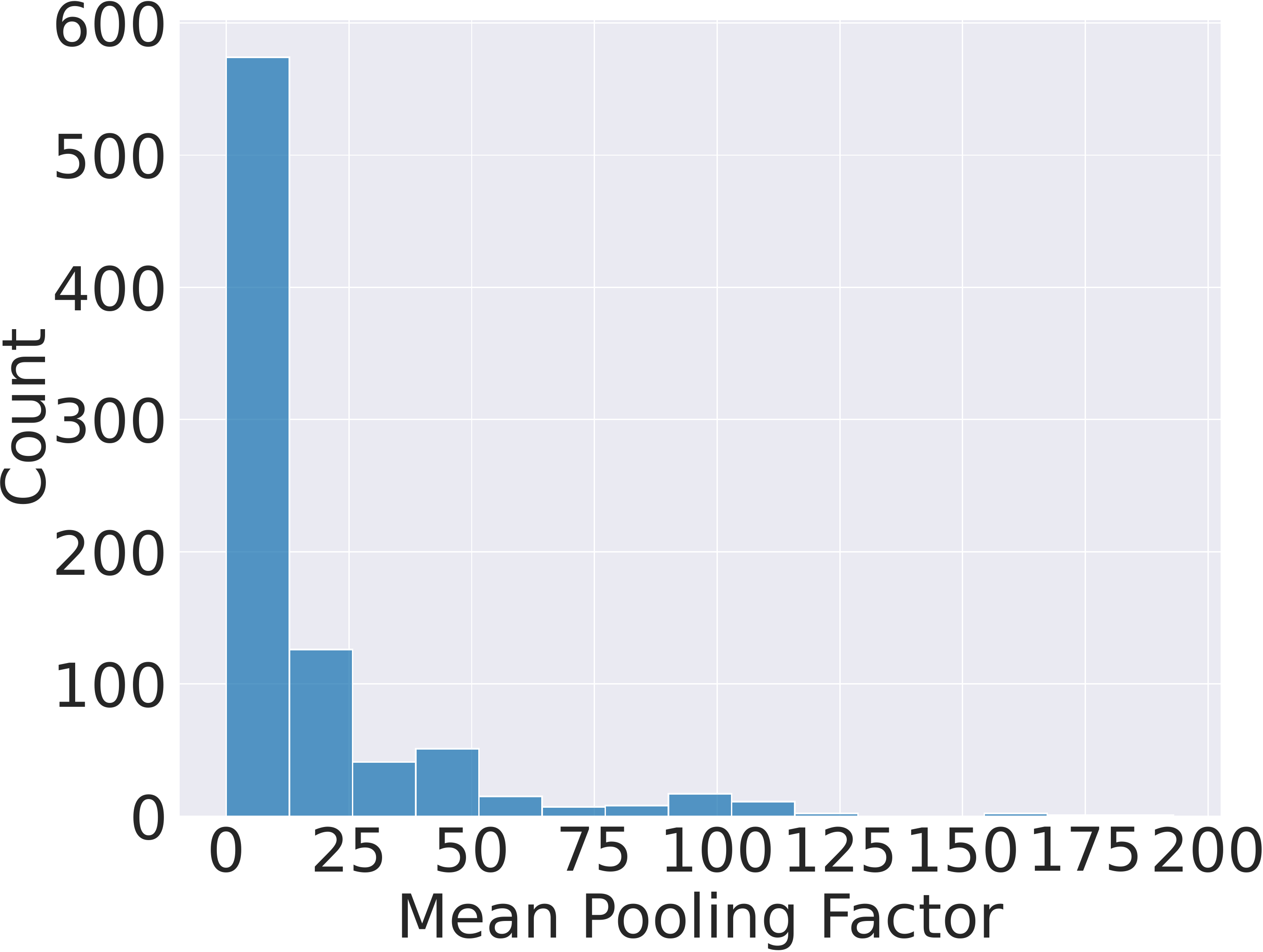}
  \caption{Pooling factor distribution.}
  \label{fig:distributionpooling}
\end{figure}

Figure~\ref{fig:distributionpooling} shows the distribution of mean pooling factors. We find that the pooling factor generally follows a power-law distribution. Most of the tables have a pooling less than 5, while there are few tables that have a pooling factor larger than 100 (some certain tables can have a pooling factor of up to 200). Recall that the pooling factor is one of the most important factors that decide the computation workloads. The power-law distribution will make the computation easily imbalanced across devices.

\begin{figure}[h!]
  \centering
    \includegraphics[width=0.5\textwidth]{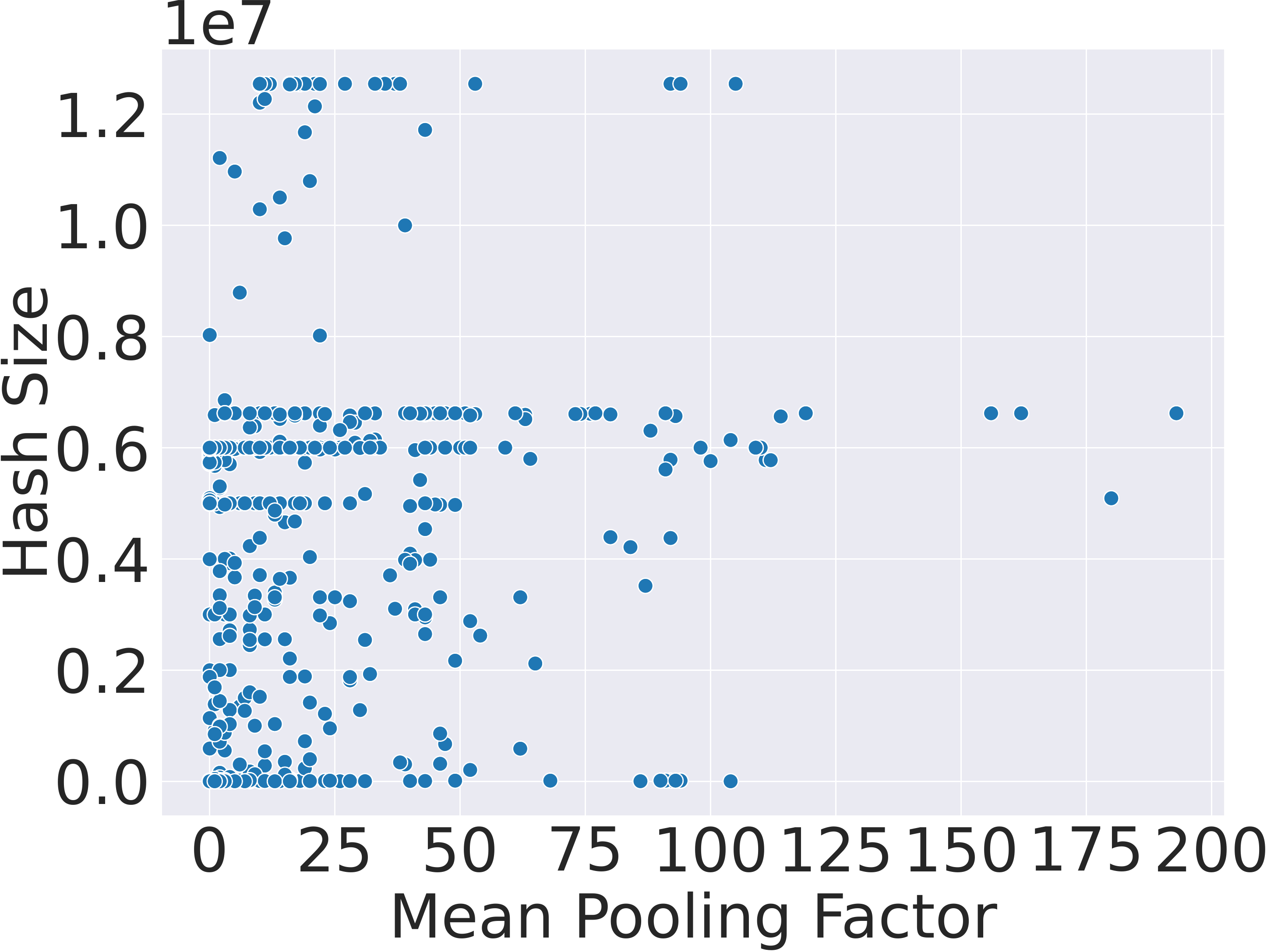}
  \caption{Hash size vs. pooling factor.}
  \label{fig:hashvspooling}
\end{figure}

We are interested in studying whether the pooling factor and the hash size have a positive correlation. The intuition is that if a table has more values, more rows could be selected when performing embedding lookup. If they have a positive correlation, balancing one of them could also lead to a balance of the other. We plot their relationship in Figure~\ref{fig:hashvspooling}. We observe that there is no clear relationship between the hash size and pooling factor. Thus, an ideal algorithm may need to balance both of them to achieve the best results.

\begin{figure}[h!]
  \centering
    \includegraphics[width=0.5\textwidth]{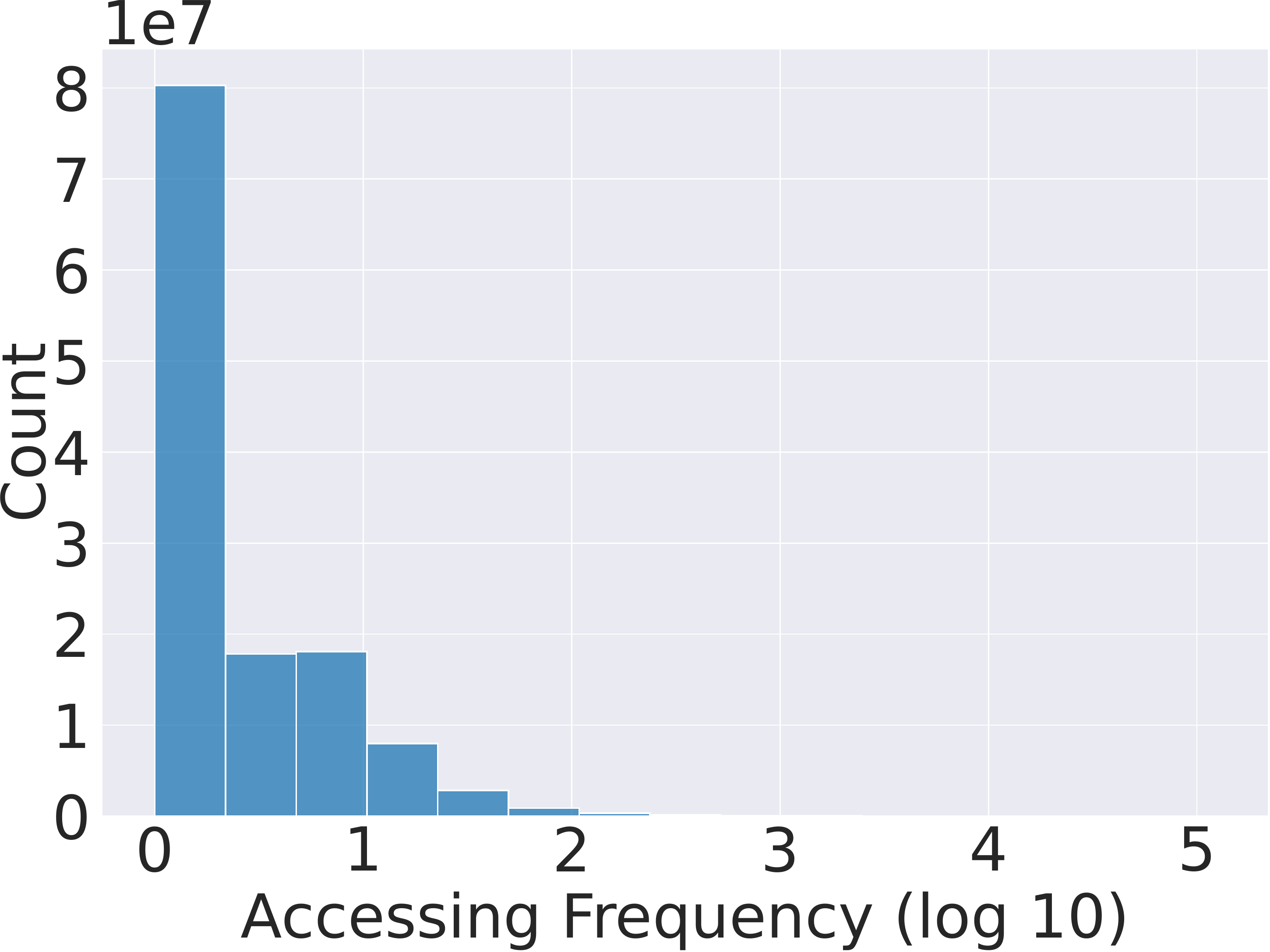}
  \caption{Indices frequency distribution.}
  \label{fig:distributionindex}
\end{figure}

Figure~\ref{fig:distributionindex} illustrates the indices accessing frequency distribution. We observe that most of the indices are accessed less than ten times, while some of them can reach $10^5$. Similarly, the diverse indices accessing frequency will easily lead to imbalances.

Overall, we find that the table features are quite diverse, and can easily lead to imbalances. Specifically, if we do not carefully partition the tables, some tables with high computation costs can be easily put into the same device, resulting in a very high cost for the device. Meanwhile, the imbalances may also lead to heavy communication costs. Thus, an ideal placement algorithm need to globally balance different aspects to achieve the best results. This motivates us to develop learning-based algorithms for embedding table placement.

\subsection{Data Processing}

Since the DLRM dataset does not specify the table dimension, we set the table dimension to be 16 for all the tables. We purposely make the dimensions small to facilitate reproducibility on GPUs with small memory. Note that our production dataset in general has a much larger table dimension that is up to 768. In addition, each table can have a different dimension. The large and diverse table dimensions will make the embedding table placement problem more challenging since imbalanced dimensions will significantly and negatively impact both the computation and the communication times. Nevertheless, our results in Table~\ref{tbl:mainresults} suggest that DreamShard can well deal with the tables that have larger and more diverse dimensions, showing the effectiveness and flexibility of our algorithm.

\section{Details of the Baselines}
\label{appendix:D}

We compare DreamShard with two types of baselines, including human expert strategies~\cite{sethi2022recshard,acun2021understanding,lui2021understanding}, and an RNN-based placement algorithm~\cite{mirhoseini2017device}. We will elaborate on them below.

\subsection{Human Expert Strategies}
These strategies have been mentioned or used in previous work of distributed recommender systems~\cite{sethi2022recshard,acun2021understanding,lui2021understanding}, and we have adopted them in our internal training workflow for years. The main idea is to use a greedy algorithm to balance the costs, where the costs are estimated based on a specific table feature, or a combination of the table features. These strategies consist of two steps as follows.

\begin{itemize}
    \item \textbf{Cost function:} Each table will be assigned an estimated cost, which serves as the target to be balanced.
    \item \textbf{Greedy algorithm:} The greedy algorithm tries to balance the sum of the costs in each device. Specifically, it first sorts all the embedding tables in descending order based on their costs. In this way, we can more easily achieve a balance if placing the tables greedily. Then starting from the table with the highest cost, we make a greedy decision in each step by placing the current table to the device that has the lowest sum of the cost so far. In the end, each device will roughly have the same or a similar sum of the costs so that we can achieve the goal of load balance.
\end{itemize}

The various expert strategies mainly differ in how the cost function is designed, i.e., the balancing objective. Specifically, the following cost functions are used as baselines to balance different aspects of the workloads: 
\begin{itemize}
    \item \textbf{Size-based:} We use the table size to estimate the cost. The intuition is that the table size is related to both the dimension and the hash size, which can reflect the workloads. In addition, balancing the size can reduce the risk of memory explosion.
    \item \textbf{Dim-based:} We use the table dimension to estimate the cost. Recall that in Section~\ref{appendix:A3}, table dimension can determine both computation and communication workloads. In particular, dimension is the only factor for communication workloads theoretically. Thus, balancing the sums of dimensions is a natural idea. 
    \item \textbf{Lookup-based:} We use the product of the table dimension and the pooling factor to estimate the cost. The motivation is that the table dimension and the pooling factor determine the computation workload in lookup.
    \item \textbf{Size-lookup-based:} We use the product of the table dimension, the pooling factor, and the table size to estimate the cost. This is the most comprehensive estimation (but it may not necessarily be the best).
\end{itemize}

The human expert strategies have several drawbacks. \textbf{First,} the estimation could be inaccurate. As shown in Section~\ref{appendix:a31}, the actual cost has a non-linear relationship with all the table features and can not be simply approximated with products. \textbf{Second,} it only optimizes the sum of the costs and can not model the operation fusion, as analyzed in Section~\ref{appendix:a32}. \textbf{Third,} while these strategies could achieve good performance in different scenarios, none of them can accommodate all scenarios. For example, if the communication bandwidths are low and communication is the bottleneck, the dim-based strategy could work better. Whereas, if the computation is the bottleneck, the lookup-based strategy may work better. It is difficult to select the most suitable one for real-world applications.

DreadShard addressed all of the above drawbacks with a learning-based cost network and a learning-based placement policy. The cost network directly approximates the multi-table costs in a data-driven manner, which can model the non-linear relationship between the cost and the table features. It can also inherently consider the operation fusion effect since it directly approximates the multi-table costs. Moreover, the RL-based placement policy makes decisions in a data-driven manner so that it can accommodate different scenarios.

\subsection{RNN-based Algorithm}

The main motivation for adopting this baseline is that embedding table placement also belongs to general device placement problems. The state-of-the-art algorithms leverage RL to optimize the device placement~\cite{mirhoseini2017device,gao2018spotlight,addanki2019placeto}. Thus, adapting the existing device placement algorithms to the embedding table placement problem is a natural idea. We focus on the RNN-based method proposed in~\cite{mirhoseini2017device} because it is a pioneering work that applies RL to device placement problems, and many of the follow-up studies are motivated by and developed based on this work.

The original RNN-based algorithm uses an RNN controller to sequentially make decisions for device placement, and the RNN controller is updated with the RL loss. \textbf{First,} each operation is represented as some operation features, such as data types and output shapes. \textbf{Second,} the operation representations are sequentially fed into an RNN architecture. \textbf{Third,} an attention layer is applied to the hidden states. \textbf{Fourth,} the representation obtained after the attention layer is followed by a policy head to make predictions. \textbf{Finally,} the RNN controller will be updated using the standard policy gradient loss.

We have adapted the original RNN-based algorithm so that it can be applied to our embedding table placement problem. Specifically, we replace the operation features with the 21 table features used in DreamShard. Additionally, we use the same feature extraction MLP with the same architecture as DreamShard. The policy head of the RNN-based baseline also has the same architecture as the policy head in DreadShard. The main difference is that we use an RNN and an attention layer to process the feature representations. We note that such design can not generalize across different numbers of devices due to architecture constraints of RNN.

\section{Details of the Experimental Configurations}
\label{appendix:E}
In this section, we provide more details of how we perform the experiments to test the generalizability of DreamShard. We consider three types of generalizability, including 1) unseen placement tasks (i.e., the combination of the tables is different, but the individual tables may or may not be seen in training), 2) unseen embedding tables, and 3) different numbers of tables/devices. Our experiments are designed to maximally test all these three types of generalizability.

To test 1) and 2), we control the table pools for training and testing. Specifically, we divide all the tables in half to construct a training pool and a testing pool, where the training tasks are constructed only based on the training pool, and the testing tasks are constructed only based on the testing pool. Since there is no overlap of tables between training and testing pools, all the tables in the testing tasks are unseen. To construct each training/testing task, we randomly sample a subset of tables from the corresponding pool, and the number of tables varies from the set $\{10, 20, 30, 40, 50, 60, 70, 80, 90, 100\}$; that is, we consider different combinations of the tables, and we consider the cases from very few tables to many tables. All the table combinations in the testing tasks are naturally unseen by the algorithm. To test 3), we conducted experiments by directly transferring a trained DreamShard to a task with a different number of tables and/or devices without fine-tuning.

Our comprehensive analysis shows that DreamShard can generalize across different table combinations and numbers of tables and/or devices, making it desirable for real-world applications.

\section{Additional Results of DreamShard against Baselines}
\label{appendix:F}

\begin{table}[h!]
\centering
%~\yuandong{Do we have number of steps, which is independent of the machine used for reproducibility.}
% \yuandong{I would prefer a figure rather than a table, or at least a table with decimated rows.}
\caption{Additional results of running time in milliseconds and relative speedups over random placement on DLRM tasks, measured on 4 GPUs.}

\tiny
\setlength{\tabcolsep}{1.3pt}
\begin{tabular}{l|l|l|l|l|l|l|l|l}
\toprule
\multicolumn{2}{c|}{\multirow{2}{*}{Task}} & No strategy & \multicolumn{4}{c|}{Human Experts} & \multicolumn{2}{c}{RL} \\
\cline{3-9}
\multicolumn{2}{c|}{~} & \multicolumn{1}{c|}{Random} & \multicolumn{1}{c|}{Size-based} & \multicolumn{1}{c|}{Dim-based} & \multicolumn{1}{c|}{Lookup-based} & \multicolumn{1}{c|}{Size-lookup-based} & \multicolumn{1}{c|}{RNN-based} & \multicolumn{1}{c}{DreamShard} \\
\midrule
\multirow{2}{*}{DLRM-10 (4)} & Train & 14.8$\pm$0.3 & 13.0$\pm$0.0 (+13.8\%) & 12.8$\pm$0.0 (+15.6\%) & 11.9$\pm$0.0 (+24.4\%) & 12.0$\pm$0.0 (+23.3\%) & 13.3$\pm$0.3 (+11.3\%) & \textbf{11.6$\pm$0.3 (+27.6\%)} \\
~ & Test & 13.6$\pm$0.3 & 13.0$\pm$0.0 (+4.6\%) & 12.6$\pm$0.0 (+7.9\%) & 11.1$\pm$0.0 (+22.5\%) & 9.6$\pm$0.0 (+21.4\%) & 12.4$\pm$0.1 (+9.7\%) & \textbf{10.9$\pm$0.3 (+24.8\%)} \\
\midrule

\multirow{2}{*}{DLRM-30 (4)} & Train & 32.3$\pm$0.5 & 31.0$\pm$0.0 (+4.2\%) & 28.8$\pm$0.1 (+12.2\%) & 26.1$\pm$0.0 (+23.8\%) & 26.2$\pm$0.0 (+23.3\%) & 30.7$\pm$0.8 (+5.2\%) & \textbf{25.4$\pm$0.3 (+27.2\%)} \\
~ & Test & 31.8$\pm$0.2 & 30.3$\pm$0.0 (+5.0\%) & 28.4$\pm$0.1 (+12.0\%) & 25.4$\pm$0.0 (+25.2\%) & 25.5$\pm$0.0 (+24.7\%) & 29.7$\pm$0.5 (+7.1\%) & \textbf{24.6$\pm$0.2 (+29.3\%)} \\
\midrule

\multirow{2}{*}{DLRM-50 (4)} & Train & 49.8$\pm$0.6 & 49.7$\pm$0.0 (+0.2\%) & 46.5$\pm$0.0 (+7.1\%) & 41.2$\pm$0.0 (+20.9\%) & 41.7$\pm$0.1 (+19.4\%) & 48.2$\pm$1.2 (+3.3\%) & \textbf{40.4$\pm$0.5 (+23.3\%)} \\
~ & Test & 49.8$\pm$0.3 & 49.8$\pm$0.0 (0.0\%) & 45.8$\pm$0.1 (+8.7\%) & 41.3$\pm$0.0 (+20.6\%) & 41.4$\pm$0.0 (+20.3\%) & 48.1$\pm$1.2 (+3.5\%) & \textbf{40.4$\pm$0.6 (+23.3\%)} \\
\midrule

\multirow{2}{*}{DLRM-70 (4)} & Train & 66.3$\pm$1.0 & 67.8$\pm$0.1 (-2.2\%) & 63.1$\pm$0.0 (+5.1\%) & 56.6$\pm$0.1 (+17.1\%) & 57.5$\pm$0.1 (+15.3\%) & 70.8$\pm$13.2 (-6.4\%) & \textbf{55.2$\pm$0.4 (+20.1\%)} \\
~ & Test & 66.7$\pm$0.7 & 69.4$\pm$0.1 (-3.9\%) & 61.9$\pm$0.2 (+7.8\%) & 56.5$\pm$0.0 (+18.1\%) & 57.2$\pm$0.0 (+16.6\%) & 71.8$\pm$15.3 (-7.1\%) & \textbf{55.2$\pm$0.8 (+20.8\%)} \\
\midrule

\multirow{2}{*}{DLRM-90 (4)} & Train & 83.0$\pm$1.5 & 82.9$\pm$0.0 (+0.1\%) & 77.9$\pm$0.3 (+6.5\%) & 73.1$\pm$0.0 (+13.5\%) & 73.5$\pm$0.0 (+12.9\%) & 92.4$\pm$13.3 (-10.2\%) & \textbf{70.0$\pm$0.4 (+18.6\%)} \\
~ & Test & 82.3$\pm$1.4 & 87.2$\pm$0.2 (-5.6\%) & 77.9$\pm$0.4 (+5.6\%) & 71.8$\pm$0.2 (+14.6\%) & 72.3$\pm$0.2 (+13.8\%) & 92.9$\pm$15.6 (-11.4\%) & \textbf{69.4$\pm$0.7 (+18.6\%)} \\

\bottomrule
\end{tabular}
\end{table}

\begin{table}[h!]
\centering
\caption{Running time in milliseconds and relative speedups over random placement on DLRM tasks, measured on 2 GPUs. We observe that DreamShard is comparable with human experts (slightly better than the size-lookup-based method). A possible reason is that these tasks are relatively simple so the expert placements are already near-optimal. Nevertheless, DreamShard still shows strong performance for all the tasks and achieves the best performance in 7 out of 10 tasks.}
\tiny
\setlength{\tabcolsep}{1.7pt}
\begin{tabular}{l|l|l|l|l|l|l|l|l}
\toprule
\multicolumn{2}{c|}{\multirow{2}{*}{Task}} & No strategy & \multicolumn{4}{c|}{Human Experts} & \multicolumn{2}{c}{RL} \\
\cline{3-9}
\multicolumn{2}{c|}{~} & \multicolumn{1}{c|}{Random} & \multicolumn{1}{c|}{Size-based} & \multicolumn{1}{c|}{Dim-based} & \multicolumn{1}{c|}{Lookup-based} & \multicolumn{1}{c|}{Size-lookup-based} & \multicolumn{1}{c|}{RNN-based} & \multicolumn{1}{c}{DreamShard} \\
\midrule
\multirow{2}{*}{DLRM-10 (2)} & Train & 17.9$\pm$0.2 & 16.4$\pm$0.0 (+9.1\%) & 16.5$\pm$0.0 (+8.5\%) & 14.8$\pm$0.0 (+20.9\%) & \textbf{14.7$\pm$0.0 (+21.8\%)} & 17.0$\pm$0.2 (+5.3\%) & 15.1$\pm$0.3 (+18.5\%) \\
~ & Test & 16.5$\pm$0.4 & 16.0$\pm$0.1 (+3.1\%) & 16.0$\pm$0.0 (+3.1\%) & 13.9$\pm$0.0 (+18.7\%) & \textbf{13.7$\pm$0.1 (+20.4\%)} & 16.0$\pm$0.2 (+3.1\%) & 13.9$\pm$0.2 (+18.7\%) \\
\midrule
\multirow{2}{*}{DLRM-20 (2)} & Train & 31.6$\pm$0.6 & 30.8$\pm$0.0 (+2.6\%) & 30.6$\pm$0.0 (+3.3\%) & 27.4$\pm$0.0 (+15.3\%) & 27.3$\pm$0.0 (+15.8\%) & 30.6$\pm$0.2 (+3.3\%) & \textbf{27.1$\pm$0.2 (+16.6\%)} \\
~ & Test & 29.9$\pm$0.4 & 29.3$\pm$0.0 (+2.0\%) & 28.8$\pm$0.0 (+3.8\%) & 26.3$\pm$0.0 (+13.7\%) & 26.0$\pm$0.0 (+15.0\%) & 28.8$\pm$0.2 (+3.8\%) & \textbf{25.8$\pm$0.2 (+15.9\%)} \\
\midrule
\multirow{2}{*}{DLRM-30 (2)} & Train & 44.6$\pm$0.6 & 43.4$\pm$0.0 (+2.8\%) & 43.0$\pm$0.0 (+3.7\%) & 39.5$\pm$0.0 (+12.9\%) & \textbf{39.3$\pm$0.0 (+13.5\%)} & 43.1$\pm$0.5 (+3.5\%) & \textbf{39.3$\pm$0.3 (+13.5\%)} \\
~ & Test & 43.7$\pm$0.4 & 42.6$\pm$0.1 (+2.6\%) & 42.1$\pm$0.0 (+3.8\%) & 38.9$\pm$0.1 (+12.3\%) & \textbf{38.5$\pm$0.0 (+13.5\%)} & 42.4$\pm$0.1 (+3.1\%) & 38.6$\pm$0.4 (+13.2\%) \\
\midrule
\multirow{2}{*}{DLRM-40 (2)} & Train & 58.7$\pm$0.6 & 57.1$\pm$0.1 (+2.8\%) & 56.2$\pm$0.1 (+4.4\%) & 53.0$\pm$0.0 (+10.8\%) & 52.5$\pm$0.0 (+11.8\%) & 57.5$\pm$0.7 (+2.1\%) & \textbf{52.3$\pm$0.3 (+12.2\%)} \\
~ & Test & 58.6$\pm$0.7 & 56.9$\pm$0.0 (+3.0\%) & 56.9$\pm$0.0 (+3.0\%) & 52.5$\pm$0.0 (+11.6\%) & 52.4$\pm$0.0 (+11.8\%) & 56.5$\pm$0.4 (+3.7\%) & \textbf{51.9$\pm$0.1 (+12.9\%)} \\
\midrule
\multirow{2}{*}{DLRM-50 (2)} & Train & 72.2$\pm$1.2 & 71.2$\pm$0.0 (+1.4\%) & 70.0$\pm$0.0 (+3.1\%) & 66.0$\pm$0.0 (+9.4\%) & \textbf{65.5$\pm$0.0 (+10.2\%)} & 71.5$\pm$0.4 (+1.0\%) & \textbf{65.5$\pm$0.2 (+10.2\%)} \\
~ & Test & 72.7$\pm$0.6 & 70.6$\pm$0.0 (+3.0\%) & 70.7$\pm$0.0 (+2.8\%) & 65.7$\pm$0.0 (+10.7\%) & 65.6$\pm$0.0 (+10.8\%) & 70.8$\pm$0.5 (+2.7\%) & \textbf{65.5$\pm$0.3 (+11.0\%)} \\
\bottomrule
\end{tabular}
\end{table}

\section{Additional Results of Generalizability}
\label{appendix:G}

\begin{table}[h!]
\centering
\caption{Additional results of the generalization performance of DreamShard from source tasks to target tasks w.r.t. to different numbers of tables. In general, DreamShard can transfer to tasks with different numbers of tables with competitive or even better performances.}
\scriptsize
\setlength{\tabcolsep}{5.0pt}
\begin{tabular}{l|c|c|c|c|c}
\toprule

\diagbox [width=7em,trim=l] {Source}{Target} & DLRM-20 (4) & DLRM-40 (4) & DLRM-60 (4) & DLRM-80 (4) & DLRM-100 (4)  \\
\midrule
DLRM-20 (4) & - & 32.5$\pm$0.3 & 47.8$\pm$0.2 & 62.8$\pm$0.4 & 77.9$\pm$0.4 \\
DLRM-40 (4) & 17.6$\pm$0.1 & - & 47.8$\pm$0.4 & 62.7$\pm$0.5 & 78.0$\pm$0.5 \\
DLRM-60 (4) &17.7$\pm$0.1 & 32.5$\pm$0.2 & - & 63.1$\pm$0.4 & 78.2$\pm$0.5 \\
DLRM-80 (4) & 17.6$\pm$0.1 & 32.4$\pm$0.2 & 47.8$\pm$0.3 & - & 78.1$\pm$0.5 \\
DLRM-100 (4) & 17.7$\pm$0.3 & 32.7$\pm$0.4 & 48.1$\pm$0.6 & 63.2$\pm$0.9 & - \\
\midrule
DreamShard trained on target & 17.6$\pm$0.2 & 32.4$\pm$0.3 & 47.9$\pm$0.7 & 62.7$\pm$0.3 & 77.8$\pm$0.8 \\

\bottomrule
\end{tabular}
\end{table}

\begin{table}[h!]
\centering
\caption{Additional results of the generalization performance of DreamShard from source tasks with 4 GPUs to target tasks with 2 GPUs w.r.t. to different numbers of tables. In general, DreamShard can transfer to tasks with different numbers of tables and fewer GPUs with competitive or even better performances.}
\scriptsize
\setlength{\tabcolsep}{5.0pt}
\begin{tabular}{l|c|c|c|c|c}
\toprule

\diagbox [width=7em,trim=l] {Source}{Target} & DLRM-10 (2) & DLRM-20 (2) & DLRM-30 (2) & DLRM-40 (2) & DLRM-50 (2)  \\
\midrule
DLRM-10 (4) & 14.1$\pm$0.2 & 26.2$\pm$0.3 & 38.7$\pm$0.5 & 52.2$\pm$0.7 & 65.3$\pm$1.2 \\
DLRM-20 (4) & 13.9$\pm$0.1 & 25.8$\pm$0.1 & 38.1$\pm$0.1 & 51.4$\pm$0.2 & 64.5$\pm$0.1 \\
DLRM-30 (4) & 14.1$\pm$0.1 & 26.1$\pm$0.2 & 38.5$\pm$0.2 & 52.0$\pm$0.2 & 65.2$\pm$0.2 \\
DLRM-40 (4) & 14.3$\pm$0.1 & 26.2$\pm$0.1 & 38.6$\pm$0.2 & 52.0$\pm$0.3 & 65.1$\pm$0.2 \\
DLRM-50 (4) & 14.3$\pm$0.4 & 26.3$\pm$0.3 & 38.6$\pm$0.3 & 52.1$\pm$0.4 & 65.3$\pm$0.6 \\
\midrule
DreamShard trained on target & 13.9$\pm$0.2 & 25.8$\pm$0.2 & 38.6$\pm$0.4 & 51.9$\pm$0.1 & 65.5$\pm$0.3 \\

\bottomrule
\end{tabular}
\end{table}

\begin{table}[h!]
\centering
\caption{Additional results of the generalization performance of DreamShard from source tasks with 2 GPUs to target tasks with 4 GPUs w.r.t. to different numbers of tables. In general, DreamShard can transfer to tasks with different numbers of tables and more GPUs with competitive or even better performances.}
\scriptsize
\setlength{\tabcolsep}{5.0pt}
\begin{tabular}{l|c|c|c|c|c}
\toprule

\diagbox [width=7em,trim=l] {Source}{Target} & DLRM-10 (4) & DLRM-20 (4) & DLRM-30 (4) & DLRM-40 (4) & DLRM-50 (4)  \\
\midrule
DLRM-10 (2) & 10.8$\pm$0.3 & 18.3$\pm$0.4 & 25.6$\pm$0.6 & 33.8$\pm$0.7 & 41.7$\pm$0.9 \\
DLRM-20 (2) & 10.6$\pm$0.1 & 17.8$\pm$0.3 & 25.0$\pm$0.4 & 32.9$\pm$0.4 & 40.7$\pm$0.6 \\
DLRM-30 (2) &10.9$\pm$0.3 & 18.0$\pm$0.4 & 25.0$\pm$0.6 & 32.9$\pm$0.7 & 40.7$\pm$0.7 \\
DLRM-40 (2) & 10.8$\pm$0.1 & 17.8$\pm$0.2 & 24.8$\pm$0.2 & 32.6$\pm$0.3 & 40.2$\pm$0.3 \\
DLRM-50 (2) & 10.7$\pm$0.1 & 17.6$\pm$0.1 & 24.6$\pm$0.1 & 32.3$\pm$0.2 & 40.0$\pm$0.3 \\
\midrule
DreamShard trained on target & 10.9$\pm$0.3 & 17.6$\pm$0.2 & 24.6$\pm$0.2 & 32.4$\pm$0.3 & 40.4$\pm$0.6 \\

\bottomrule
\end{tabular}
\end{table}

\newpage
\section{Additional Results of Training Efficiency}
\label{appendix:H}

\begin{figure}[h!]
  \centering

  \begin{subfigure}[b]{0.25\textwidth}
    \centering
    \includegraphics[width=0.99\textwidth]{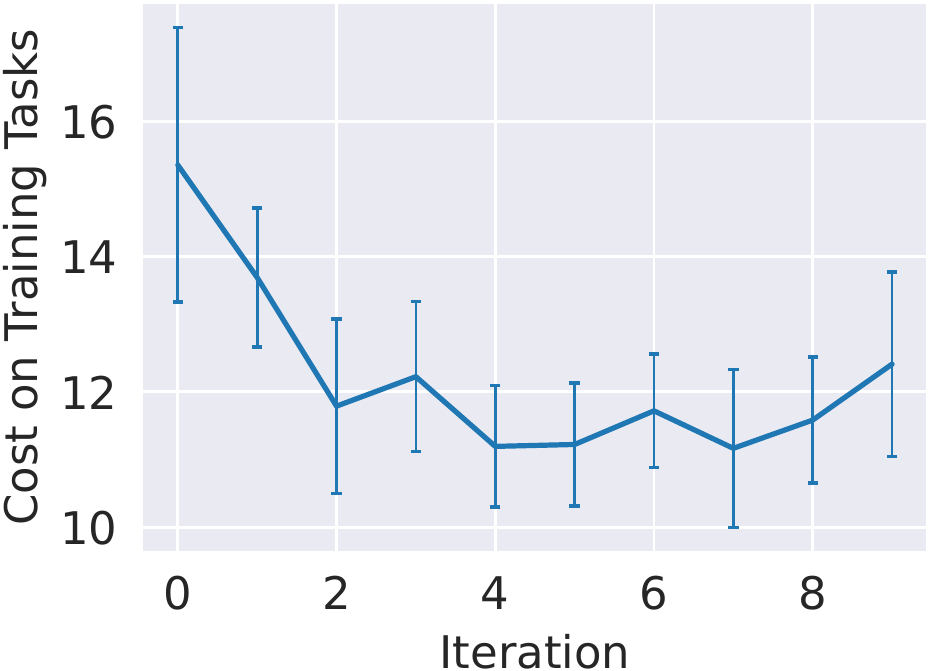}
    \caption{DLRM-10}
  \end{subfigure}%
  \begin{subfigure}[b]{0.25\textwidth}
    \centering
    \includegraphics[width=0.99\textwidth]{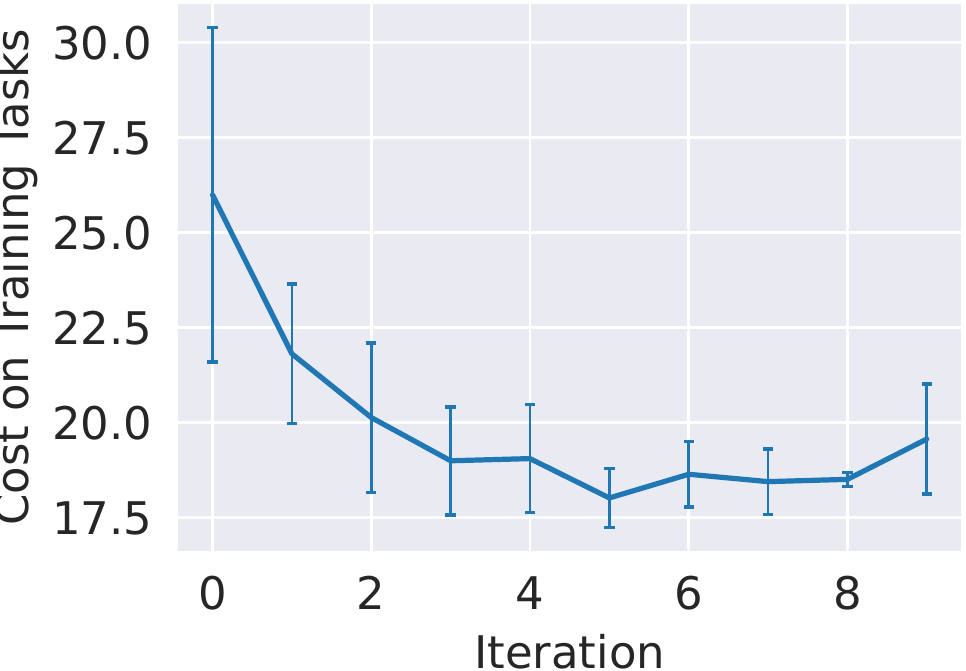}
    \caption{DLRM-20}
  \end{subfigure}%
  \begin{subfigure}[b]{0.25\textwidth}
    \centering
    \includegraphics[width=0.99\textwidth]{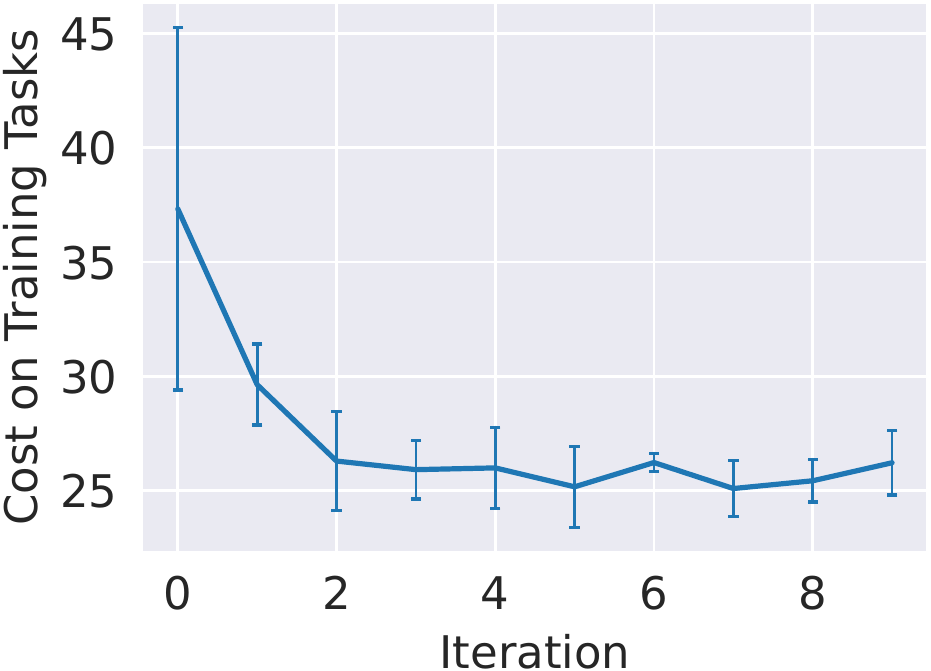}
    \caption{DLRM-30}
  \end{subfigure}%
  \begin{subfigure}[b]{0.25\textwidth}
    \centering
    \includegraphics[width=0.99\textwidth]{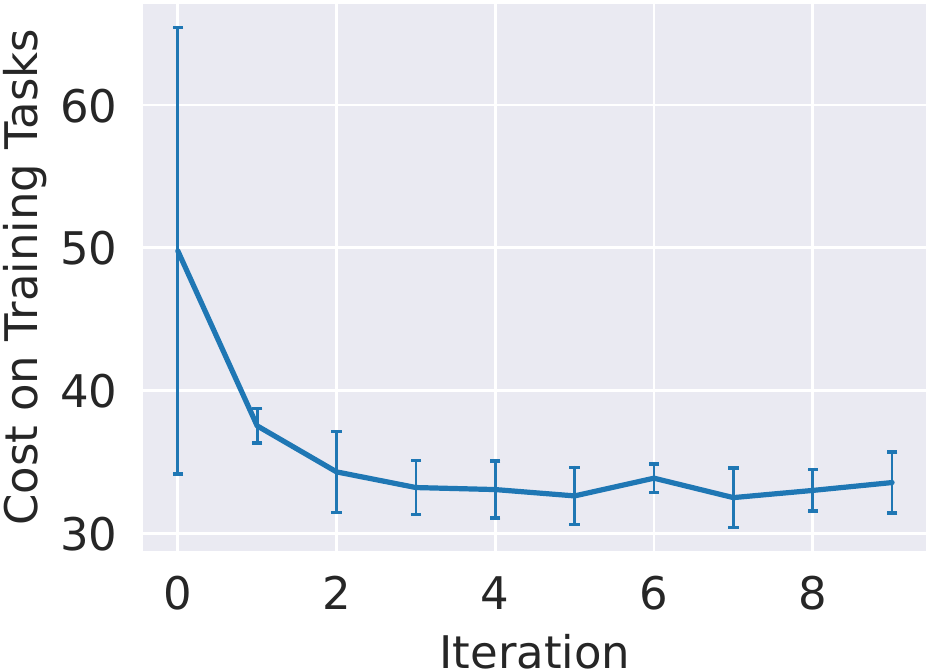}
    \caption{DLRM-40}
  \end{subfigure}%
  
  \begin{subfigure}[b]{0.25\textwidth}
    \centering
    \includegraphics[width=0.99\textwidth]{figs/rq3/50_iteration.pdf}
    \caption{DLRM-50}
  \end{subfigure}%
  \begin{subfigure}[b]{0.25\textwidth}
    \centering
    \includegraphics[width=0.99\textwidth]{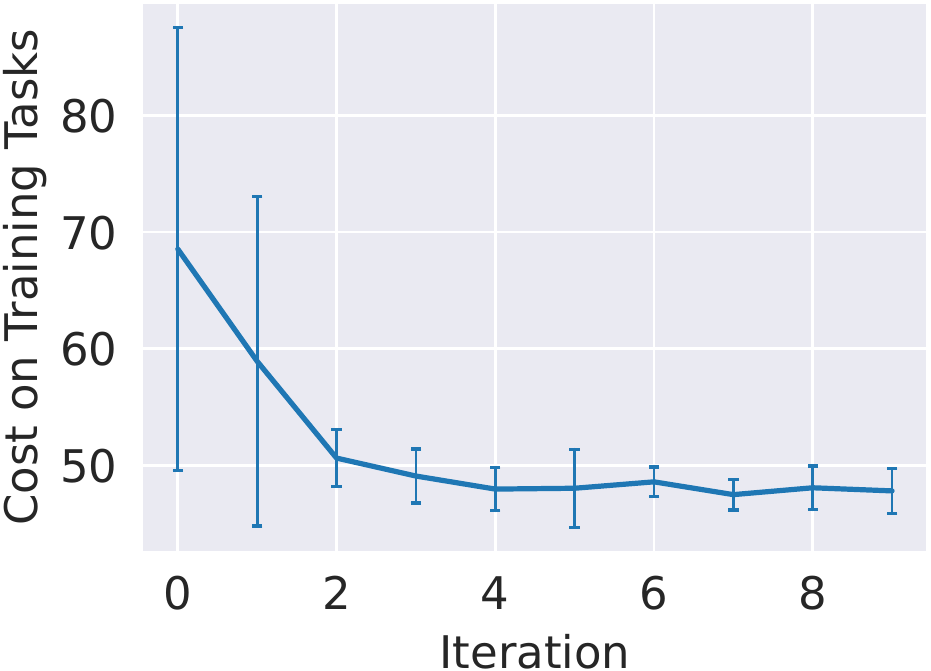}
    \caption{DLRM-60}
  \end{subfigure}%
  \begin{subfigure}[b]{0.25\textwidth}
    \centering
    \includegraphics[width=0.99\textwidth]{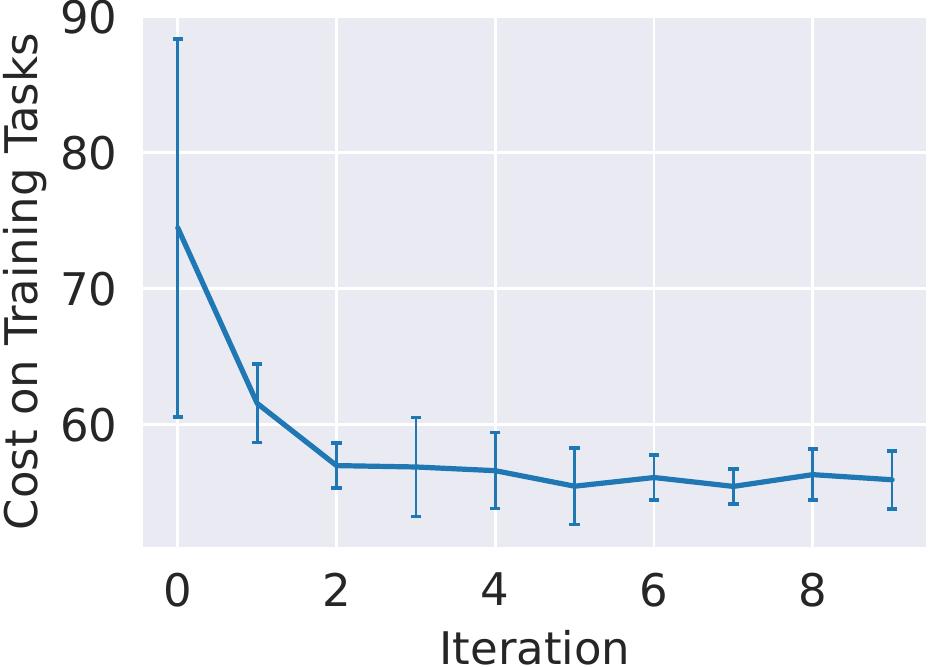}
    \caption{DLRM-70}
  \end{subfigure}%
  \begin{subfigure}[b]{0.25\textwidth}
    \centering
    \includegraphics[width=0.99\textwidth]{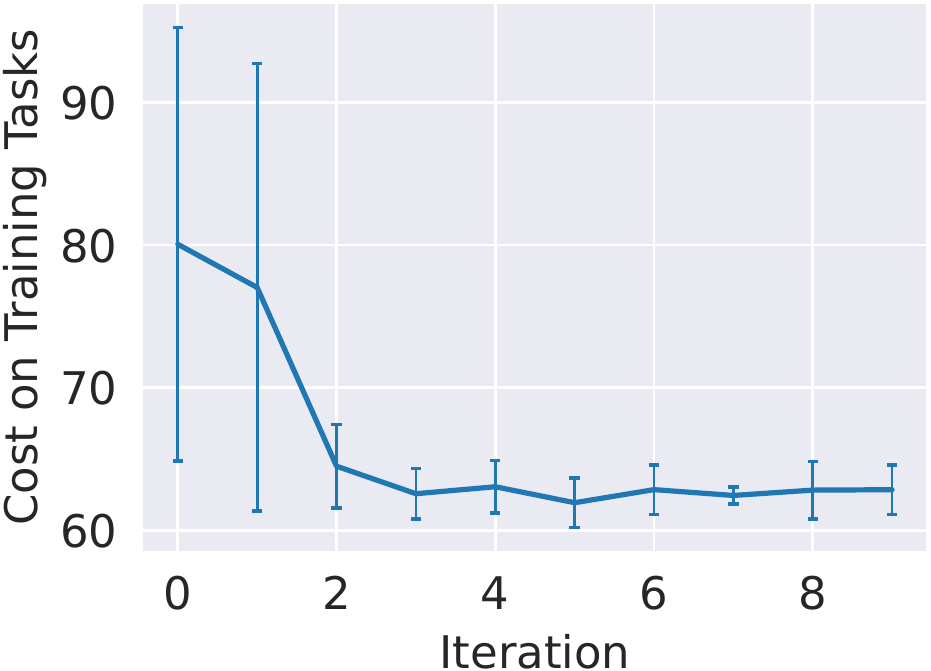}
    \caption{DLRM-80}
  \end{subfigure}%
  
  \begin{subfigure}[b]{0.25\textwidth}
    \centering
    \includegraphics[width=0.99\textwidth]{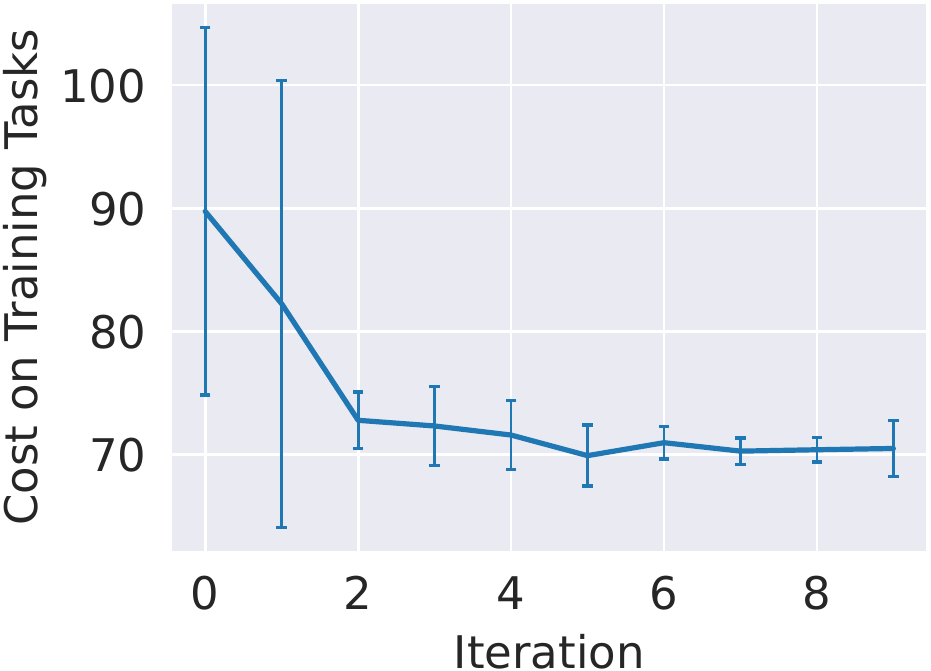}
    \caption{DLRM-90}
  \end{subfigure}%
  \begin{subfigure}[b]{0.25\textwidth}
    \centering
    \includegraphics[width=0.99\textwidth]{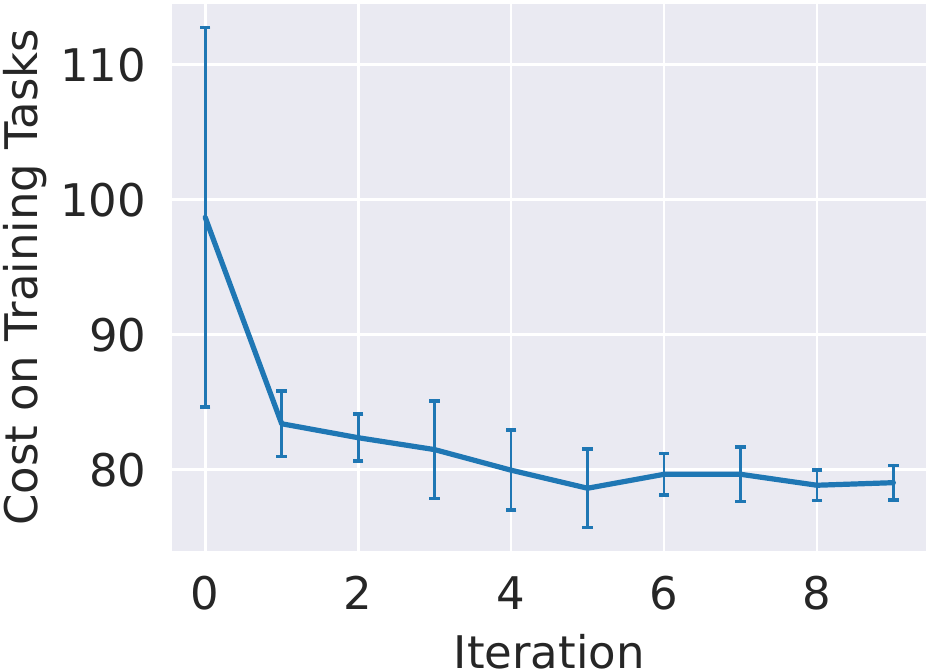}
    \caption{DLRM-100}
  \end{subfigure}%

  \caption{Performance ($\downarrow$) of DreamShard w.r.t. the numbers of iterations. DreamShard achieves strong performance with very few iterations.}
\end{figure}

\begin{figure}[h!]
  \centering

  \begin{subfigure}[b]{0.25\textwidth}
    \centering
    \includegraphics[width=0.99\textwidth]{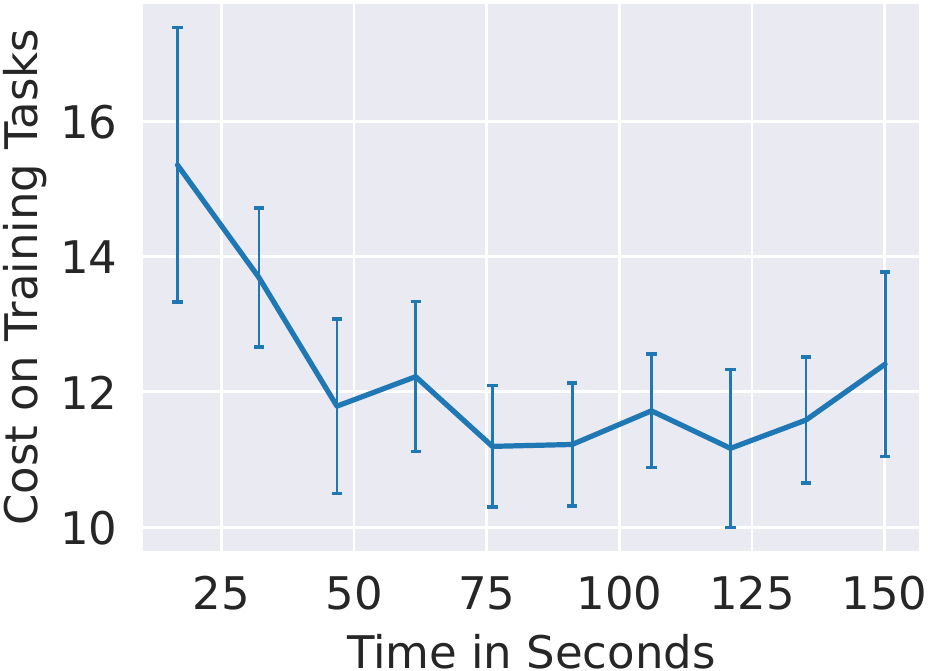}
    \caption{DLRM-10}
  \end{subfigure}%
  \begin{subfigure}[b]{0.25\textwidth}
    \centering
    \includegraphics[width=0.99\textwidth]{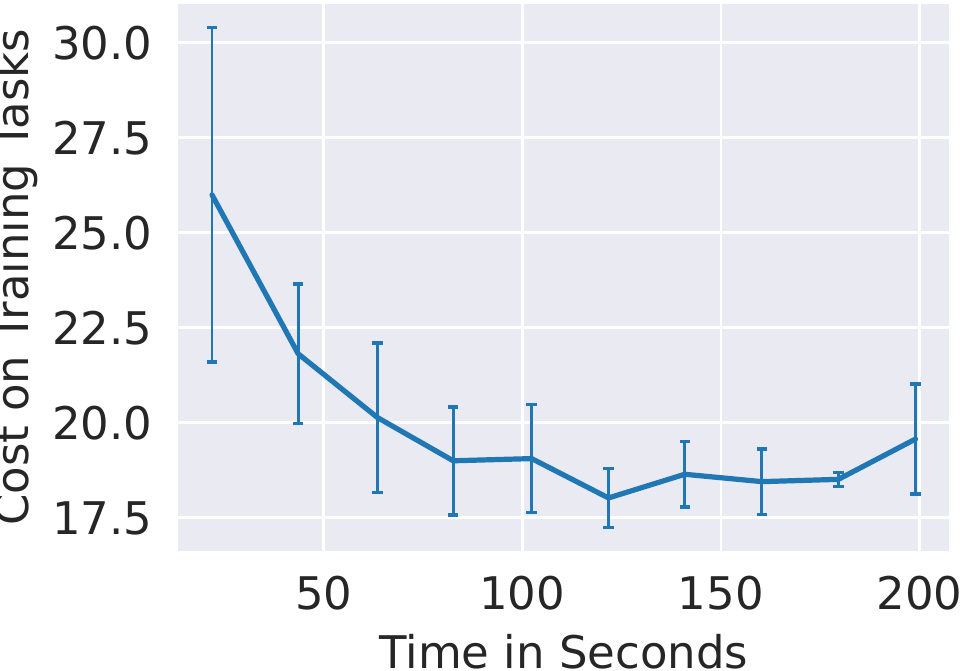}
    \caption{DLRM-20}
  \end{subfigure}%
  \begin{subfigure}[b]{0.25\textwidth}
    \centering
    \includegraphics[width=0.99\textwidth]{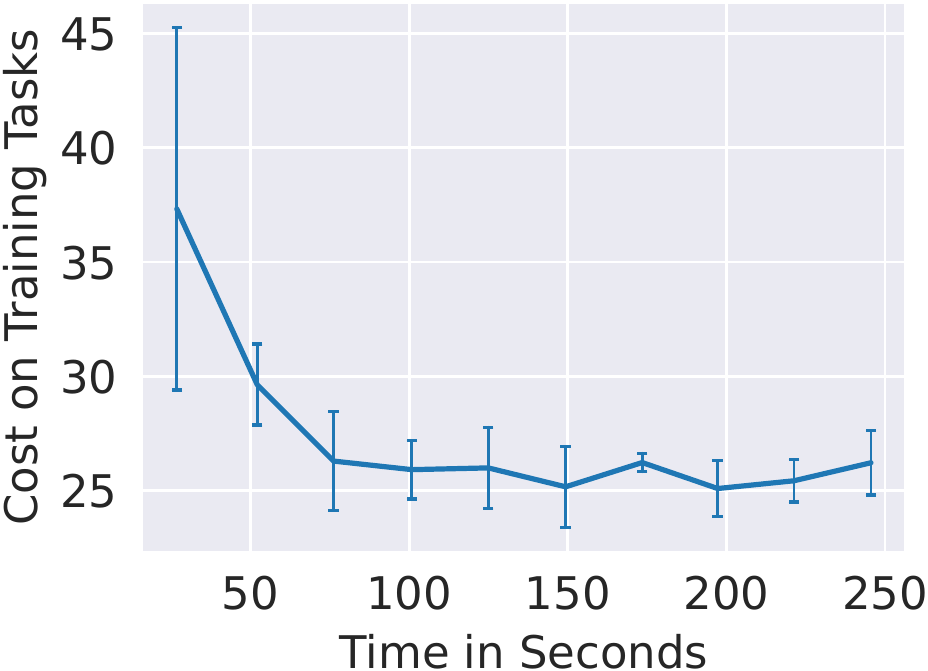}
    \caption{DLRM-30}
  \end{subfigure}%
  \begin{subfigure}[b]{0.25\textwidth}
    \centering
    \includegraphics[width=0.99\textwidth]{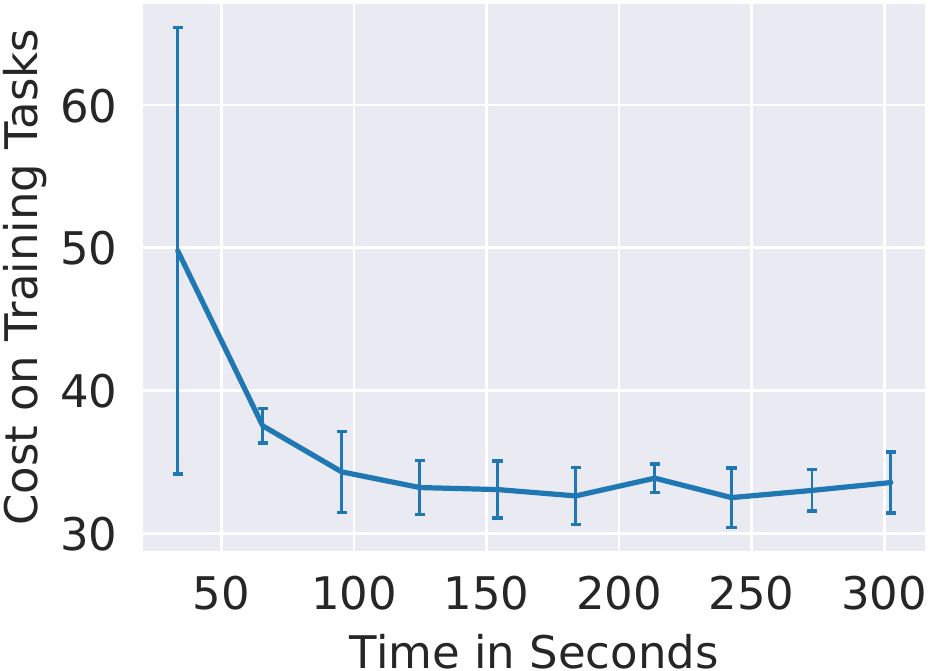}
    \caption{DLRM-40}
  \end{subfigure}%
  
  \begin{subfigure}[b]{0.25\textwidth}
    \centering
    \includegraphics[width=0.99\textwidth]{figs/rq3/50_time.pdf}
    \caption{DLRM-50}
  \end{subfigure}%
  \begin{subfigure}[b]{0.25\textwidth}
    \centering
    \includegraphics[width=0.99\textwidth]{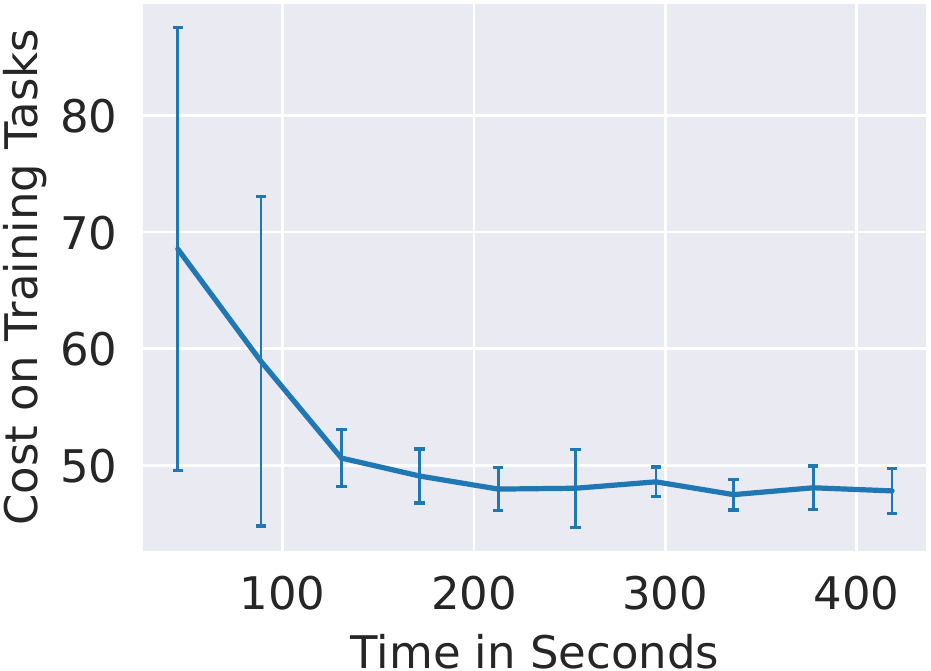}
    \caption{DLRM-60}
  \end{subfigure}%
  \begin{subfigure}[b]{0.25\textwidth}
    \centering
    \includegraphics[width=0.99\textwidth]{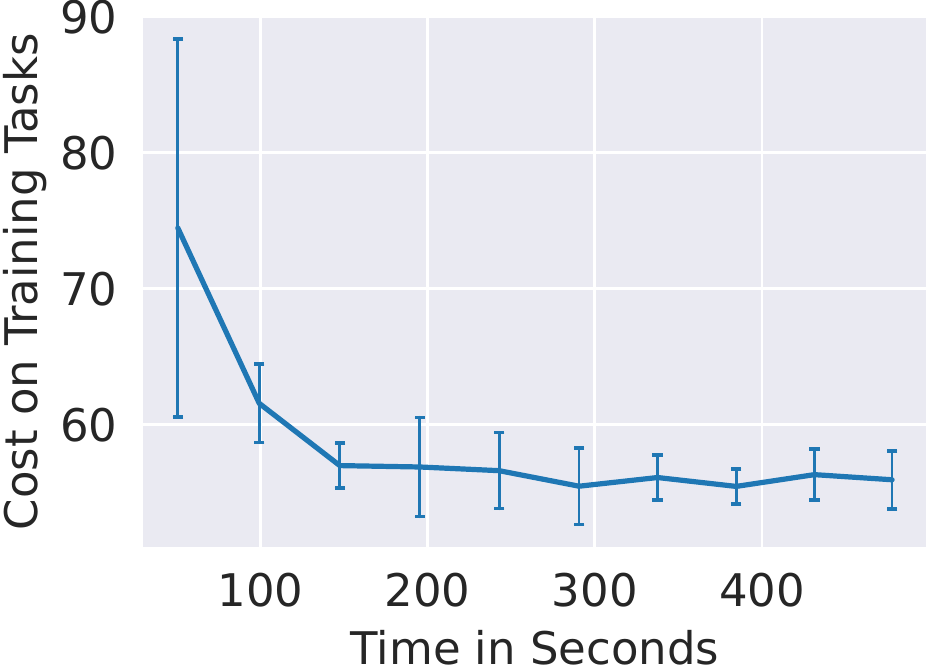}
    \caption{DLRM-70}
  \end{subfigure}%
  \begin{subfigure}[b]{0.25\textwidth}
    \centering
    \includegraphics[width=0.99\textwidth]{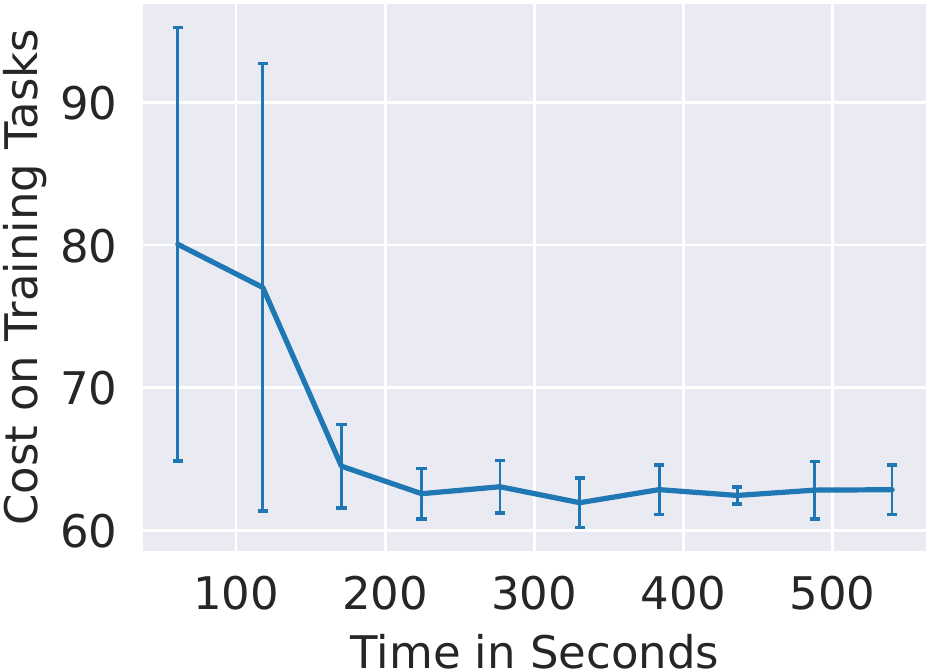}
    \caption{DLRM-80}
  \end{subfigure}%
  
  \begin{subfigure}[b]{0.25\textwidth}
    \centering
    \includegraphics[width=0.99\textwidth]{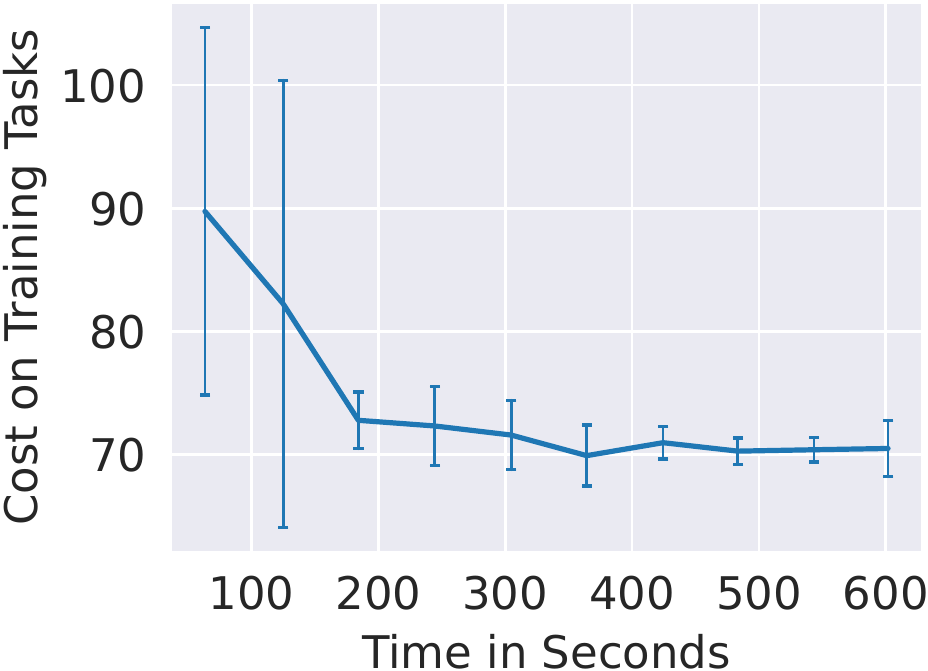}
    \caption{DLRM-90}
  \end{subfigure}%
  \begin{subfigure}[b]{0.25\textwidth}
    \centering
    \includegraphics[width=0.99\textwidth]{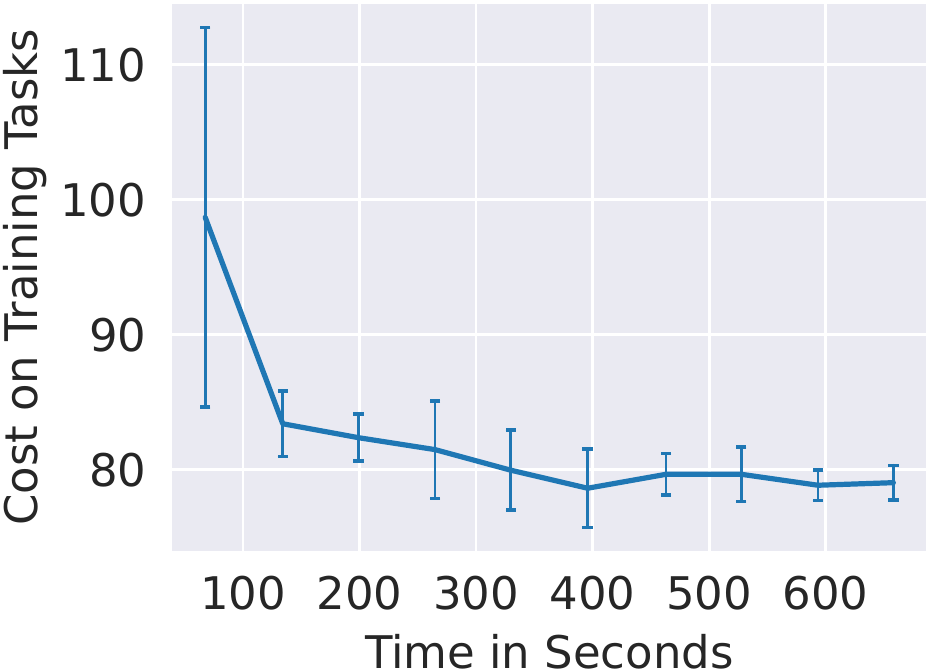}
    \caption{DLRM-100}
  \end{subfigure}%

  \caption{Performance ($\downarrow$) of DreamShard w.r.t. running time. DreamShard achieves strong performance in a short time.}
\end{figure}

\clearpage
\newpage
\section{Additional Results of Hyperparameter Study}
\label{appendix:I}

\begin{figure}[h!]
  \centering

  \begin{subfigure}[b]{0.25\textwidth}
    \centering
    \includegraphics[width=0.99\textwidth]{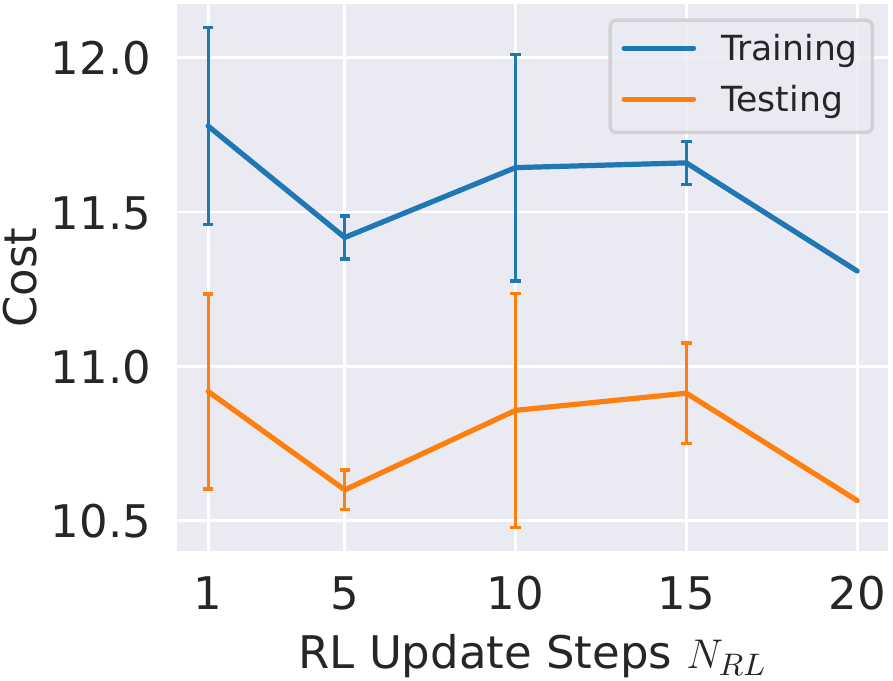}
    \caption{DLRM-10}
  \end{subfigure}%
  \begin{subfigure}[b]{0.25\textwidth}
    \centering
    \includegraphics[width=0.99\textwidth]{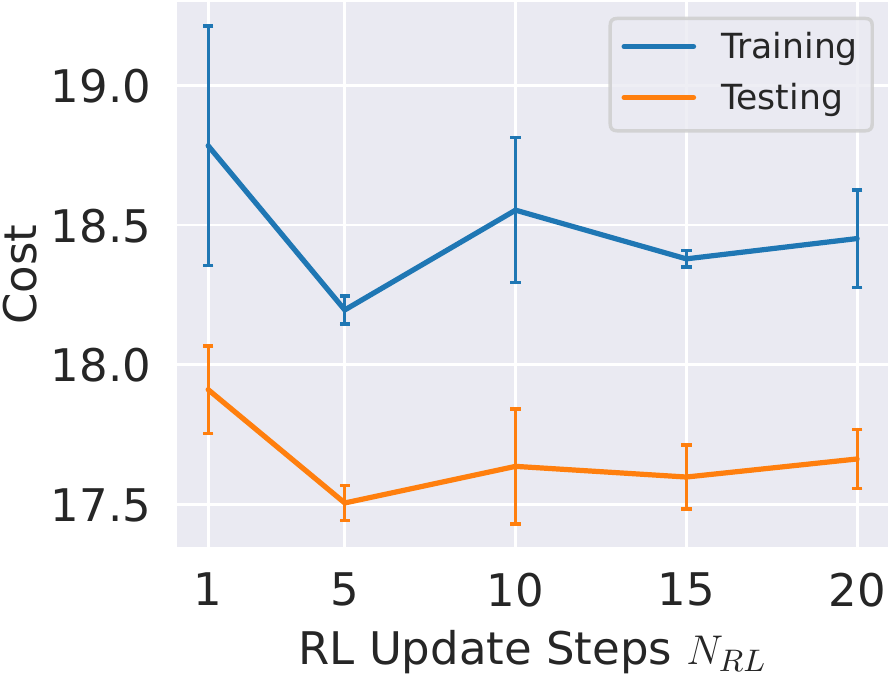}
    \caption{DLRM-20}
  \end{subfigure}%
  \begin{subfigure}[b]{0.25\textwidth}
    \centering
    \includegraphics[width=0.99\textwidth]{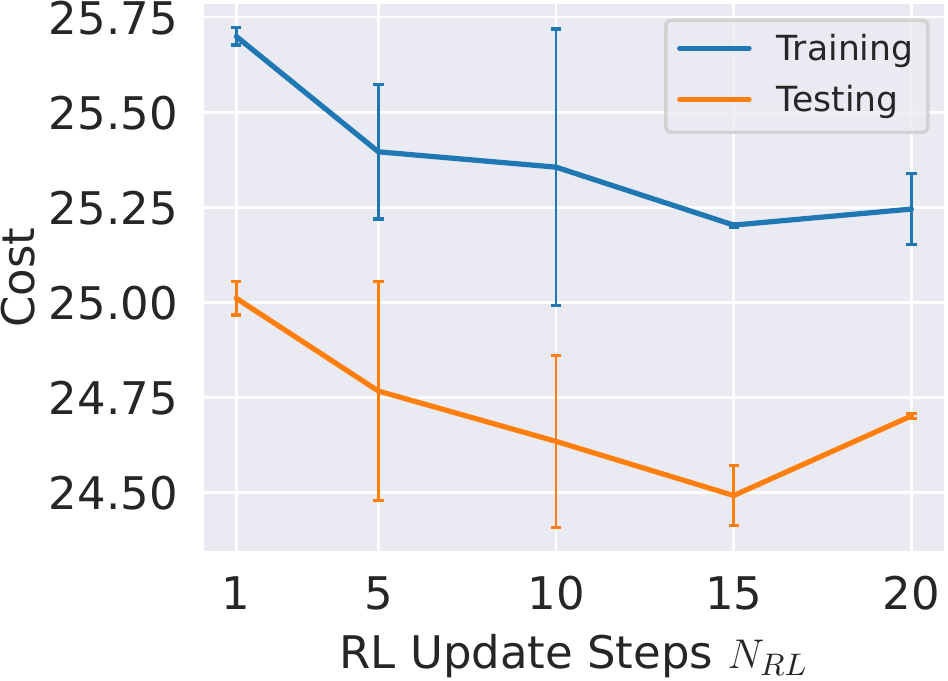}
    \caption{DLRM-30}
  \end{subfigure}%
  \begin{subfigure}[b]{0.25\textwidth}
    \centering
    \includegraphics[width=0.99\textwidth]{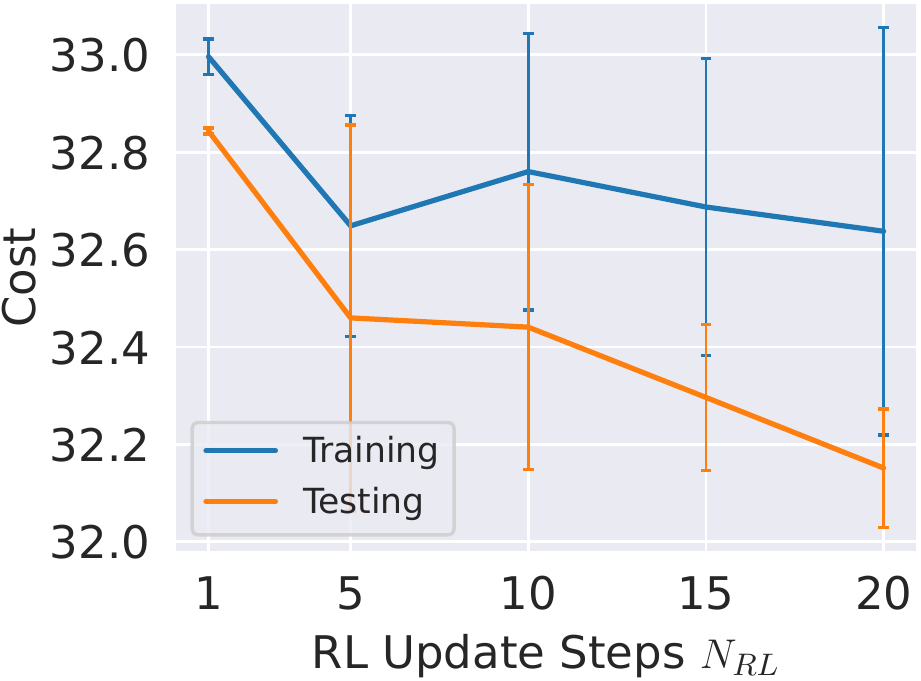}
    \caption{DLRM-40}
  \end{subfigure}%
  
  \begin{subfigure}[b]{0.25\textwidth}
    \centering
    \includegraphics[width=0.99\textwidth]{figs/rq4_1/50.pdf}
    \caption{DLRM-50}
  \end{subfigure}%
  \begin{subfigure}[b]{0.25\textwidth}
    \centering
    \includegraphics[width=0.99\textwidth]{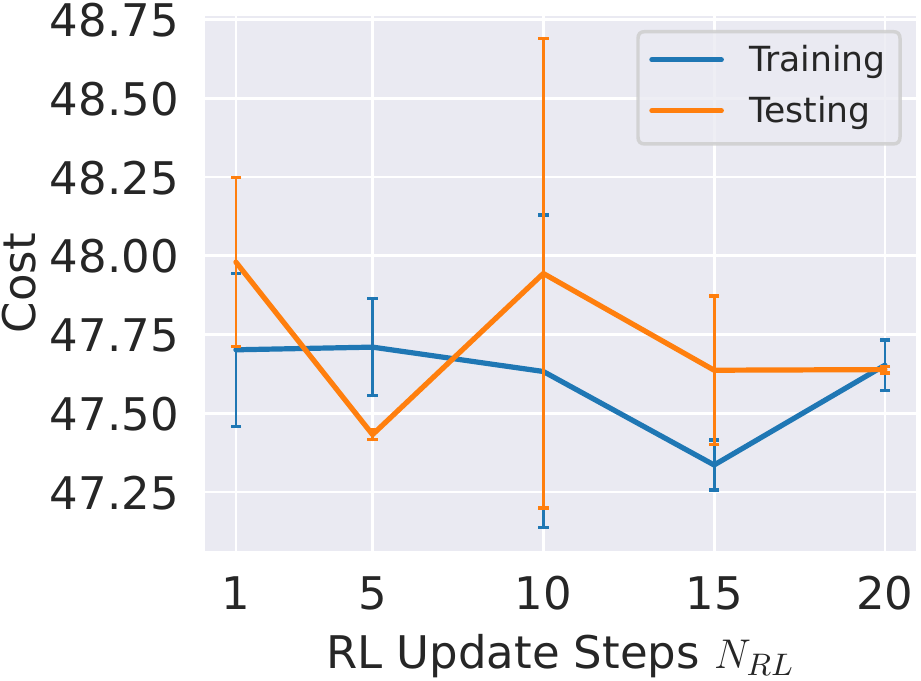}
    \caption{DLRM-60}
  \end{subfigure}%
  \begin{subfigure}[b]{0.25\textwidth}
    \centering
    \includegraphics[width=0.99\textwidth]{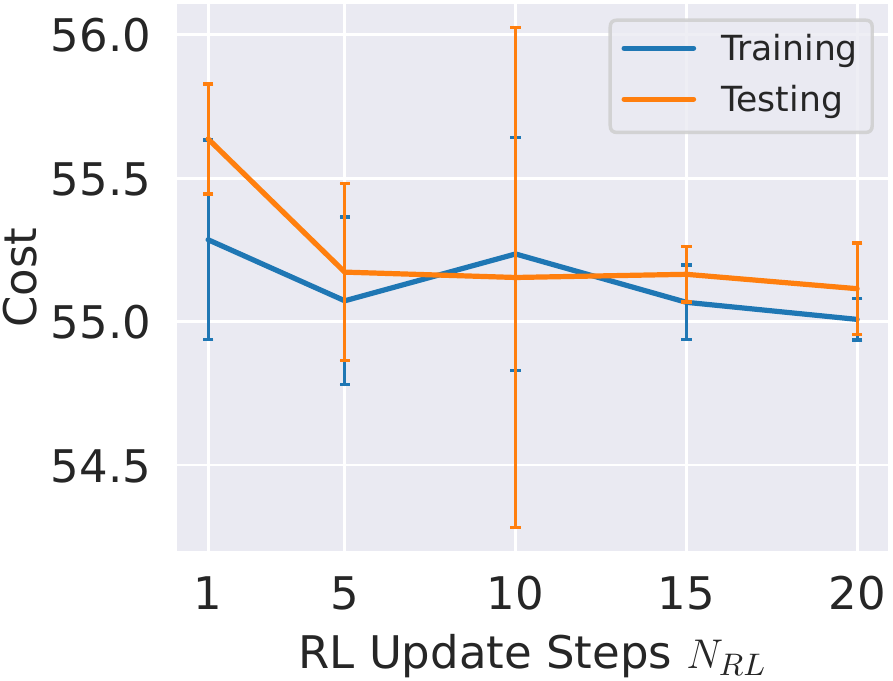}
    \caption{DLRM-70}
  \end{subfigure}%
  \begin{subfigure}[b]{0.25\textwidth}
    \centering
    \includegraphics[width=0.99\textwidth]{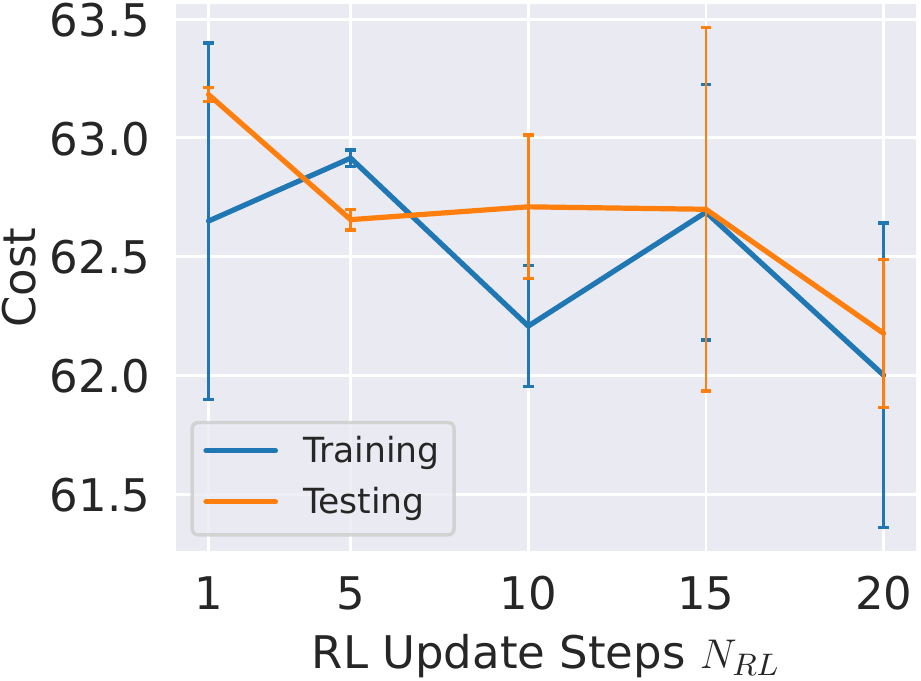}
    \caption{DLRM-80}
  \end{subfigure}%
  
  \begin{subfigure}[b]{0.25\textwidth}
    \centering
    \includegraphics[width=0.99\textwidth]{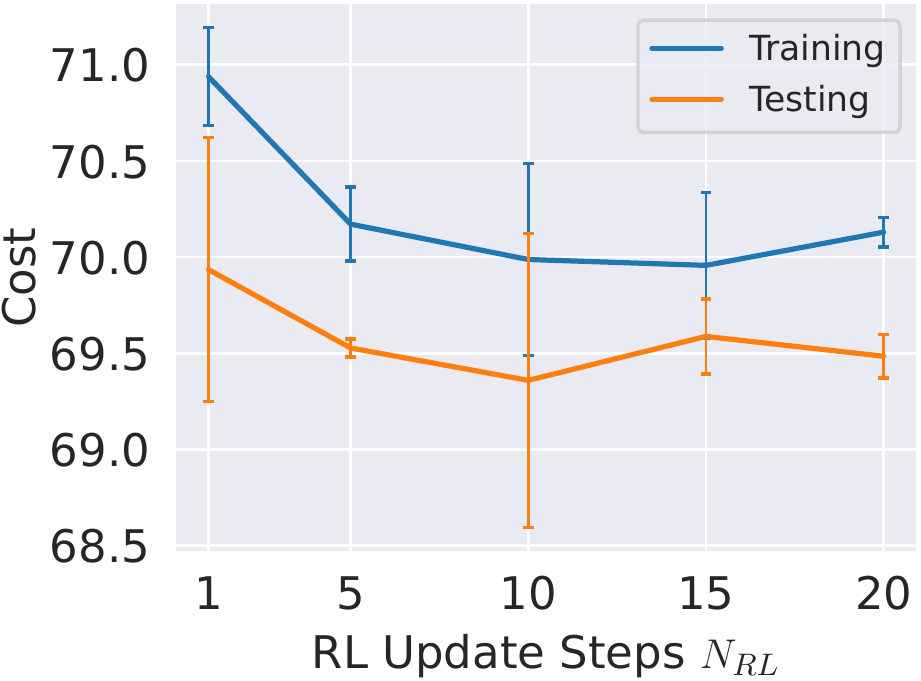}
    \caption{DLRM-90}
  \end{subfigure}%
  \begin{subfigure}[b]{0.25\textwidth}
    \centering
    \includegraphics[width=0.99\textwidth]{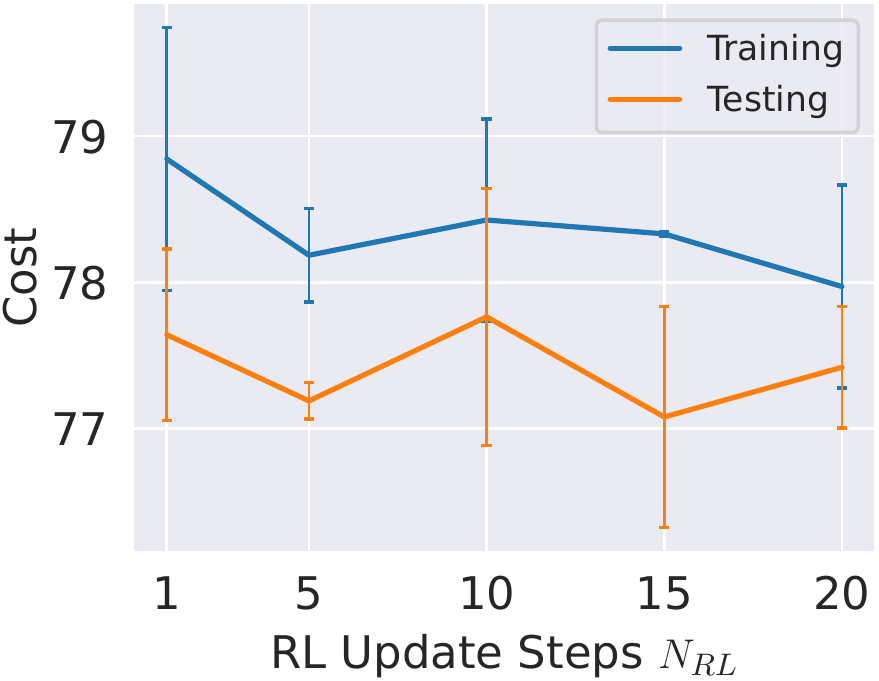}
    \caption{DLRM-100}
  \end{subfigure}%

  \caption{Impact of the number of RL update steps $N_{\text{RL}}$. In general, a too small $N_{\text{RL}}$ will degrade the performance. However, when $N_{\text{RL}} > 10$, increasing $N_{\text{RL}}$ does not lead to a clear benefit, but may cause more computation cost in training RL.}
\end{figure}

\begin{figure}[h!]
  \centering

  \begin{subfigure}[b]{0.25\textwidth}
    \centering
    \includegraphics[width=0.99\textwidth]{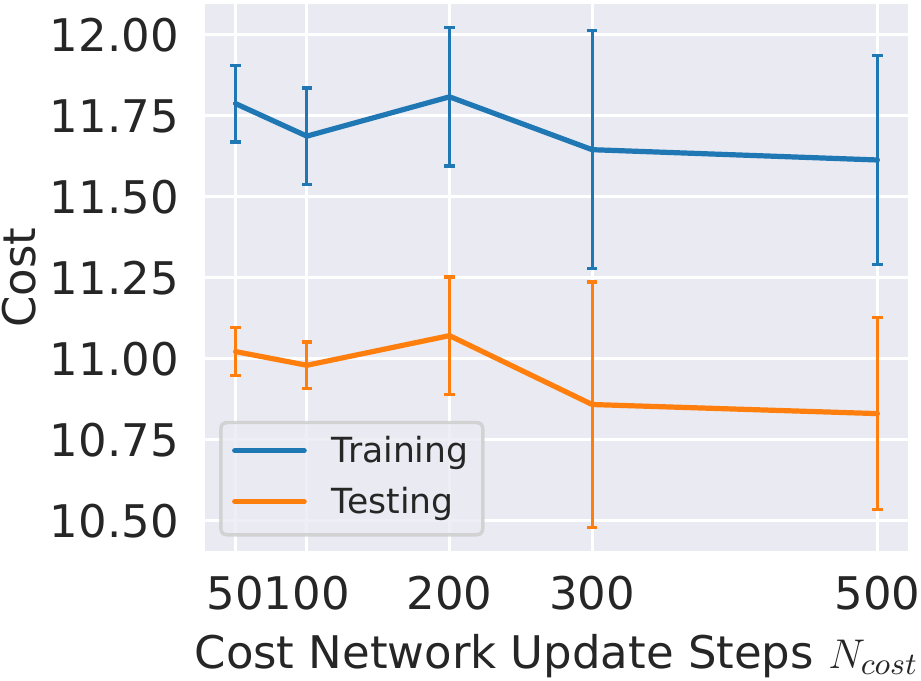}
    \caption{DLRM-10}
  \end{subfigure}%
  \begin{subfigure}[b]{0.25\textwidth}
    \centering
    \includegraphics[width=0.99\textwidth]{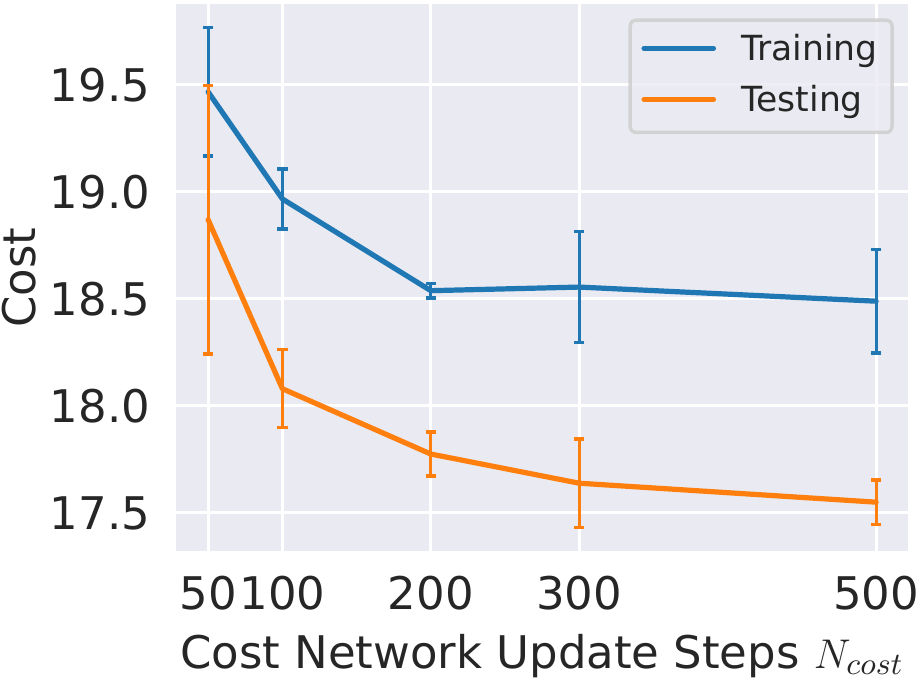}
    \caption{DLRM-20}
  \end{subfigure}%
  \begin{subfigure}[b]{0.25\textwidth}
    \centering
    \includegraphics[width=0.99\textwidth]{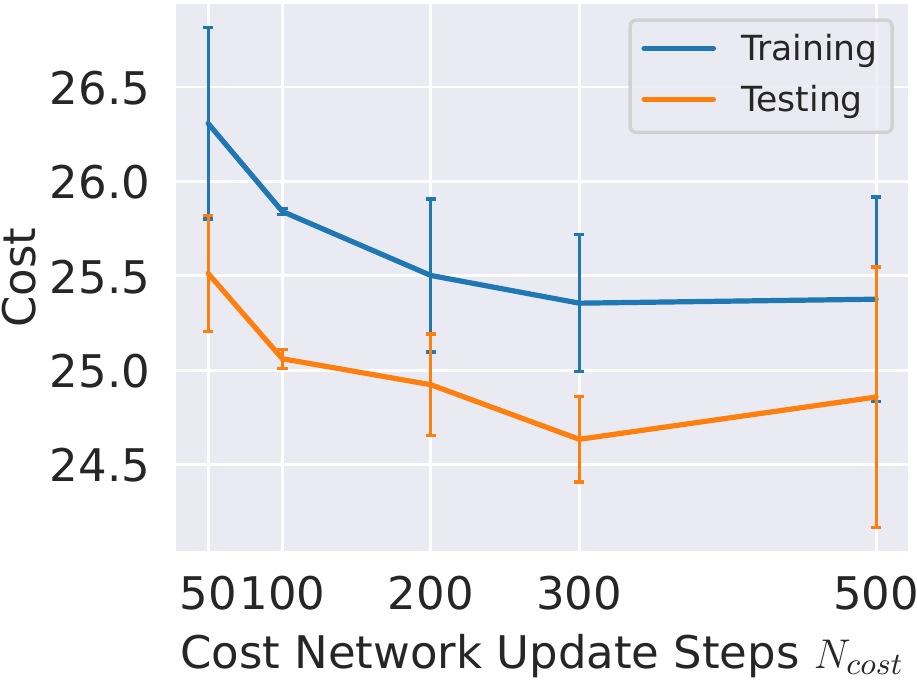}
    \caption{DLRM-30}
  \end{subfigure}%
  \begin{subfigure}[b]{0.25\textwidth}
    \centering
    \includegraphics[width=0.99\textwidth]{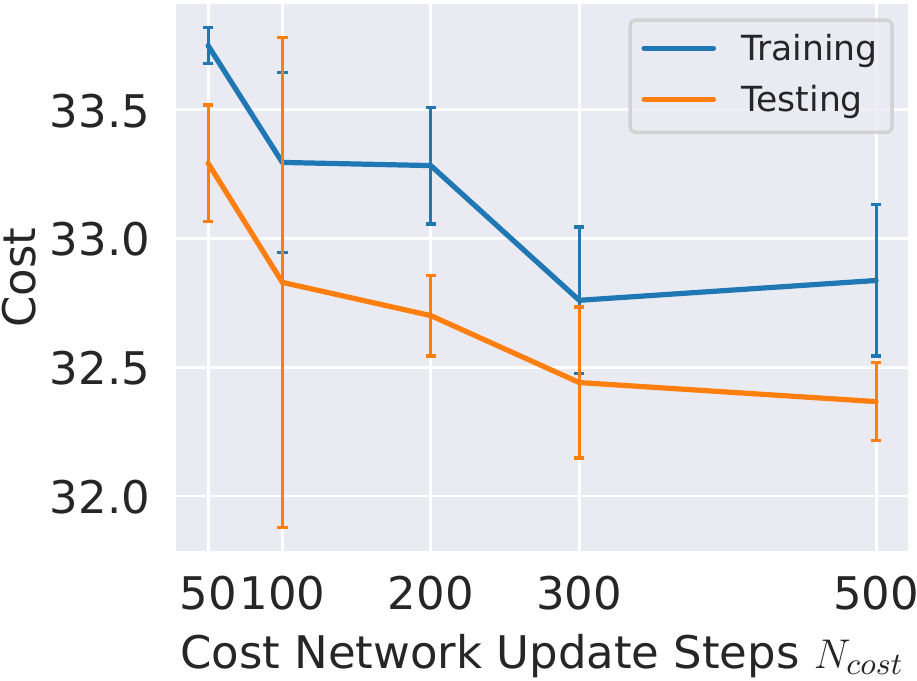}
    \caption{DLRM-40}
  \end{subfigure}%
  
  \begin{subfigure}[b]{0.25\textwidth}
    \centering
    \includegraphics[width=0.99\textwidth]{figs/rq4_2/50.pdf}
    \caption{DLRM-50}
  \end{subfigure}%
  \begin{subfigure}[b]{0.25\textwidth}
    \centering
    \includegraphics[width=0.99\textwidth]{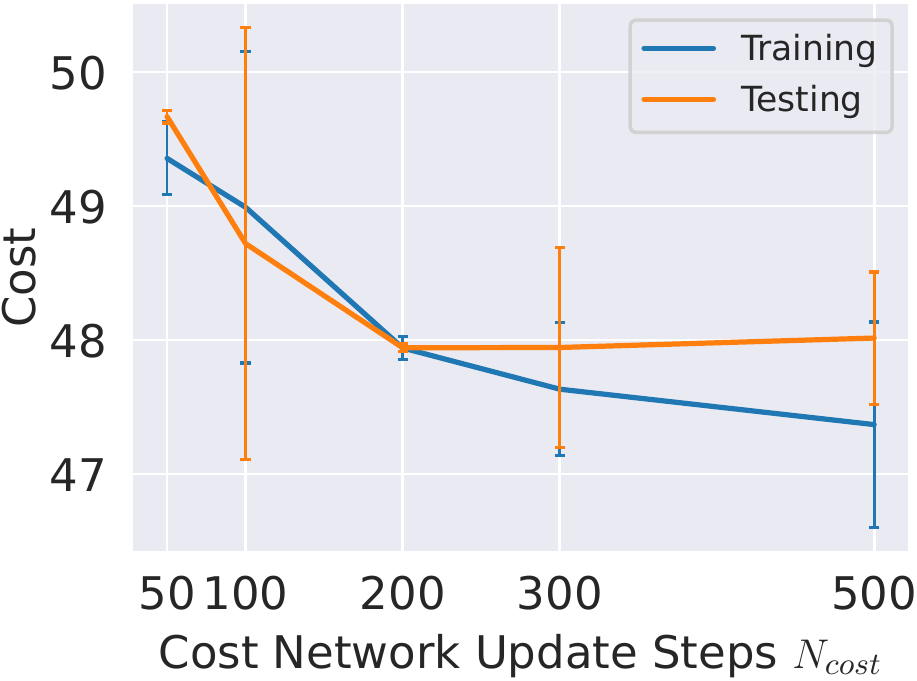}
    \caption{DLRM-60}
  \end{subfigure}%
  \begin{subfigure}[b]{0.25\textwidth}
    \centering
    \includegraphics[width=0.99\textwidth]{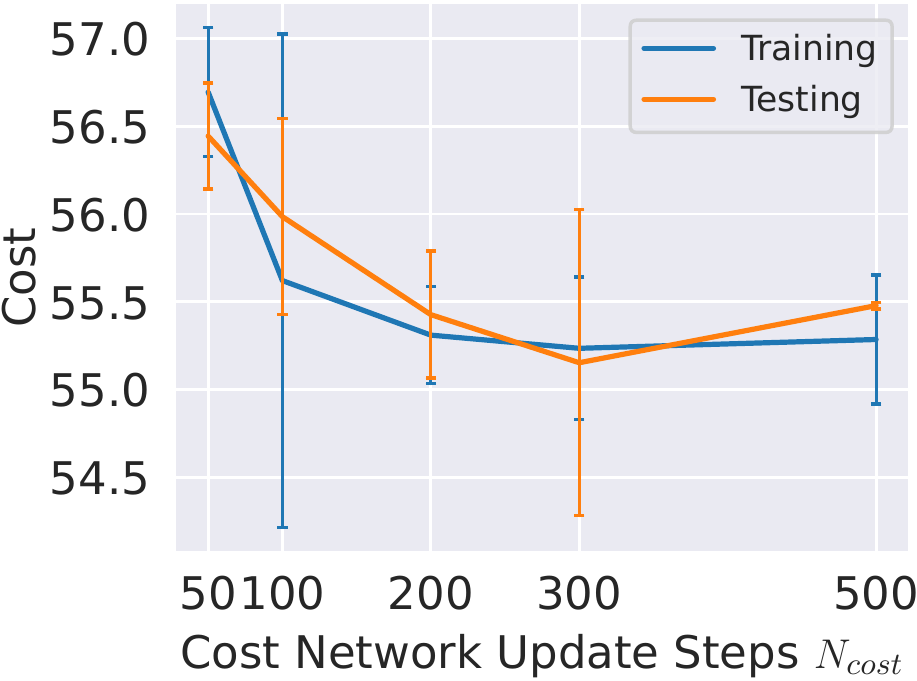}
    \caption{DLRM-70}
  \end{subfigure}%
  \begin{subfigure}[b]{0.25\textwidth}
    \centering
    \includegraphics[width=0.99\textwidth]{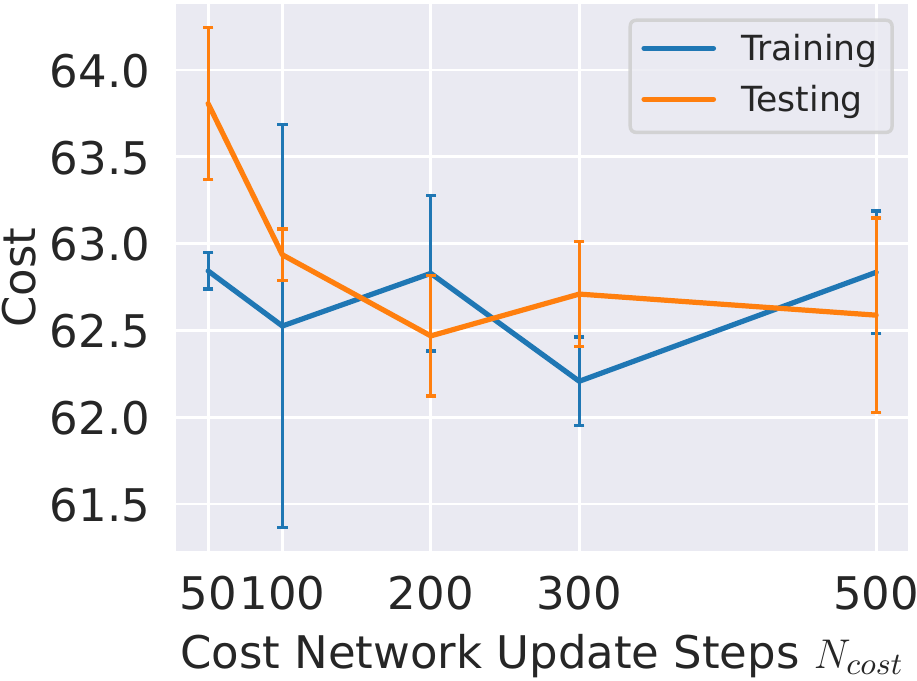}
    \caption{DLRM-80}
  \end{subfigure}%
  
  \begin{subfigure}[b]{0.25\textwidth}
    \centering
    \includegraphics[width=0.99\textwidth]{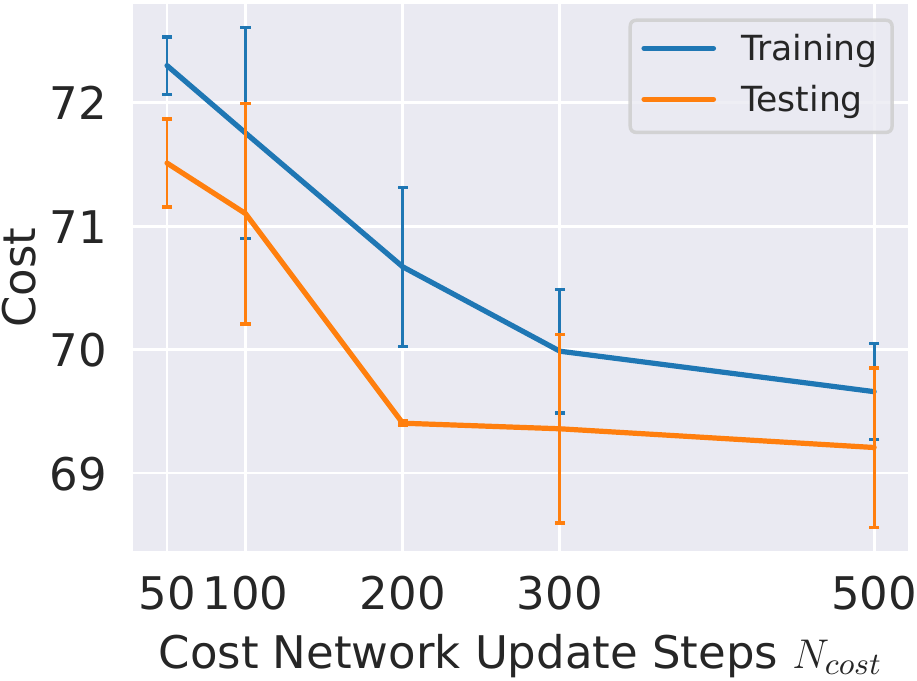}
    \caption{DLRM-90}
  \end{subfigure}%
  \begin{subfigure}[b]{0.25\textwidth}
    \centering
    \includegraphics[width=0.99\textwidth]{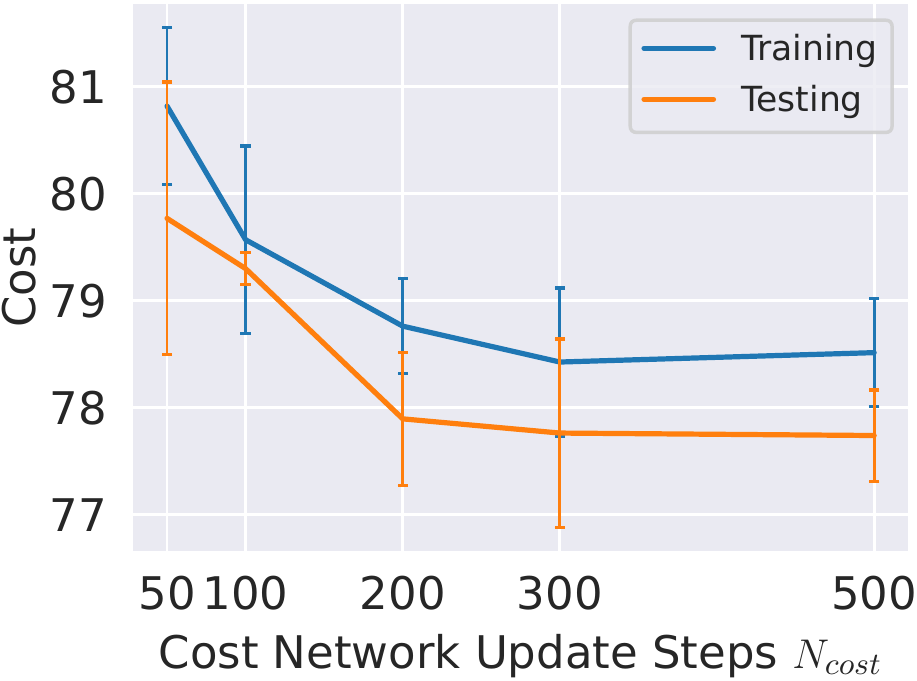}
    \caption{DLRM-100}
  \end{subfigure}%

  \caption{Impact of the cost network update steps $N_{\text{cost}}$. In general, a too small $N_{\text{cost}}$ will degrade the performance. However, when $N_{\text{cost}} > 300$, increasing $N_{\text{cost}}$ does not lead to a clear benefit, but may cause more computation cost in training the cost network.}
\end{figure}

\newpage
\section{Additional Results of Ablation Study}
\label{appendix:J}

\begin{table}[h!]
\centering
\caption{Ablation Study. The pooling factor and the dimension appear to be the most important table features. The proposed cost features are very effective since w/o cost has poor performances. }
\scriptsize
\setlength{\tabcolsep}{1.0pt}
\begin{tabular}{l|c|c|c|c|c|c|c|c|c}
\toprule

\multicolumn{2}{c|}{Task} & w/o dim & w/o row & w/o pooling factor & w/o table size & w/o distribution & w/o cost & w/ RNN & DreamShard  \\
\midrule
\multirow{2}{*}{DLRM-10 (4)} & Train & 11.8$\pm$0.2 & \textbf{11.5$\pm$0.1} & 12.9$\pm$0.2 & 11.5$\pm$0.0 & 11.8$\pm$0.3 & 13.4$\pm$0.6 &  11.7$\pm$0.1 & 11.6$\pm$0.3  \\
~ & Test & 10.9$\pm$0.0 & \textbf{10.7$\pm$0.0} & 12.4$\pm$0.2 & \textbf{10.7$\pm$0.0} & 11.0$\pm$0.4 & 12.4$\pm$0.6 & 10.8$\pm$0.0 & 10.9$\pm$0.3  \\
\midrule
\multirow{2}{*}{DLRM-20 (4)} & Train & 18.5$\pm$0.1 & 18.6$\pm$0.3 & 21.6$\pm$0.1 & \textbf{18.3$\pm$0.1} & 18.8$\pm$0.2 & 22.0$\pm$0.1 & 18.7$\pm$0.2 & 18.6$\pm$0.2  \\
~ & Test & 17.7$\pm$0.2 & 17.7$\pm$0.3 & 20.4$\pm$0.2 & \textbf{17.5$\pm$0.1} & 18.0$\pm$0.0 & 20.8$\pm$0.3 & 17.8$\pm$0.2 & 17.6$\pm$0.2  \\
\midrule
\multirow{2}{*}{DLRM-30 (4)} & Train & \textbf{25.3$\pm$0.2} & 25.5$\pm$0.1 & 29.6$\pm$0.2 & 25.2$\pm$0.1 & 25.7$\pm$0.4 & 29.8$\pm$0.5 & 25.2$\pm$0.2 & 25.4$\pm$0.3  \\
~ & Test & 24.7$\pm$0.3 & 24.9$\pm$0.1 & 29.2$\pm$0.4 & \textbf{24.5$\pm$0.1} & 24.9$\pm$0.2 & 29.0$\pm$0.6 & 24.7$\pm$0.3 & 24.6$\pm$0.2  \\
\midrule
\multirow{2}{*}{DLRM-40 (4)} & Train & 33.1$\pm$0.8 & 32.9$\pm$0.3 & 37.9$\pm$0.3 & 33.2$\pm$0.4 & 32.9$\pm$0.1 & 38.1$\pm$0.2 & \textbf{32.6$\pm$0.2} & 32.8$\pm$0.3  \\
~ & Test & 33.3$\pm$0.8 & \textbf{32.4$\pm$0.2} & 37.9$\pm$0.6 & 32.9$\pm$0.4 & 32.5$\pm$0.2 & 37.2$\pm$0.1 & \textbf{32.3$\pm$0.1} & 32.4$\pm$0.3  \\
\midrule
\multirow{2}{*}{DLRM-50 (4)} & Train & 40.8$\pm$0.4 & 40.7$\pm$0.1 & 46.3$\pm$0.3 & 40.8$\pm$0.4 & 40.6$\pm$0.2 & 47.5$\pm$1.2 & 40.5$\pm$0.2 & \textbf{40.4$\pm$0.5}  \\
~ & Test & 40.9$\pm$0.6 & 40.6$\pm$0.3 & 47.2$\pm$0.1 & 40.6$\pm$0.7 & 40.5$\pm$0.2 & 46.3$\pm$0.1 & 40.5$\pm$0.1 & \textbf{40.4$\pm$0.6}  \\
\midrule
\multirow{2}{*}{DLRM-60 (4)} & Train & 48.5$\pm$0.7 & 47.6$\pm$0.4 & 54.3$\pm$0.2 & 47.8$\pm$0.3 & 48.0$\pm$0.1 & 53.9$\pm$1.3 & \textbf{47.5$\pm$0.0} & 47.6$\pm$0.4  \\
~ & Test & 48.9$\pm$0.5 & \textbf{47.7}$\pm$0.4 & 54.8$\pm$0.3 & 48.0$\pm$0.3 & 48.1$\pm$0.3 & 54.7$\pm$0.8 & \textbf{47.7$\pm$0.1} & 47.9$\pm$0.7  \\
\midrule
\multirow{2}{*}{DLRM-70 (4)} & Train & 56.0$\pm$0.5 & \textbf{55.2$\pm$0.1} & 62.9$\pm$0.2 & 55.5$\pm$0.2 & 55.3$\pm$0.1 & 62.5$\pm$0.6 & \textbf{55.0$\pm$0.1} & 55.2$\pm$0.4  \\
~ & Test & 56.1$\pm$0.2 & 55.5$\pm$0.2 & 62.8$\pm$0.6 & 55.5$\pm$0.2 & \textbf{55.6$\pm$0.0} & 58.3$\pm$0.5 & 55.0$\pm$0.0 & 55.2$\pm$0.8  \\
\midrule
\multirow{2}{*}{DLRM-80 (4)} & Train & 64.2$\pm$0.6 & 62.8$\pm$0.1 & 70.3$\pm$0.7 & 62.6$\pm$0.1 & 62.9$\pm$0.1 & 71.4$\pm$0.7 & 62.5$\pm$0.2 & \textbf{62.2$\pm$0.2}  \\
~ & Test & 64.2$\pm$0.6 & 62.6$\pm$0.1 & 71.0$\pm$1.1 & 62.9$\pm$0.0 & 63.0$\pm$0.4 & 71.1$\pm$1.4 & \textbf{62.1$\pm$0.4} & 62.7$\pm$0.3  \\
\midrule
\multirow{2}{*}{DLRM-90 (4)} & Train & 71.8$\pm$1.1 & 71.0$\pm$0.7 & 79.1$\pm$1.0 & 70.7$\pm$0.1 & 70.4$\pm$0.4 & 79.8$\pm$0.9 & 70.8$\pm$0.3 & \textbf{70.0$\pm$0.4}  \\
~ & Test & 70.8$\pm$1.2 & 70.3$\pm$0.8 & 77.5$\pm$0.4 & 69.6$\pm$0.3 & 70.1$\pm$0.0 & 79.0$\pm$1.3 & 70.2$\pm$0.1 & \textbf{69.4$\pm$0.7}  \\
\midrule
\multirow{2}{*}{DLRM-100 (4)} & Train & 79.6$\pm$0.3 & 79.1$\pm$0.6 & 87.6$\pm$0.6 & 78.7$\pm$0.2 & 78.9$\pm$0.1 & 89.1$\pm$2.8 & 78.6$\pm$0.6 & \textbf{78.4$\pm$0.6}  \\
~ & Test & 78.1$\pm$0.8 & 78.1$\pm$0.5 & 86.2$\pm$0.8 & 78.0$\pm$0.4 & 78.0$\pm$0.2 & 87.7$\pm$1.5 & \textbf{77.8$\pm$0.6} & \textbf{77.8$\pm$0.8}  \\

\bottomrule
\end{tabular}
\end{table}

\begin{table}[h!]
\centering
\caption{Ablation study of the table features on the Prod dataset. We did not use the DLRM dataset because its tables have the same table dimension, which could make the prediction less dependent on the table dimension. We collect a million samples and split 80\%/10\%/10\% as training/validation/testing sets. We fully train a cost network with 100 epochs and report the MSE on the testing set with each individual feature being removed. We find that each table feature contributes to the prediction accuracy, and the most contributing features are table dimension, pooling factor, and distribution features.}
\scriptsize
\setlength{\tabcolsep}{5.0pt}
\begin{tabular}{l|c}
\toprule
Features & Testing MSE \\
\midrule
w/o dimension & 13.746 \\
w/o hash size & 0.307 \\
w/o pooling factor & 0.635 \\
w/o table size & 0.305 \\
w/o distribution features & 0.437\\
All features & \textbf{0.303} \\

\bottomrule
\end{tabular}
\end{table}

\section{Additional Results on Ultra-Large
Industrial Recommendation Model}

\begin{table}[h!]
\centering
\caption{Scalability test. We apply each placement algorithm (excluding the RNN-based method since we find it is very unstable and can not deliver a reasonable performance) to an ultra-large industrial recommendation model, which contains nearly a thousand embedding tables that demand multi-terabyte memory. We run all the placement algorithms on a training cluster with 128 GPUs. We measure the embedding cost and the overall training throughput, which includes embedding lookup, dense computation, data loading, etc. We report the relative improvement over random placement. Since the production model has already been optimized with many iterations, a 5\% improvement of training throughput is considered significant. DreamShard shows 27.6\% improvement over the strongest baseline. }
\scriptsize
\setlength{\tabcolsep}{5.0pt}
\begin{tabular}{l|c|c}
\toprule
Sharding Algorithm & Embedding cost & Training Throughput Improvement \\
\midrule
Random & 118.3 & 0.00\% \\
Size-based & 107.6 (+10.0\%) & +4.0\% \\
Dim-based & 90.8 (+30.3\%) & +13.9\% \\
Lookup-based & 102.4 (+15.6\%) & +11.9\% \\
Size-lookup-based & 109.2 (+8.3\%) & +12.8\% \\
DreamShard & 61.59 (+92.2\%) & +45.3\%\\

\bottomrule
\end{tabular}
\end{table}

\newpage
\section{Additional Good/Bad Case Study on Tasks with 50 Tables and 4 GPUs}
\label{appendix:K}

\begin{figure}[h!]
  \centering
  \begin{subfigure}[b]{0.8\textwidth}
    \centering
    \includegraphics[width=0.99\textwidth]{figs/case/legend.pdf}
  \end{subfigure}%
  
  \begin{subfigure}[b]{0.8\textwidth}
    \centering
    \includegraphics[width=0.99\textwidth]{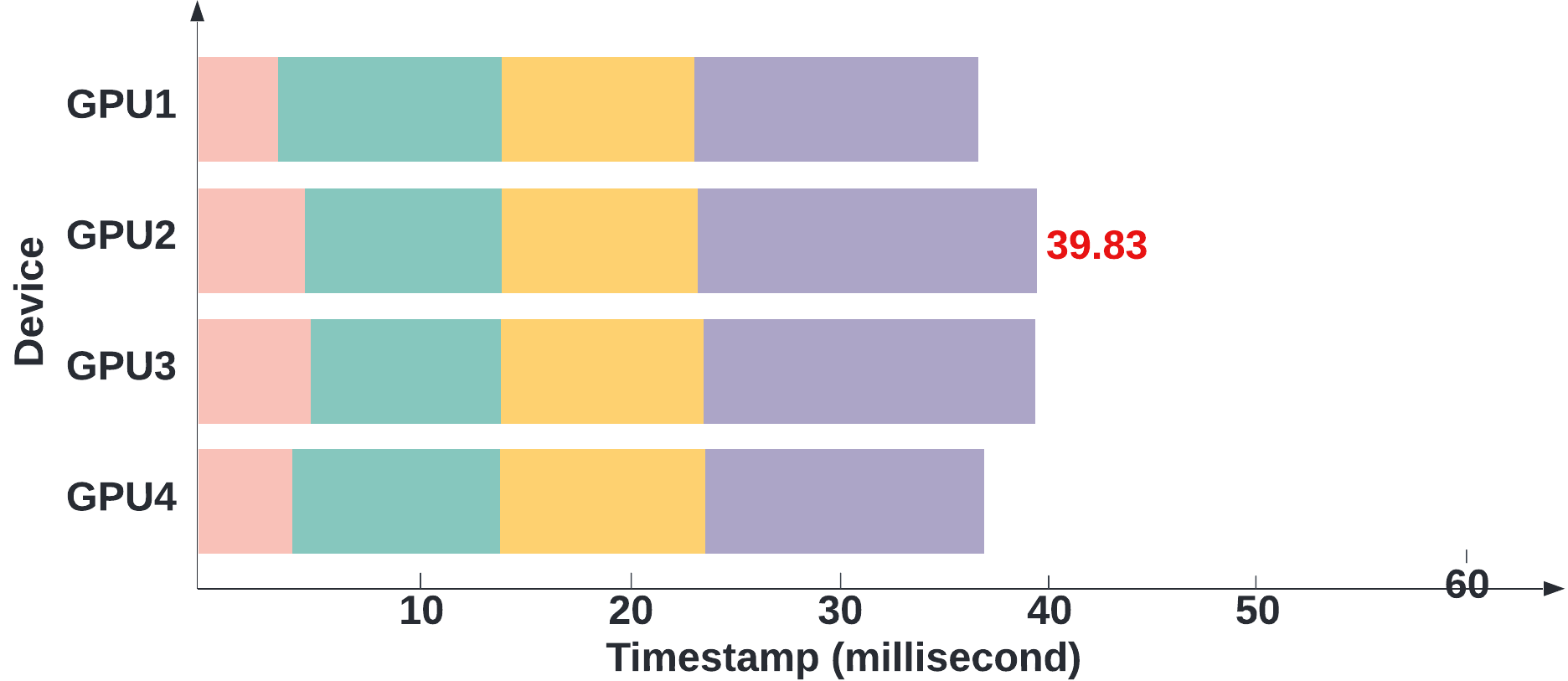}
    \subcaption{Random placement}
  \end{subfigure}%
  
  \begin{subfigure}[b]{0.8\textwidth}
    \centering
    \includegraphics[width=0.99\textwidth]{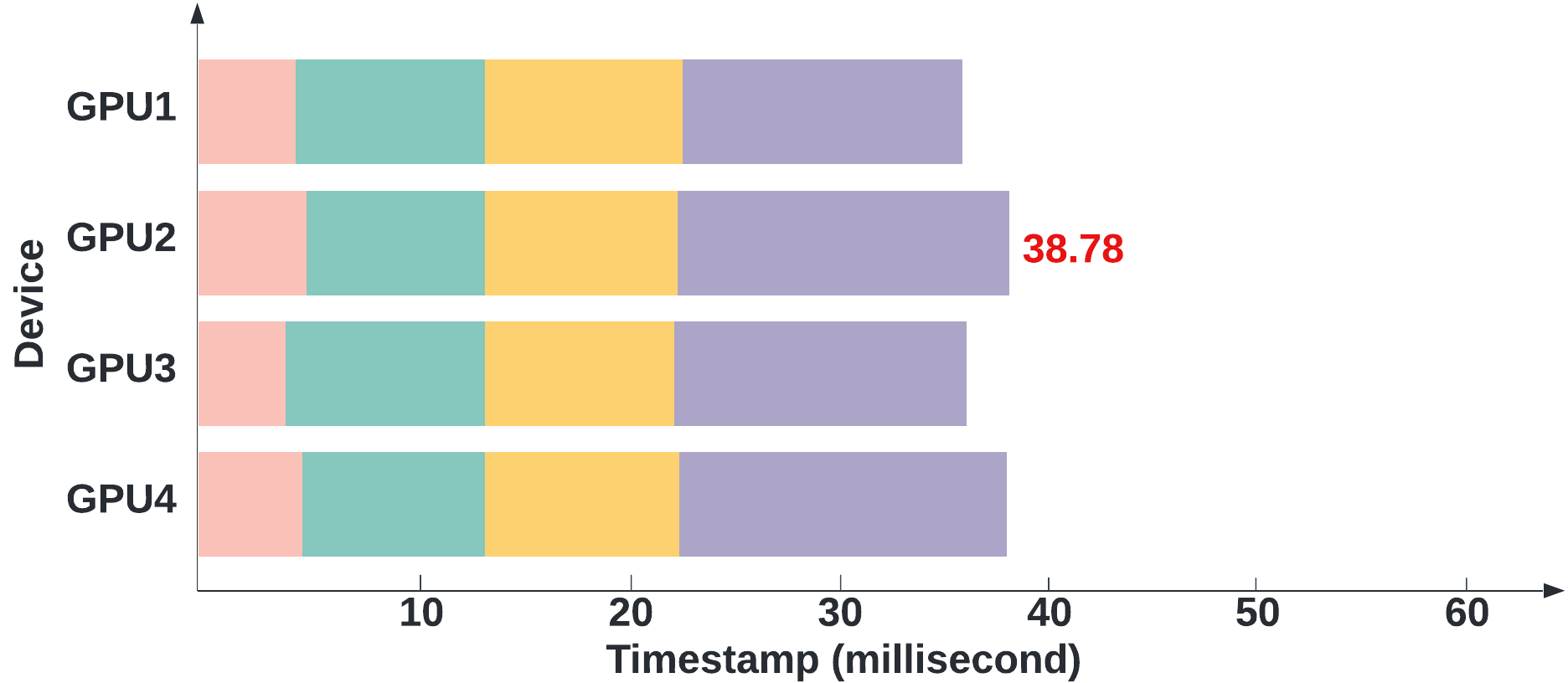}
    \subcaption{Best human expert placement}
  \end{subfigure}%
  
  \begin{subfigure}[b]{0.8\textwidth}
    \centering
    \includegraphics[width=0.99\textwidth]{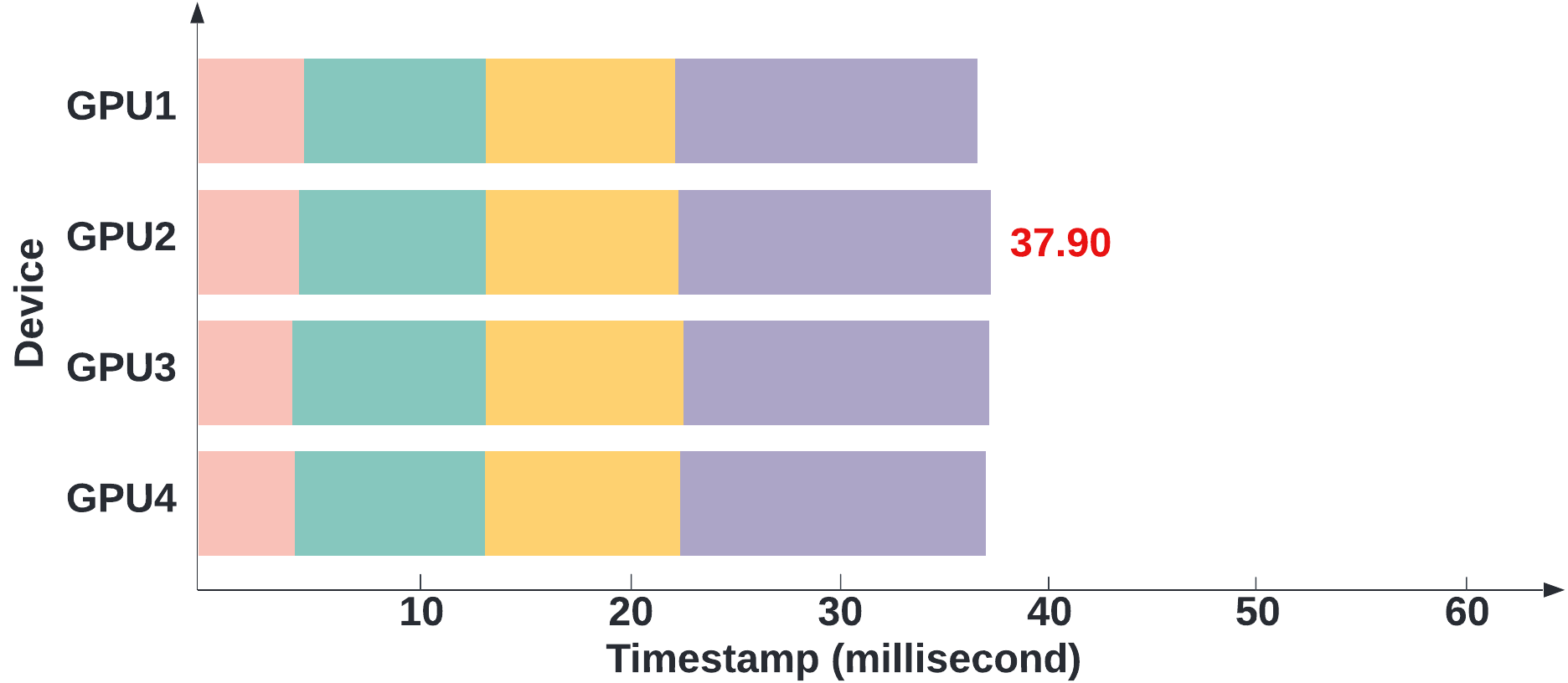}
    \subcaption{DreamShard}
  \end{subfigure}%
 
  \caption{Good case 1: visualization of DreamShard, the best heuristic algorithm, and random placement on a task of placing 50 tables to 4 GPUs. DreamShard outperforms the baselines with a better balance. }
\end{figure}

\begin{figure}[h!]
  \centering
  \begin{subfigure}[b]{0.8\textwidth}
    \centering
    \includegraphics[width=0.99\textwidth]{figs/case/legend.pdf}
  \end{subfigure}%
  
  \begin{subfigure}[b]{0.8\textwidth}
    \centering
    \includegraphics[width=0.99\textwidth]{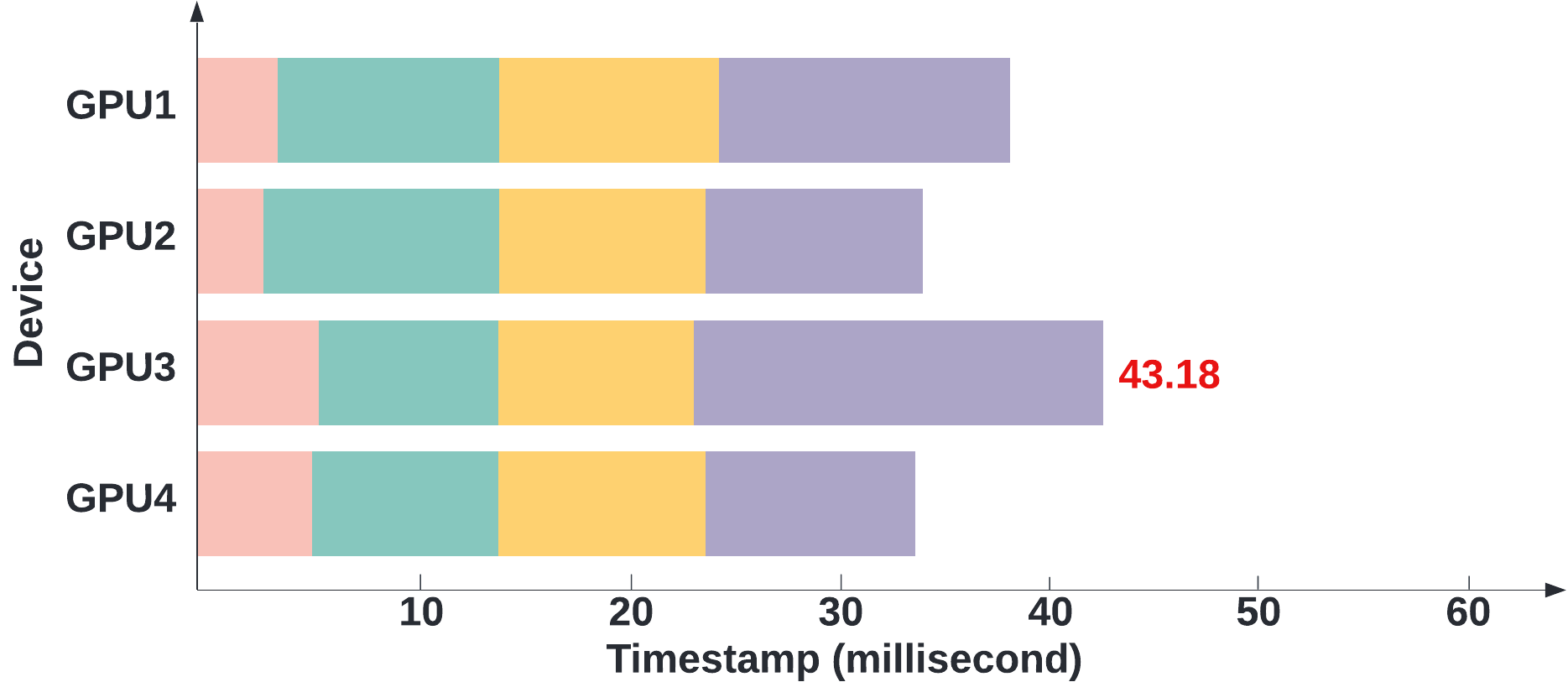}
    \subcaption{Random placement}
  \end{subfigure}%
  
  \begin{subfigure}[b]{0.8\textwidth}
    \centering
    \includegraphics[width=0.99\textwidth]{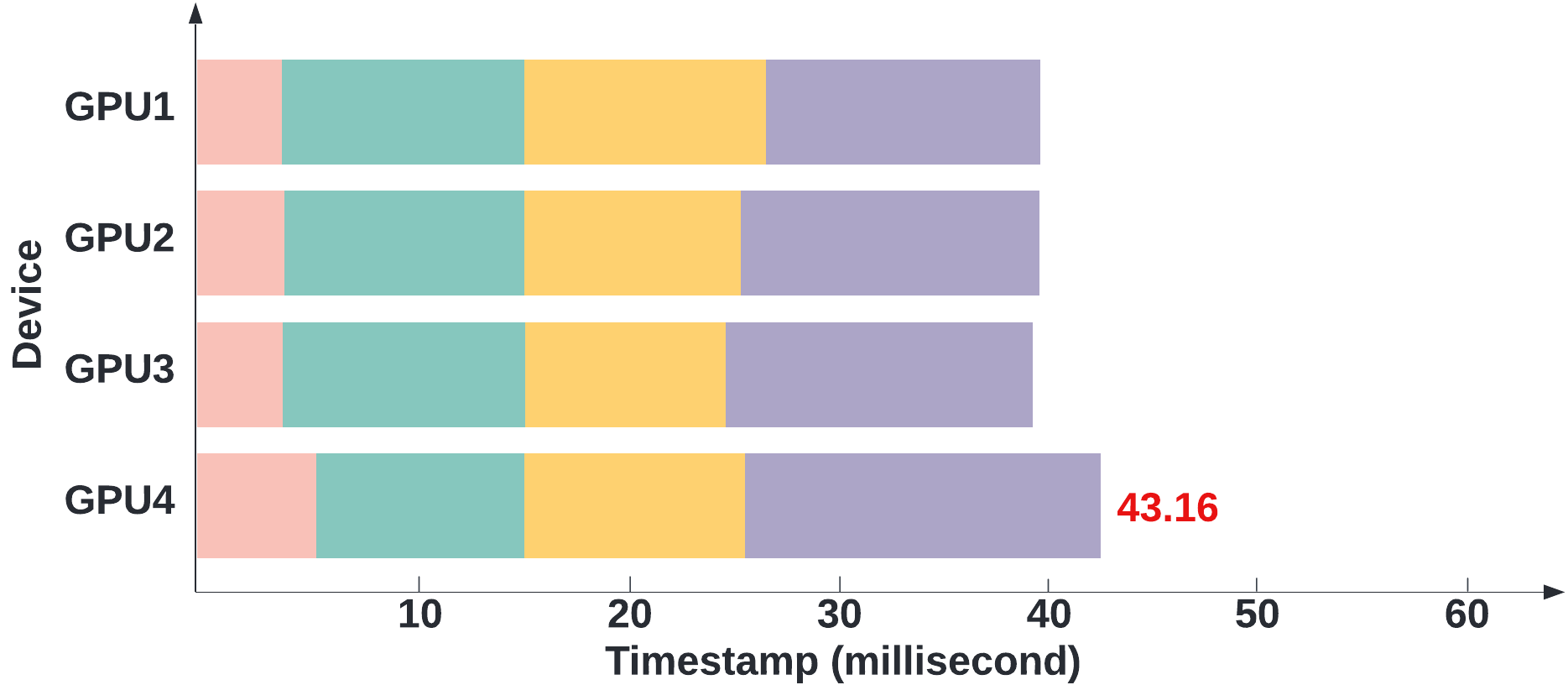}
    \subcaption{Best human expert placement}
  \end{subfigure}%
  
  \begin{subfigure}[b]{0.8\textwidth}
    \centering
    \includegraphics[width=0.99\textwidth]{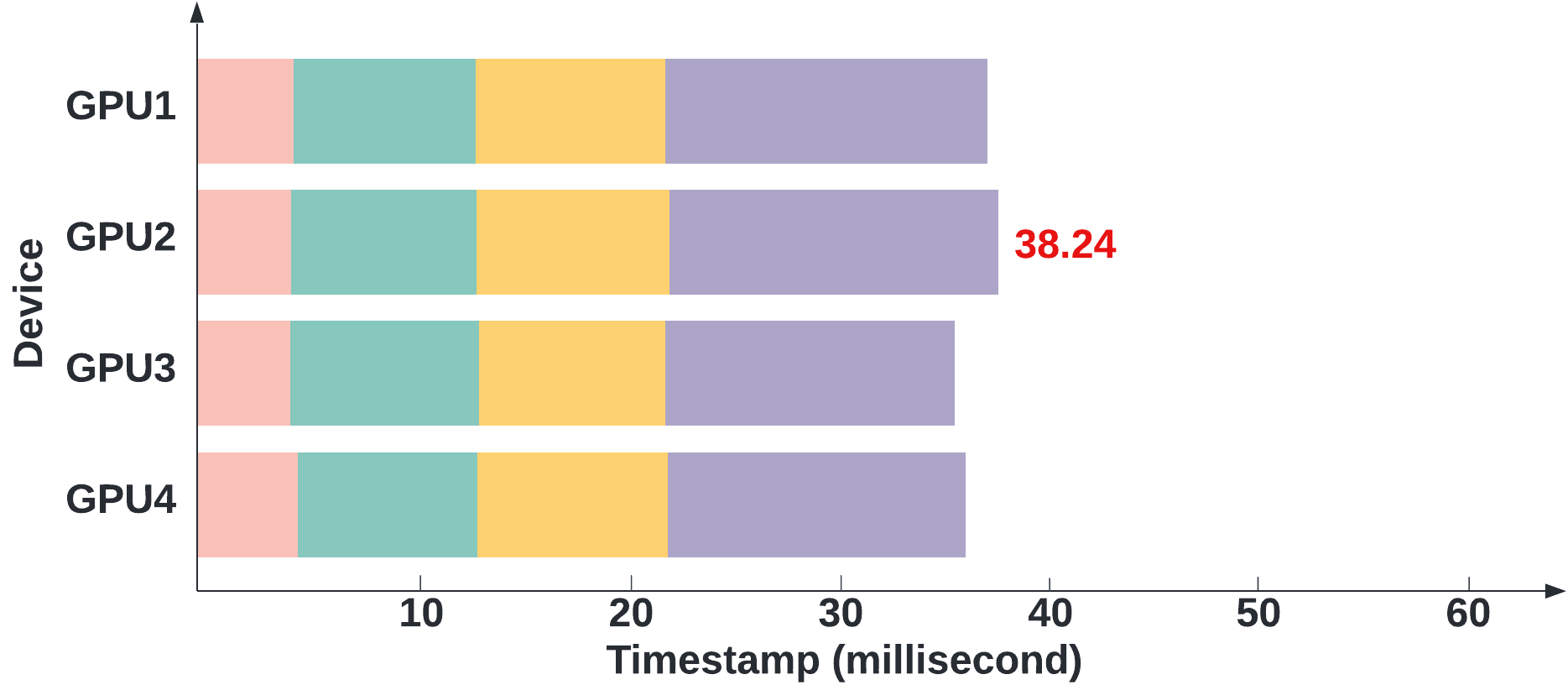}
    \subcaption{DreamShard}
  \end{subfigure}%
 
  \caption{Good case 2: visualization of DreamShard, the best heuristic algorithm, and random placement on a task of placing 50 tables to 4 GPUs. DreamShard achieves better balance as well as less communication time, leading to significantly lower overall cost.}
\end{figure}

\begin{figure}[h!]
  \centering
  \begin{subfigure}[b]{0.8\textwidth}
    \centering
    \includegraphics[width=0.99\textwidth]{figs/case/legend.pdf}
  \end{subfigure}%
  
  \begin{subfigure}[b]{0.8\textwidth}
    \centering
    \includegraphics[width=0.99\textwidth]{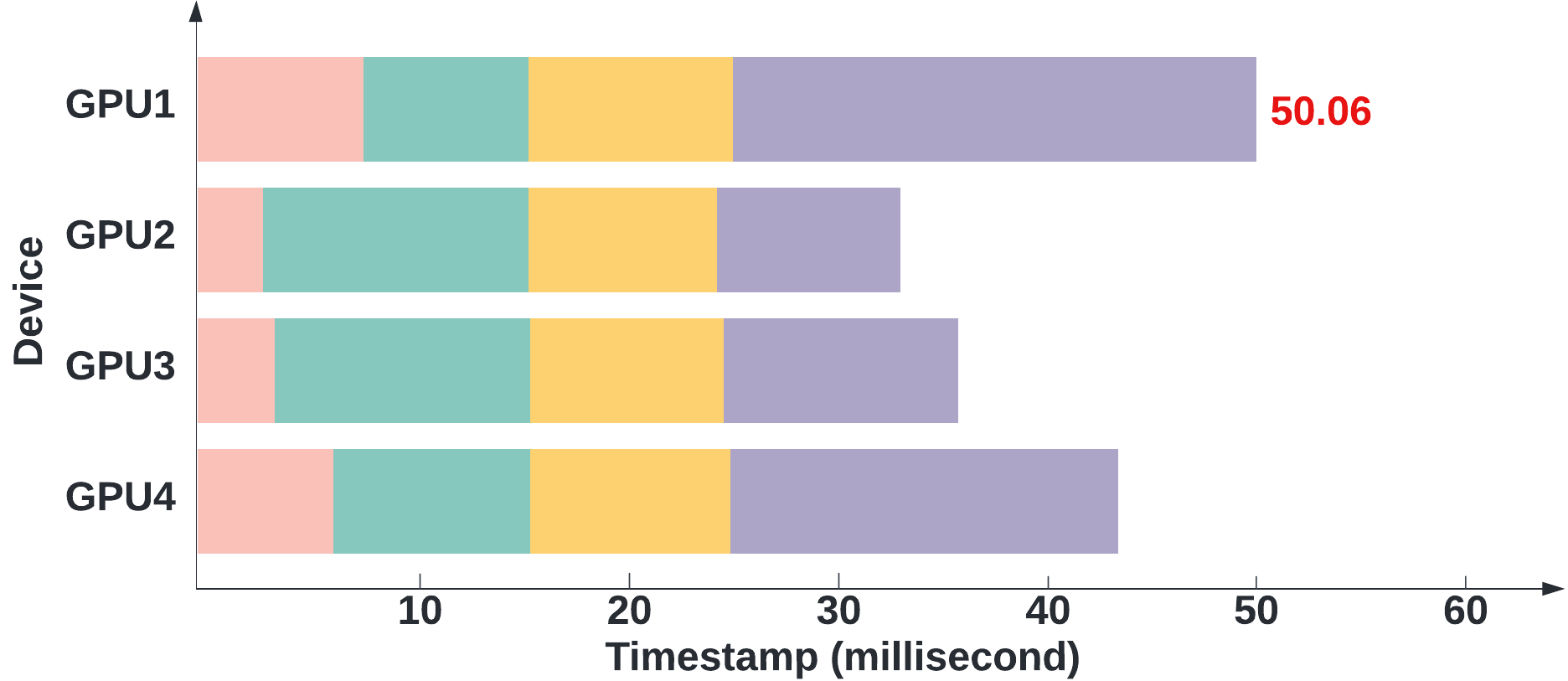}
    \subcaption{Random placement}
  \end{subfigure}%
  
  \begin{subfigure}[b]{0.8\textwidth}
    \centering
    \includegraphics[width=0.99\textwidth]{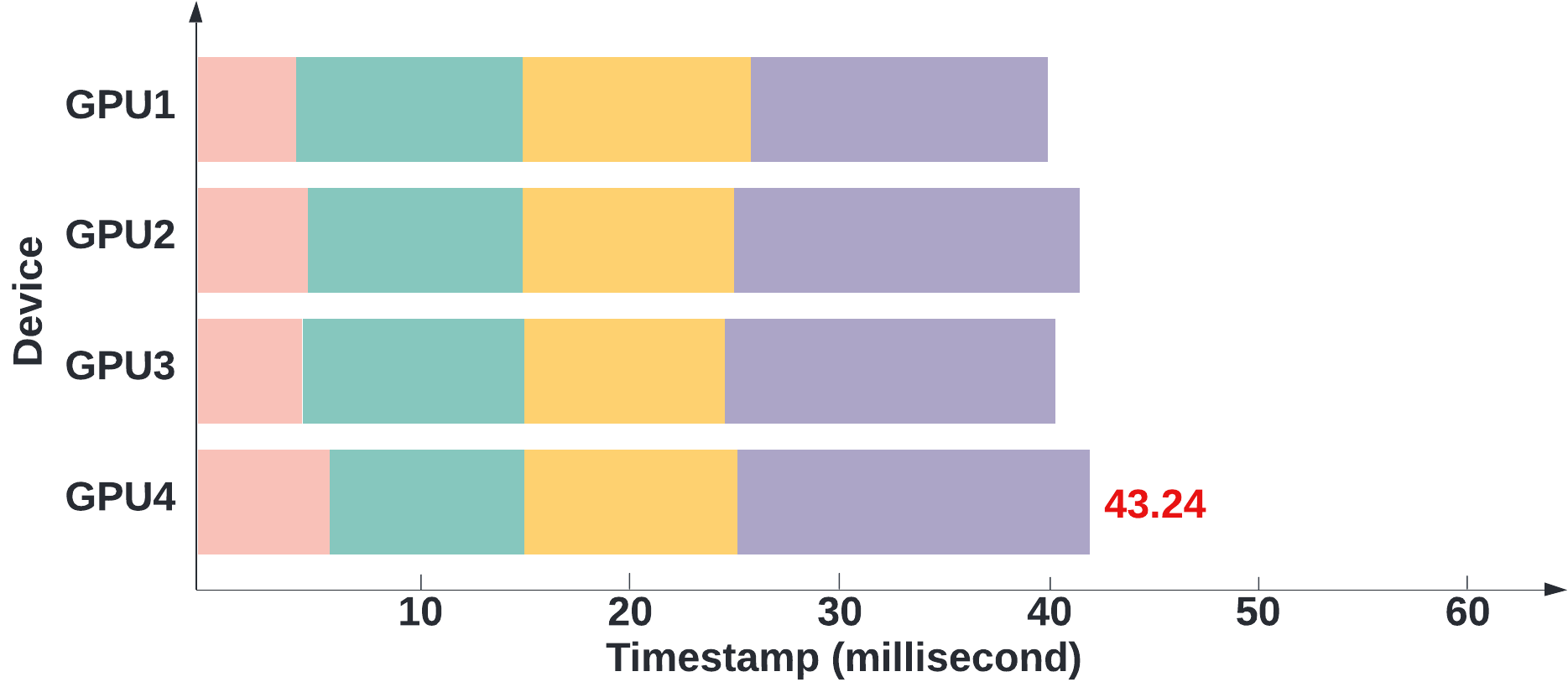}
    \subcaption{Best human expert placement}
  \end{subfigure}%
  
  \begin{subfigure}[b]{0.8\textwidth}
    \centering
    \includegraphics[width=0.99\textwidth]{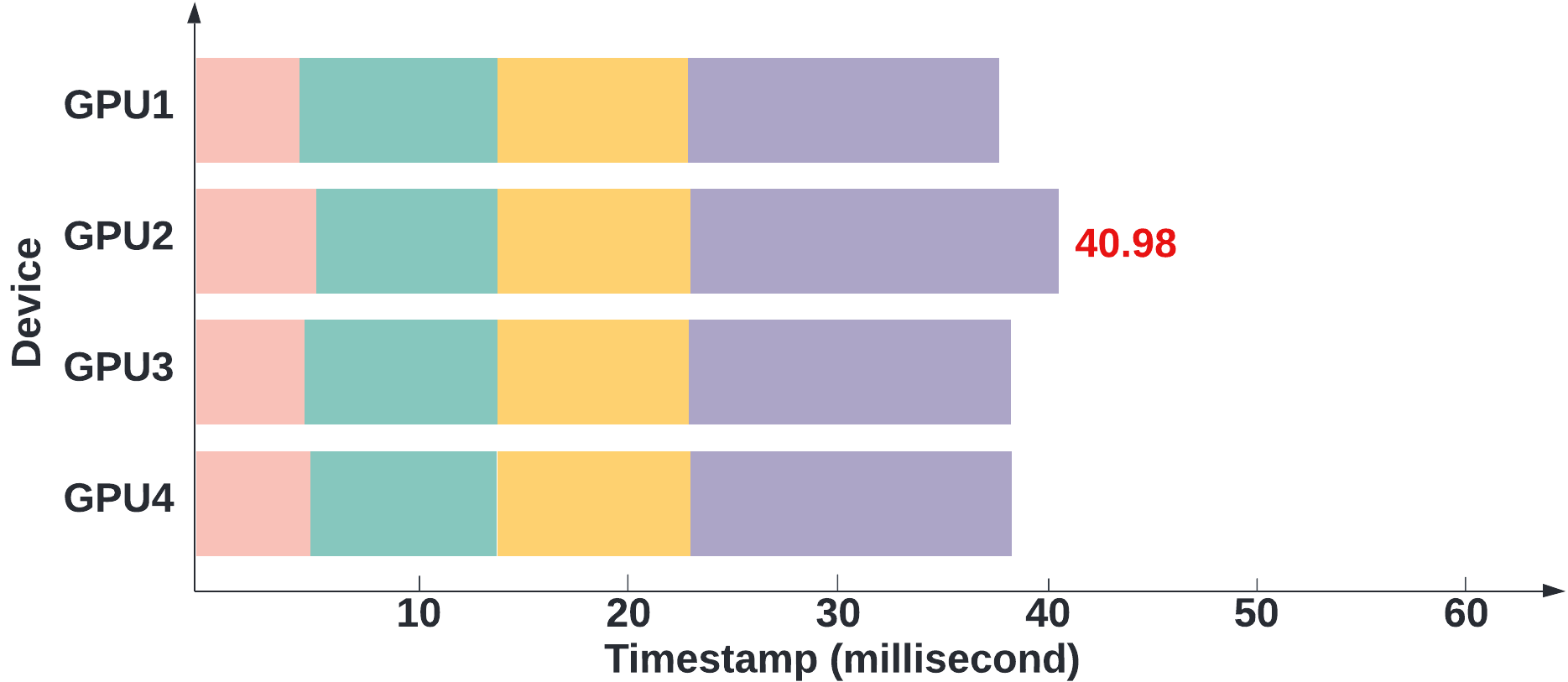}
    \subcaption{DreamShard}
  \end{subfigure}%
 
  \caption{Good case 3: visualization of DreamShard, the best heuristic algorithm, and random placement on a task of placing 50 tables to 4 GPUs. While DreamShard does not achieve a very good overall balance, the communication time appears to be less than those of the baselines potentially due to a better balance in terms of communication. As such, it still leads to a significant overall speedup.}
\end{figure}

\begin{figure}[h!]
  \centering
  \begin{subfigure}[b]{0.8\textwidth}
    \centering
    \includegraphics[width=0.99\textwidth]{figs/case/legend.pdf}
  \end{subfigure}%
  
  \begin{subfigure}[b]{0.8\textwidth}
    \centering
    \includegraphics[width=0.99\textwidth]{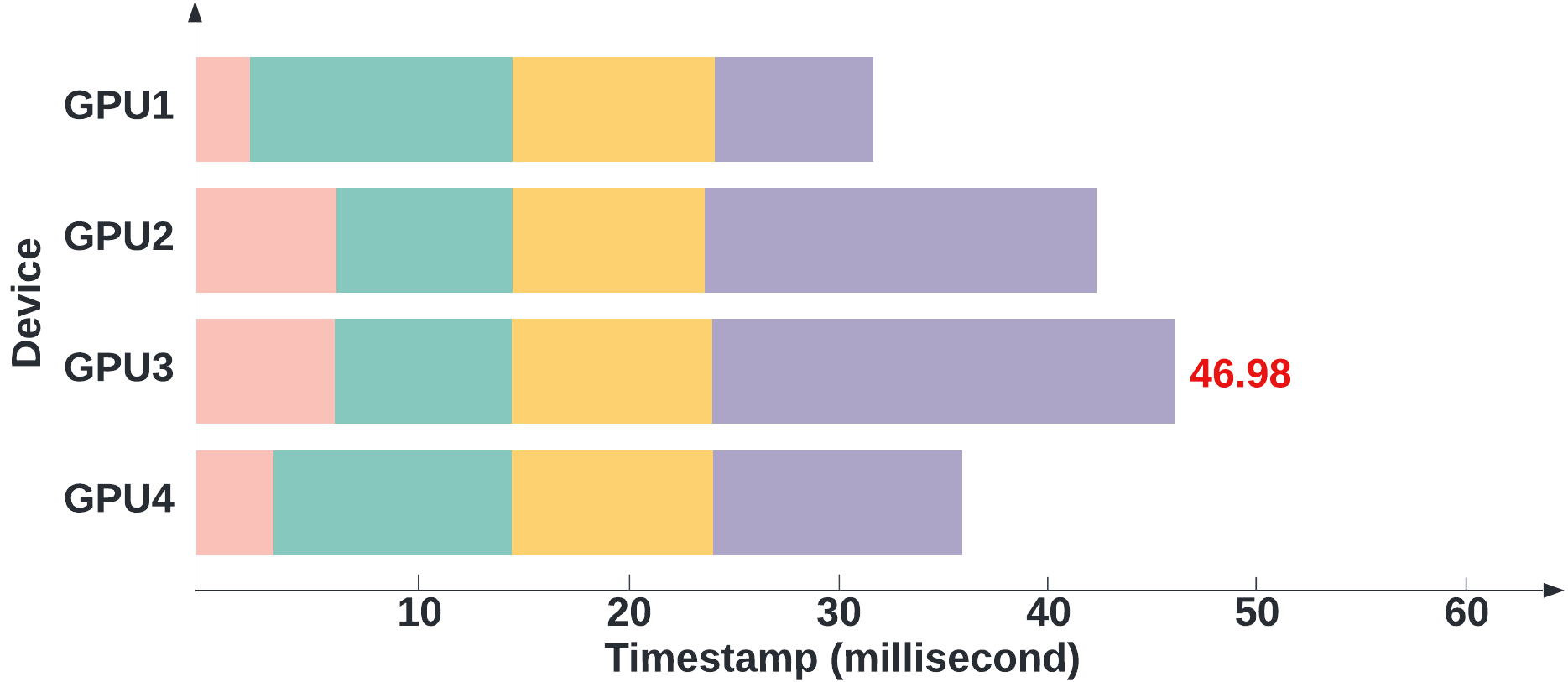}
    \subcaption{Random placement}
  \end{subfigure}%
  
  \begin{subfigure}[b]{0.8\textwidth}
    \centering
    \includegraphics[width=0.99\textwidth]{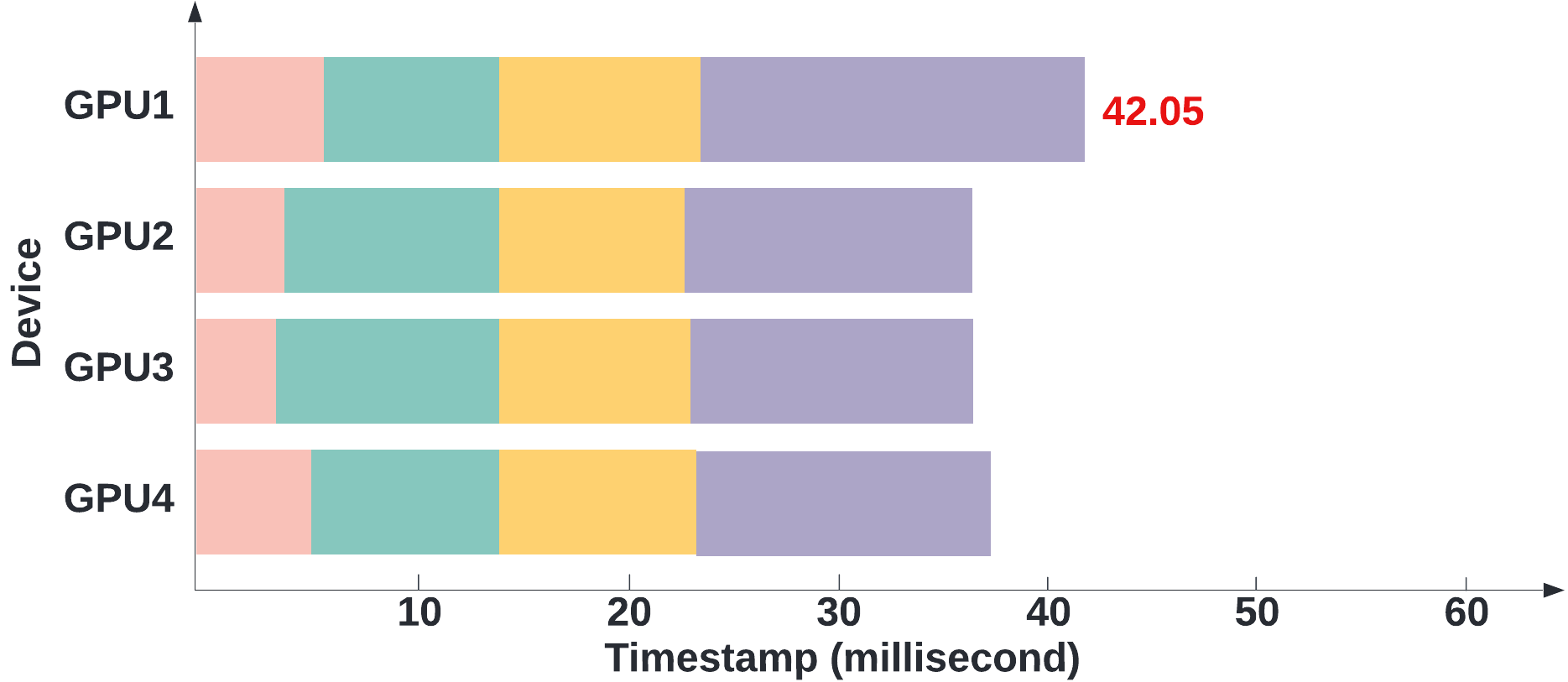}
    \subcaption{Best human expert placement}
  \end{subfigure}%
  
  \begin{subfigure}[b]{0.8\textwidth}
    \centering
    \includegraphics[width=0.99\textwidth]{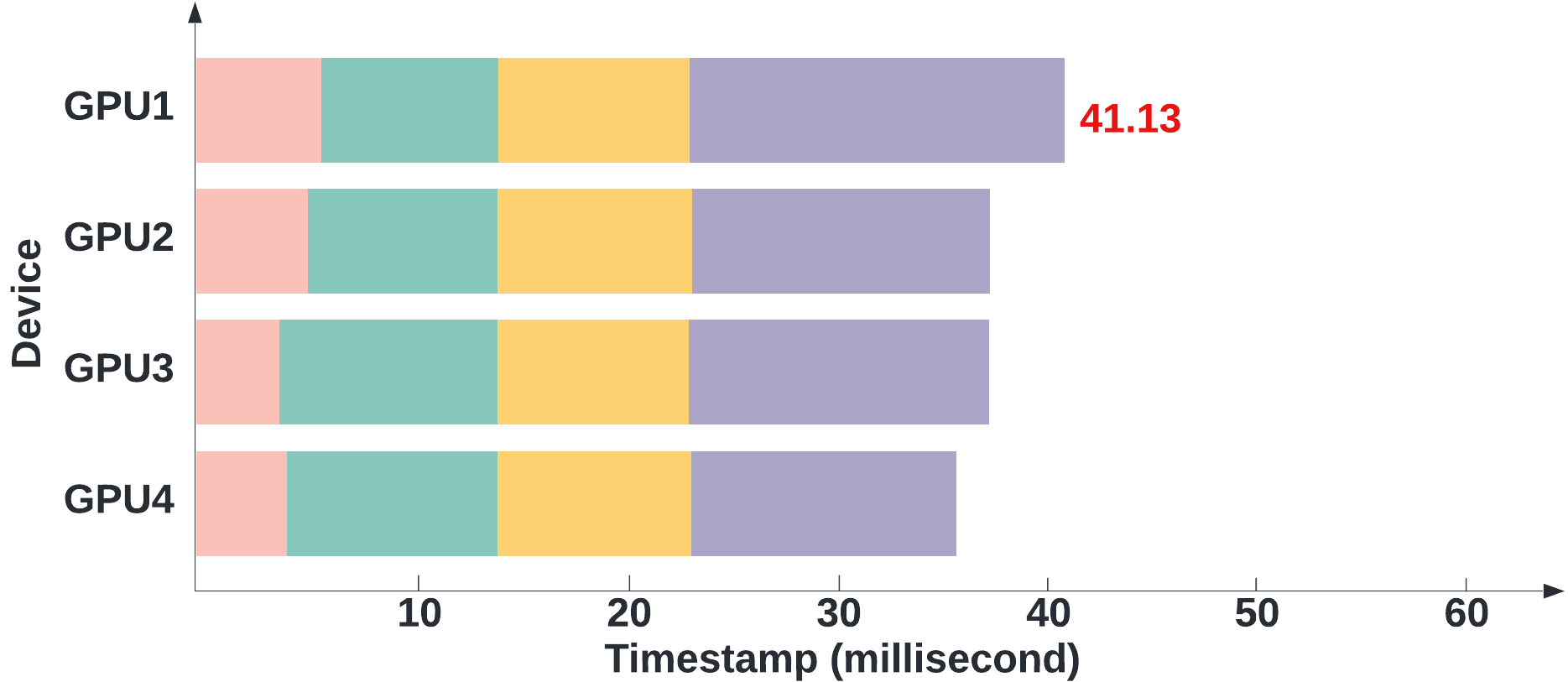}
    \subcaption{DreamShard}
  \end{subfigure}%
 
  \caption{Bad case 1: visualization of DreamShard, the best heuristic algorithm, and random placement on a task of placing 50 tables to 4 GPUs. The costs of DreamShard are not very balanced. Nevertheless, DreamShard still slightly outperforms the baselines.}
\end{figure}

\begin{figure}[h!]
  \centering
  \begin{subfigure}[b]{0.8\textwidth}
    \centering
    \includegraphics[width=0.99\textwidth]{figs/case/legend.pdf}
  \end{subfigure}%
  
  \begin{subfigure}[b]{0.8\textwidth}
    \centering
    \includegraphics[width=0.99\textwidth]{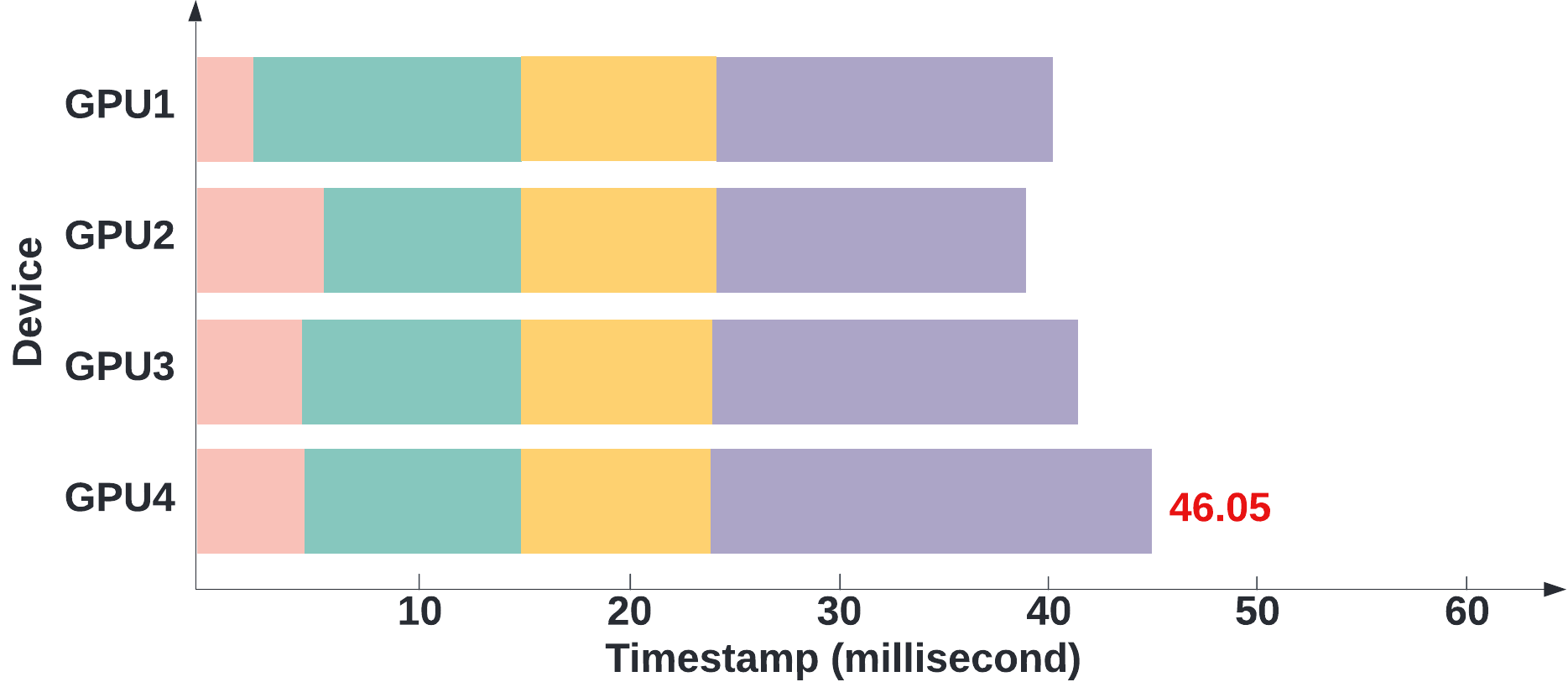}
    \subcaption{Random placement}
  \end{subfigure}%
  
  \begin{subfigure}[b]{0.8\textwidth}
    \centering
    \includegraphics[width=0.99\textwidth]{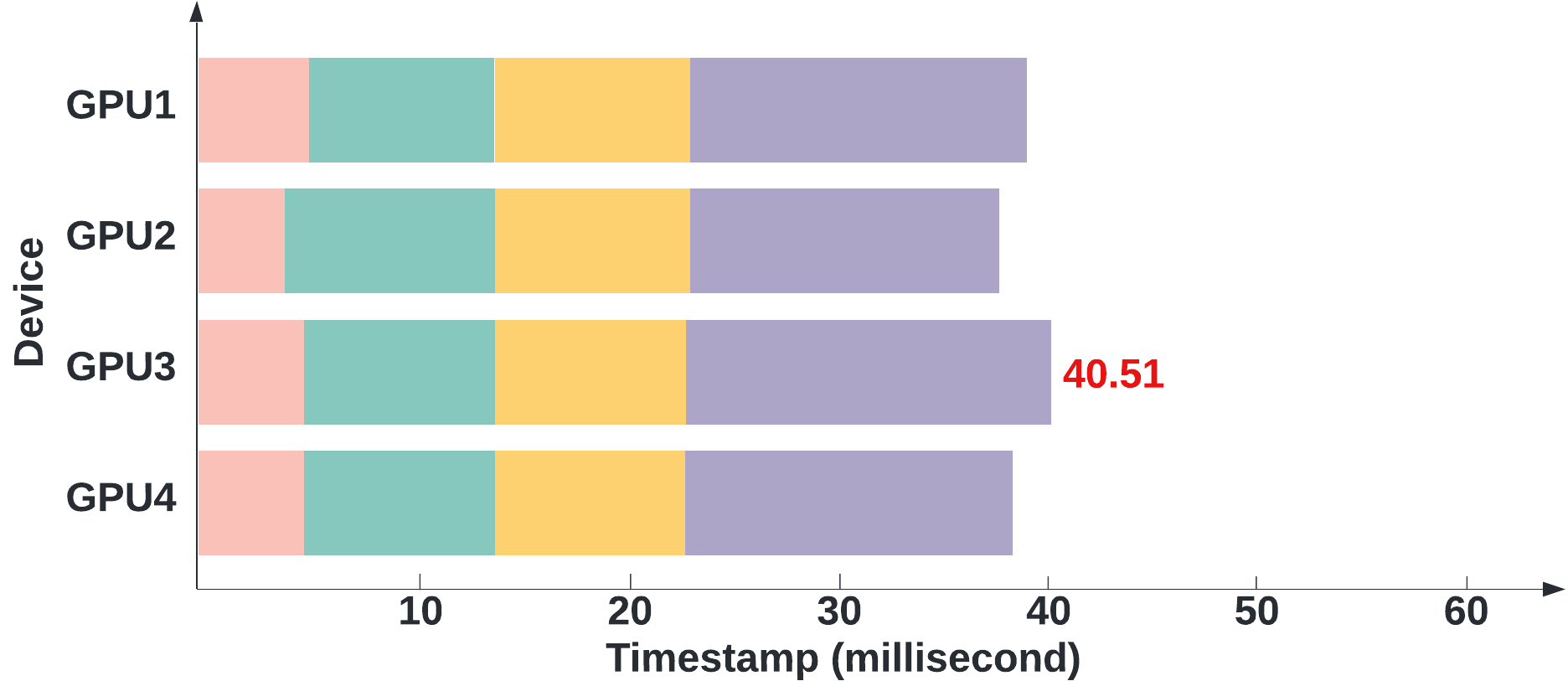}
    \subcaption{Best human expert placement}
  \end{subfigure}%
  
  \begin{subfigure}[b]{0.8\textwidth}
    \centering
    \includegraphics[width=0.99\textwidth]{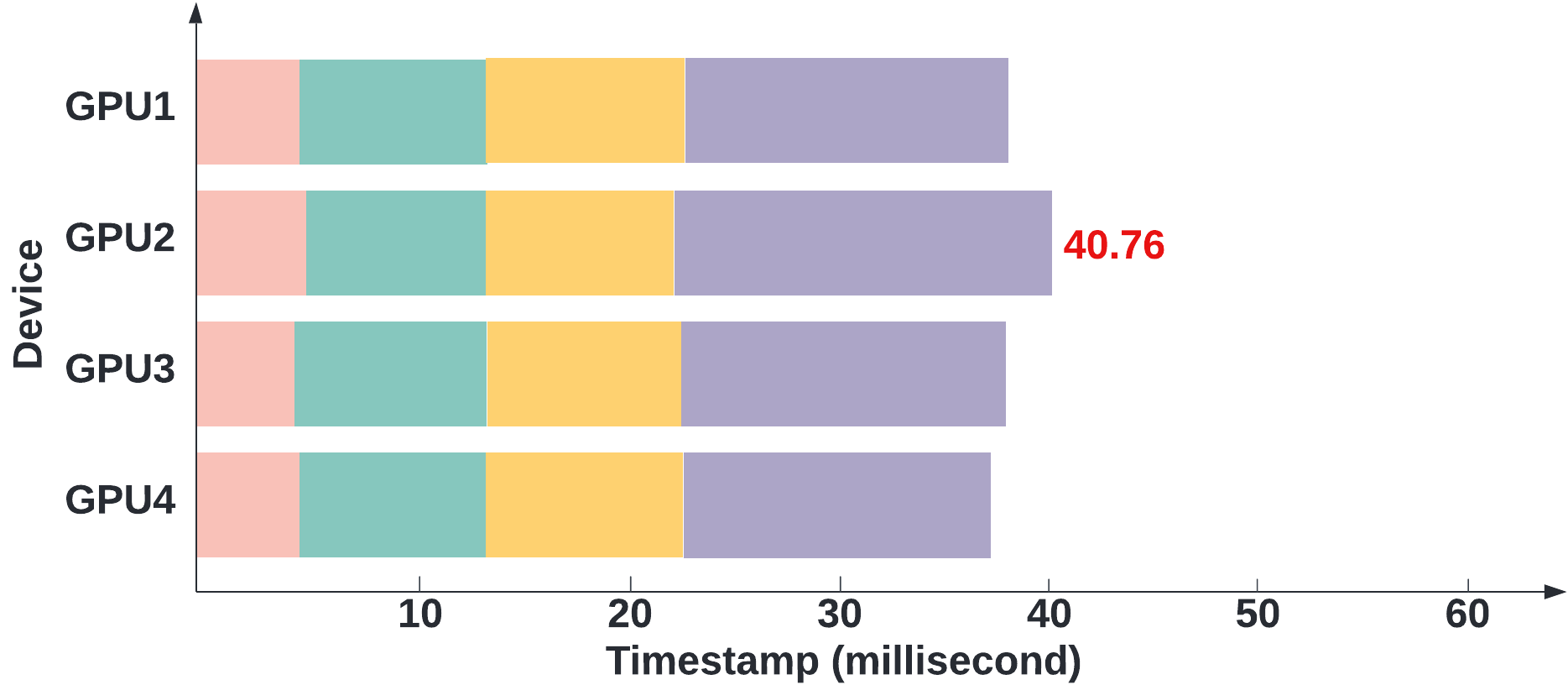}
    \subcaption{DreamShard}
  \end{subfigure}%
 
  \caption{Bad case 2: visualization of DreamShard, the best heuristic algorithm, and random placement on a task of placing 50 tables to 4 GPUs. The costs of DreamShard are slightly worse than the best heuristic. However, we find that this is actually a very rare case. In most tasks, DreamShard is either significantly better than the best heuristic or has a competitive performance with the heuristic.}
\end{figure}

\begin{figure}[h!]
  \centering
  \begin{subfigure}[b]{0.8\textwidth}
    \centering
    \includegraphics[width=0.99\textwidth]{figs/case/legend.pdf}
  \end{subfigure}%
  
  \begin{subfigure}[b]{0.8\textwidth}
    \centering
    \includegraphics[width=0.99\textwidth]{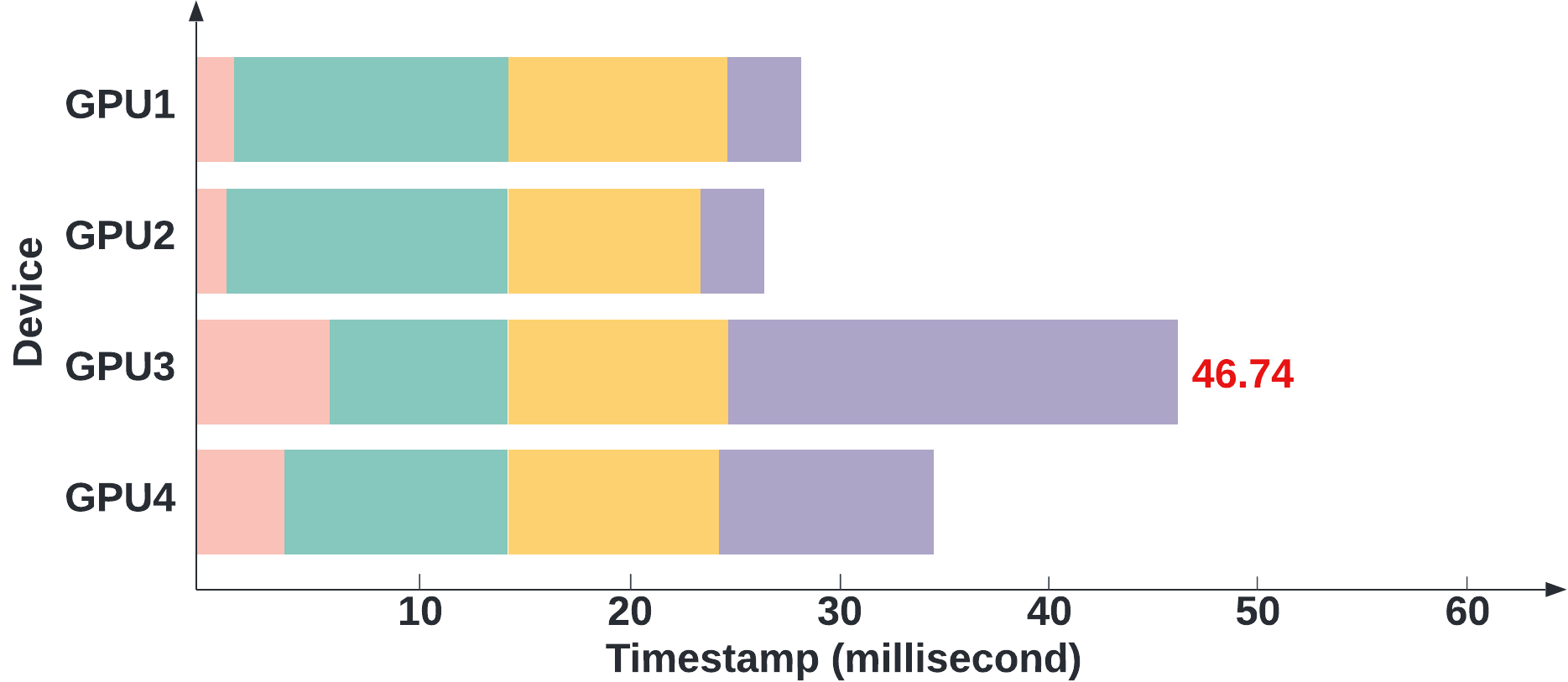}
    \subcaption{Random placement}
  \end{subfigure}%
  
  \begin{subfigure}[b]{0.8\textwidth}
    \centering
    \includegraphics[width=0.99\textwidth]{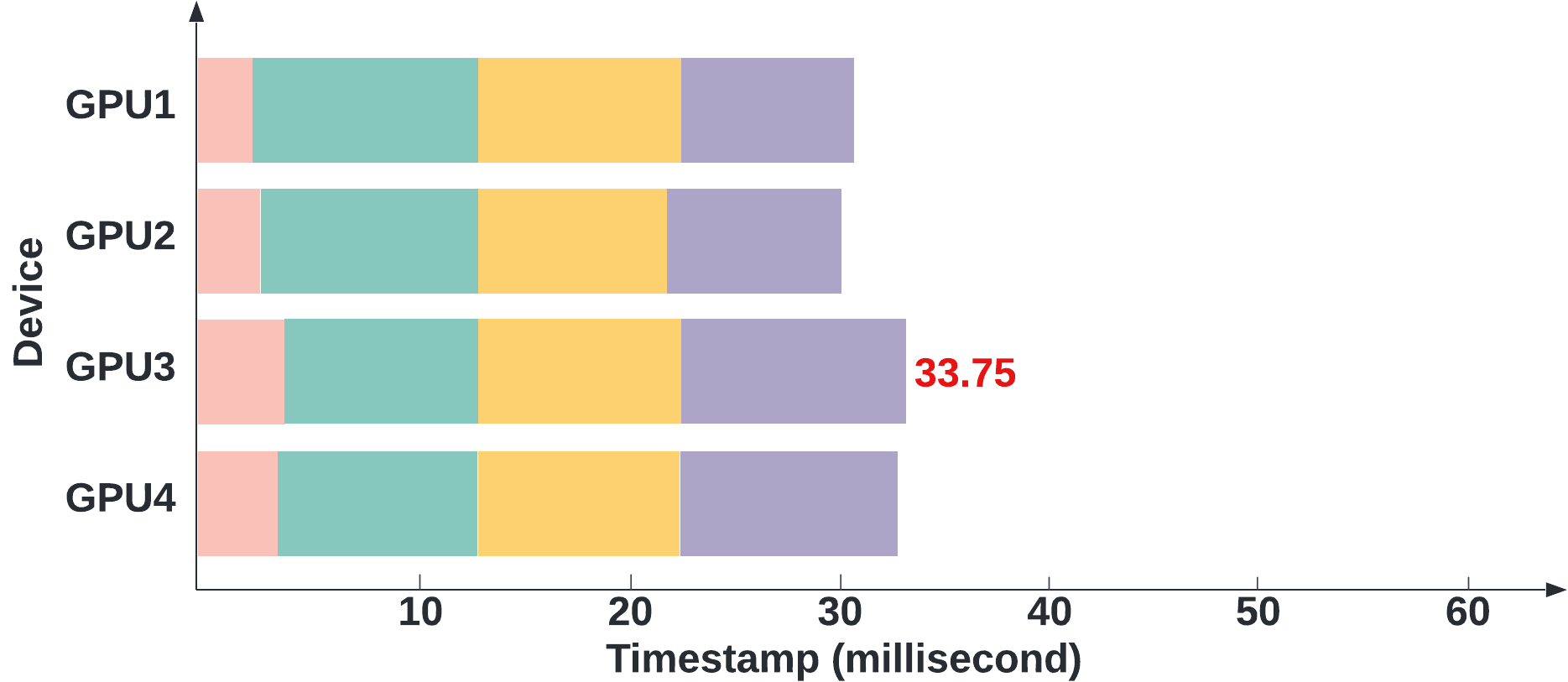}
    \subcaption{Best human expert placement}
  \end{subfigure}%
  
  \begin{subfigure}[b]{0.8\textwidth}
    \centering
    \includegraphics[width=0.99\textwidth]{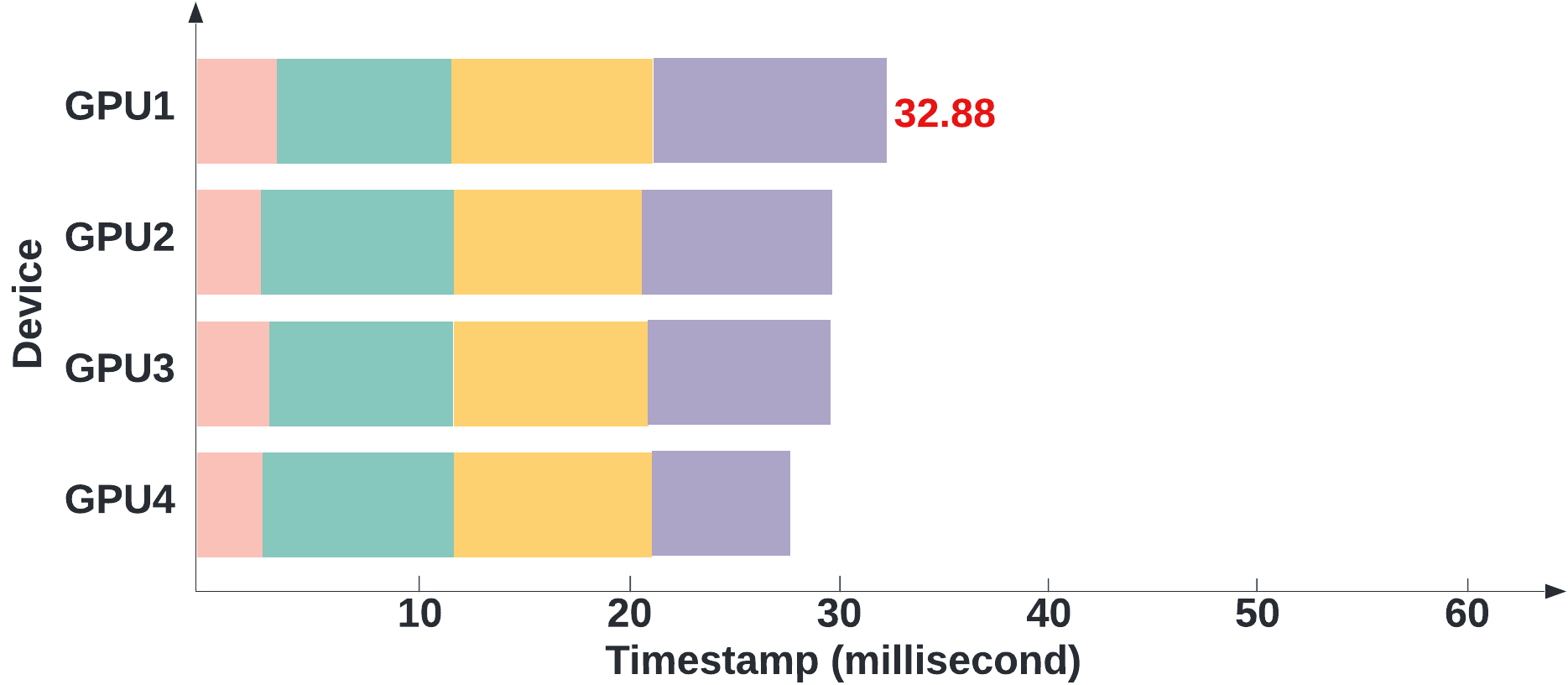}
    \subcaption{DreamShard}
  \end{subfigure}%
 
  \caption{Bad case 3: visualization of DreamShard, the best heuristic algorithm, and random placement on a task of placing 50 tables to 4 GPUs. While DreamShard achieves better results in the forward pass, it suffers from an imbalance in the backward pass. While it outperforms the best heuristic, it is still very likely to have room for improvement.}
\end{figure}

\end{document}